\pdfoutput=1
\documentclass[11pt,a4paper]{article}

\usepackage[utf8]{inputenc}
\usepackage[T1]{fontenc}
\usepackage{lmodern}
\usepackage[margin=2.4cm]{geometry}
\usepackage{amsmath,amssymb,amsthm,mathtools}
\usepackage{booktabs,array,tabularx,multirow,longtable}
\usepackage{graphicx}
\usepackage{enumitem}
\usepackage{algorithm}
\usepackage{algpseudocode}
\usepackage{microtype}
\usepackage{xcolor}
\usepackage{caption}
\usepackage{tikz}
\usetikzlibrary{positioning,arrows.meta,fit,calc,backgrounds,shapes.geometric}
\usepackage[numbers,sort&compress]{natbib}
\definecolor{teal0}{RGB}{0,110,110}
\definecolor{burnt}{RGB}{192,80,0}
\definecolor{ink}{RGB}{25,28,32}
\definecolor{softgreen}{RGB}{46,139,87}
\definecolor{grayln}{RGB}{200,200,200}
\usepackage[pdftitle={Owned Edits to a Shared Decision Rule},pdfauthor={John J. O'Hare},
            pdfcreator={},pdfproducer={},pdfkeywords={}]{hyperref}
\hypersetup{colorlinks=true,linkcolor=teal0,citecolor=teal0,urlcolor=teal0}
\theoremstyle{plain}
\newtheorem{proposition}{Proposition}
\newtheorem{lemma}[proposition]{Lemma}
\newtheorem{corollary}[proposition]{Corollary}
\theoremstyle{definition}
\newtheorem{definition}{Definition}
\newtheorem{assumption}{Assumption}
\theoremstyle{remark}
\newtheorem{remark}{Remark}

\newcommand{\X}{\mathcal{X}}
\newcommand{\Xk}[1][k]{\mathcal{X}_{#1}}
\newcommand{\Sub}{\mathcal{S}}
\newcommand{\Par}{\mathcal{P}}
\newcommand{\Hyp}{\mathcal{H}}
\newcommand{\none}{\bot}
\newcommand{\sem}[1]{[\![#1]\!]}
\newcommand{\comp}{\mathrm{comp}}
\newcommand{\kap}{\kappa}
\newcommand{\supp}{\operatorname{supp}}
\newcommand{\dec}{\delta}
\newcommand{\oplusP}{\oplus}
\newcommand{\Loc}{\Lambda}
\newcommand{\err}{\widehat{\mathrm{err}}}

\newcommand{\HC}{\mathrm{HC}}
\newcommand{\ReLU}{\operatorname{ReLU}}
\newcommand{\ind}[1]{\mathbb{1}[#1]}
\newcommand{\Z}{\mathbb{Z}}
\newcommand{\R}{\mathbb{R}}

\newcommand{\ledgerrelease}{r8}
\newcommand{\ledgerhash}{628d5b3c21a7}
\newcommand{\ledgercount}{10{,}800}
{(unbuilt)}\newcommand{\ledgerhash}{--}\newcommand{\ledgercount}{--}}

\title{\bfseries Owned Edits to a Shared Decision Rule:\\
Learning Private Exceptions as Typed, Gated, Withdrawable Weight Edits}
\author{Dr John J. O'Hare\\ \small DreamLab AI Consulting Ltd\\
\small \href{https://dreamlab-ai.com}{dreamlab-ai.com}\thanks{Every result in this paper is generated from the results ledger, release \ledgerrelease{} (\ledgercount{} entries, entry hash \texttt{\ledgerhash}); see the availability statement.}}
\date{11 November 2026}

\begin{document}
\maketitle

\begin{abstract}
\noindent
Organisations that apply the same published decision rule often hold private evidence that the rule is wrong for their own cases, and will not pool it. We describe a learning mechanism for that setting whose unit of change is an owned edit. The published rule is compiled exactly into a small ReLU network, a structured prior. Each party's learner searches a small typed space of declarations, a few named values such as subject, scope, form and size, and proposes one only when its own records pass a selection-aware evidence gate. The learning is that gated search around the prior. In two pre-registered analyses, on synthetic data whose 20 hidden truths were sampled from that typed space, the learner at 30 records per party was non-inferior, within 2 points, to all seven tuned baselines (a confined gradient learner, Ditto, FedRep, federated averaging, two pooled networks and pooled boosted trees), scoring 96.6\% against 91.7\% for the unchanged rule; a descriptive greedy ablation with no prior, over a separate space, scored 84.0\%. In a third registered test, typed edits over a network trained on shared records beat that network alone by \ensuremath{+5.4} points and the same edits over the published rule by \ensuremath{+3.1}; oracle-tuned fine-tuning and Ditto matched it at 30 records. An acceptance body recomputes each edit from its declaration and writes it into units reserved for that party, so by construction an edit changes no decision about any other subject, withdraws exactly, discloses at most $\log_2$ of the typed space's size in bits, and the artefact can hide nothing beyond its declared values; timing is not covered. We report where it loses: truths shared by all parties, relaxations, ten records, a pre-stated gate criterion that failed in both runs, and a language model given the records that decided about as accurately on one draw. On a rule-only base, a table of owned exceptions makes the same decisions.
\end{abstract}

\section{Introduction}\label{sec:intro}

\paragraph{One rule, one test record.}
A European standard for derating electronic parts in spacecraft sets the drain--source voltage that a power transistor may carry in flight at ``80 \% of rated, or the SEE safe operating area (VDS versus VGS), whichever is lower''~\cite{ecss-q-st-30-11c-rev2}. The first half of that line is arithmetic on the data sheet. The second half is a limit that only a radiation test can supply, and the same standard requires that ``only SEE radiation characterized'' transistors be flown (clause 0140136). We read an empty second term as ``pending'', not as ``accept''; that reading is ours, not the standard's. A public test compendium from NASA's Goddard Space Flight Center reports one such test on the SFF6661 power transistor, rated at 90 V, on a lot of 11 parts~\cite{gsfc-compendium-2024}. The ratio half of the rule allows 72 V. At zero gate bias the lot showed no destructive event up to 65 V, and parts failed a little above it. At the next tested bias the safe limit is 55 V, and at the highest it is 0 V: no safe drain voltage at all. For this part the test tightens the public rule, and only someone holding the test record knows by how much. Figure~\ref{fig:worked} follows this record through the mechanism.

This row happens to be public. Many are not. Of the power-transistor rows we extracted from eight public compendia, six of thirty give no result and refer the reader to the investigator or mark the data proprietary. The standard itself says that part strength ``varies from manufacturer to manufacturer, from type to type, and from lot to lot''~\cite{ecss-q-st-30-11c-rev2}, and the compendium's authors ask that their data be used cautiously, since many tests ran under lot-specific conditions~\cite{gsfc-compendium-2024}. A European Space Agency presentation on its call for ideas on commercial parts recorded ``Big demand to share information: higher when it comes to getting COTS data and lower willingness for providing it''~\cite{esa-esccon-cots}. The evidence that would correct the public rule exists, but it is scattered across the firms that paid for the tests, and each has reasons to keep it.

\paragraph{The general setting.}
Strip away the sector and the structure is this. Several organisations, which we call \emph{parties}, apply the same published decision rule to the same kind of item. Each holds private records showing where the rule is wrong, usually for items it makes, buys or tests. None will pool those records: they are commercially sensitive, sometimes export-controlled, and they reveal what the holder is working on. Yet every party would gain from a shared rule that carried the corrections the others know. Section~\ref{sec:setting} states the setting formally.

\paragraph{Why the obvious routes fail.}
\emph{Pooling} the records into a common model meets the accuracy requirement and fails the disclosure requirement by design.
\emph{Federated learning}~\cite{mcmahan-fedavg,kairouz-open-problems} keeps records local but exchanges gradients or weights, from which records can be reconstructed or inferred~\cite{zhu-dlg,geiping-inverting,melis-leakage,nasr-privacy}. Removing a party afterwards needs retraining or approximate unlearning~\cite{cao-unlearning,bourtoule-sisa,liu-federaser}. Personalised variants~\cite{collins-fedrep,li-ditto,arivazhagan-fedper} reduce interference between parties without removing it; Appendix~\ref{app:proofs} proves that a party's own records cannot tell local gradient descent to confine a correction to that party's subject.
\emph{Model editing}~\cite{meng-rome,meng-memit,mitchell-mend,hartvigsen-grace,fang-alphaedit}, including its collaborative and federated forms~\cite{zheng-collabedit,fleke-fededit} and the recent proposals to write weight edits directly~\cite{hu-wsp-scope}, edits weights freely; an audit of such edits found flaws that a reader of the weights can miss~\cite{nusrat-wsp-audit}. A free-form edit has side effects that must be tested rather than ruled out, it cannot in general be withdrawn exactly once later edits have landed, and the choice among edits that behave identically can carry information no behavioural test sees (Section~\ref{sec:fail-hidden}).
Finally, in our pilot a capable language model handed a party's records and the public rule decided held-out items as accurately as our mechanism, on one draw and a small probe (Section~\ref{sec:fail-llm}). But the records must then be in front of the model at every decision, two identical calls stated different rules, and a decision made in context has no owner and cannot be withdrawn.

\paragraph{What the mechanism does.}
The mechanism has four parts (Section~\ref{sec:mechanism}).
\begin{enumerate}[nosep]
  \item \emph{A compiled prior.} The public rule is written exactly into a small ReLU network, which makes the same decision as the rule on every valid input. This follows the knowledge-based neural network tradition~\cite{towell-kbann} and recent work compiling programs into networks~\cite{weiss-rasp,lindner-tracr}.
  \item \emph{A typed learner with an evidence gate.} Each party searches, on its own records, a small finite space of \emph{declarations}: typed statements such as ``for this subject, in this scope, tighten this limit by this much''. It submits one only if the declaration fixes enough more records than it breaks, and if a margin that large would be unlikely were the public rule right and the labels merely noisy. This is theory revision~\cite{ourston-either,richards-forte,mooney-shavlik-recap} over a deliberately small space.
  \item \emph{An owned weight patch.} A public compiler turns a declaration into a weight patch confined to network units reserved for that party's subject. The patch carries its owner, composes with other parties' patches in any order and can be withdrawn exactly, restoring the model bit for bit.
  \item \emph{An acceptance body.} A neutral party admits a declaration after re-deriving the patch itself from the declaration, so that the edit cannot differ from the stated intent, running exhaustive checks, and testing the result on a reference set that no party chose.
\end{enumerate}
What crosses from a party to the others is the declaration and nothing else. It discloses a handful of values, never the records.

\paragraph{What is proved, what is measured, and what is not shown.}
Exact compilation on the input domain, non-interference between subjects, exact withdrawal, an upper bound on disclosure and zero hidden capacity under re-derivation are properties of the construction. We prove them (Appendix~\ref{app:proofs}) and check them on every run (Section~\ref{sec:res-guarantees}). The hypothesis is a single registered endpoint: that accuracy from few records is not worse than that of seven properly sized gradient, pooled and federated baselines (Section~\ref{sec:hypothesis}). It is measured on synthetic benches with planted truths (Sections~\ref{sec:experiments} and~\ref{sec:results}). On real records, the transistor rule above compiles exactly over thirty-three million inputs, and fifteen public rows become typed declarations with no spill and exact withdrawal. That bench shows no learning: its declarations were written mechanically from the rows, and its parties are compendium year bands, not firms.

Section~\ref{sec:failures} reports where the mechanism fails, and Section~\ref{sec:limitations} states its limits. The evidence for accuracy is synthetic. The typed space was built to contain the planted truths. A language model given the records decides as accurately. The reference set the acceptance test relies on does not yet exist. Real records needed dimensions the type did not have, and the public rule, like any type that only tightens it, cannot express leniency. For a rule-initialised network, the weights add little beyond a uniform artefact format: the same mechanism works as a typed rule table. Over a trained base, on synthetic data, the combination of trained weights and typed exceptions beats each of its parts, but oracle-tuned local fine-tuning and a personalised federated method match it, and with the fewest records it loses (Section~\ref{sec:res-hybrid}).

\paragraph{Contributions.}
\begin{enumerate}[nosep]
  \item A formal setting for correcting a shared public rule from private, owner-specific evidence (Section~\ref{sec:setting}).
  \item A four-part mechanism (compiled prior, typed declarations under an evidence gate, owned and exactly withdrawable patches, and an acceptance test built on re-derivation) that meets the construction properties by design (Section~\ref{sec:mechanism}), with proofs (Appendix~\ref{app:proofs}) and checks on every run (Section~\ref{sec:res-guarantees}).
  \item A measure of hidden capacity for any edit language over weights, and a proof that typed declarations with re-derivation have zero hidden capacity, and reach the floor that the legitimate facts impose when the type's behaviour classes match those facts. In a pilot bench, undefended free-form edits hid at least thousands of bits, and defences had to close those channels one at a time (Section~\ref{sec:fail-hidden}).
  \item A pre-registered comparison against gradient, pooled and federated baselines on synthetic benches, and a compilation of a published derating rule with public test rows (Sections~\ref{sec:experiments} and~\ref{sec:results}).
  \item An account of what a deployment needs around the mechanism, including signed declarations, an append-only ledger and rationed probing of the reference set (Section~\ref{sec:protocol}), and of where the mechanism might apply in industry and how it would fail there (Section~\ref{sec:applications}).
\end{enumerate}

\paragraph{Relation to earlier work.}
The author's earlier paper~\cite{ohare-copy-ceiling} studied a boundary between a curated corpus and a language model: what the model is shown, and how much of a measured gain that exposure alone explains. This paper studies a boundary of the same kind between a party's private records and a model shared with its competitors. Here the control is placed on the artefact that crosses rather than on the input: a typed declaration whose disclosure is bounded and whose hidden capacity is zero. The work was done at DreamLab AI Consulting Ltd\footnote{\url{https://dreamlab-ai.com}}; Section~\ref{sec:protocol} describes, briefly and by maturity, which of the company's existing identity and provenance components a deployment could build on.

\section{The mechanism}\label{sec:mechanism}

This section describes the mechanism as an engineer would meet it, then states it precisely. Figure~\ref{fig:mechanism} is the whole of it in one picture. The mathematics needed to read the rest of the paper is in Sections~\ref{sec:setting} to~\ref{sec:trained-base}; the guarantees are stated in Section~\ref{sec:guarantees} and proved in Appendix~\ref{app:proofs}.

\paragraph{In one paragraph.}
Several organisations apply the same published decision rule, for example a derating standard that says which electronic parts may be flown at which voltages. Each holds private test records showing that the rule is wrong for some items it makes, buys or tests. None will pool those records. Our mechanism writes the published rule exactly into a small neural network, the \emph{compiled prior}, and lets each organisation change that network only through a \emph{declaration}: a few named values from fixed categories, such as ``for part~$k$, within a stated range of gate bias, reject above a stated voltage''. A learner running on the organisation's own premises searches the declarations it is allowed to make and proposes one only when its records clear an \emph{evidence gate}. An \emph{acceptance body} recomputes the weight edit from the declaration, checks it exhaustively, and admits it into units reserved for that organisation. The edit can later be withdrawn exactly, and it changes no decision about any other subject.

\begin{figure}[t]
\centering
\begin{tikzpicture}[
  font=\sffamily\footnotesize, >=Stealth, node distance=4mm,
  box/.style={draw=#1, fill=#1!8, rounded corners=2pt, align=center, inner sep=3pt},
  box/.default=teal0,
  lab/.style={font=\sffamily\scriptsize, text=ink!80, align=center},
  arr/.style={->, thick, draw=ink!70}]
\node[box=gray, minimum width=20mm] (rec) {Private records\\ $R_p$ (stay here)};
\node[box, right=6mm of rec, minimum width=22mm] (learn) {Learner: search\\ the typed space $\Hyp_p$};
\node[box=burnt, right=6mm of learn, minimum width=18mm] (gate) {Evidence gate\\ propose or decline};
\node[box, right=6mm of gate, minimum width=26mm] (decl) {Declaration $d$\\ subject\,$\cdot$\,scope\,$\cdot$\\ form\,$\cdot$\,size};
\draw[arr] (rec) -- (learn); \draw[arr] (learn) -- (gate); \draw[arr] (gate) -- (decl);
\node[fit=(rec)(learn)(gate), draw=gray!60, dashed, rounded corners, inner sep=4pt,
      label={[lab]above:Party $p$'s premises (repeated for every party)}] (prem) {};
\node[box=ink, below=12mm of decl, minimum width=40mm, text width=40mm, align=left] (acc)
  {\textbf{Acceptance body}\\[1pt]
   1.\ type and authorisation\\
   2.\ re-derive the edit $\kap(d)$\\
   3.\ exhaustive checks on $\X$:\\ \quad meaning, no spill, withdrawal\\
   4.\ evidence: gate statistic,\\ \quad reference test};
\draw[arr] (decl) -- node[lab, right] {only $d$ leaves} (acc);
\node[box=softgreen, left=24mm of acc, text width=44mm, align=left] (model)
  {\textbf{Shared model}\\[1pt]
   compiled prior $\theta_0=\comp(r)$,\\ same decisions as the rule\\ on every valid input,\\
   \emph{or} a trained base\\[2pt]
   reserved units:\ \fbox{A}\ \fbox{B}\ \fbox{C}};
\draw[arr] (acc) -- node[lab, above] {admit} node[lab, below] {write $p$'s units} (model);
\node[box=gray, left=12mm of model, text width=14mm] (rule) {Published\\ rule $r$};
\draw[arr] (rule) -- node[lab, above] {compile} (model);
\node[below=6mm of acc.south, anchor=north east, xshift=20mm, text width=140mm, align=left, lab] (g)
  {\textbf{By construction (Section~\ref{sec:guarantees}):} no decision outside the declared subject changes;
   withdrawal restores the model bit for bit; peers learn at most $\log_2|\Hyp_p|$ bits per declaration;
   no room to hide information beyond the declared values.
   \textbf{Tested (Section~\ref{sec:hypothesis}):} accuracy from at most 30 records against seven tuned baselines.};
\end{tikzpicture}
\caption{The mechanism. A published rule is compiled exactly into a small network. On each party's premises a learner searches the declarations that party may make and an evidence gate decides whether to propose one; only the declaration leaves. The acceptance body recomputes the weight edit from the declaration itself (it never accepts submitted weights), checks it on every valid input, weighs the evidence, and writes it into units reserved for that party. The base may instead be a network trained on shared reference records (Section~\ref{sec:trained-base}).}
\label{fig:mechanism}
\end{figure}

\paragraph{What is learned, and why we call it learning.}\label{sec:what-learned}
The learning in this mechanism is a search, under an evidence gate, over a small \emph{typed hypothesis space} centred on a \emph{structured prior}: the published rule. From a few dozen records the learner must decide two things: whether its records justify any change at all, and if so which of the permitted corrections they support. One registered test and one descriptive ablation show that both the prior and the learned parameters matter.
\begin{itemize}[nosep]
  \item \emph{The prior carries most of the accuracy.} In the second confirmatory run (Section~\ref{sec:experiments}), on truths sampled from the typed space itself, the typed learner scored 96.6\% at 30 records per party, against 91.7\% for the published rule left unchanged. A learner without the compiled prior, which composes up to three declarations greedily from a separate 151-element space of absolute declarations, starting from an empty rule and with no gate, scored 84.0\%. That arm differs from the typed learner in its space and its search as well as in the prior, and it ran inside run~2 without entering any registered comparison, so the figure is descriptive.\footnote{With the gate as well, the no-prior arm scored 43.7\%. That figure is degenerate rather than informative: the gate's null model flips the reference rule's decisions at its observed error rate, and from an empty rule that rate exceeds one half, so the gate almost never fires (deviation~1 of the first plan, recorded before any comparison). We report both arms and read the ungated one as the measure of the prior.}
  \item \emph{A trained base can carry what the type cannot.} When part of the truth is shared by every party and lies outside any typed declaration, a network trained on shared reference records carries that part, and typed declarations laid over its reserved units carry each party's exceptions. In a registered test this combination was \ensuremath{+5.4} points (95\% interval [\ensuremath{+4.4}, \ensuremath{+6.4}]) more accurate than the trained network alone, and \ensuremath{+3.1} points ([\ensuremath{+2.0}, \ensuremath{+4.1}]) more accurate than the same declarations over the published rule (Section~\ref{sec:trained-base}). Each of these two comparators lacks one component of the truth by design, so the test shows that the composition works; oracle-tuned private fine-tuning and Ditto pFL match it at 30 records (Section~\ref{sec:res-hybrid}). The gate cost accuracy here: in a registered comparison, gated minus ungated was \ensuremath{-1.3} points [\ensuremath{-1.5}, \ensuremath{-1.0}].
\end{itemize}
We then concede the point a sceptic will make. When the base is \emph{only} a compiled rule, a table of owned exceptions run beside the rule makes exactly the same decisions as our network (Corollary~\ref{cor:alone}), and so does a table run beside a trained network. What the network form adds over a table is that a trained component and the owned exceptions live in one artefact, hashed, checked and withdrawn by one procedure. Most of the contribution, the typed space, the gate, the acceptance test and the measurement of disclosure and hidden capacity, would carry over to a rule table unchanged. We say which claims depend on the network form wherever it matters.

\subsection{The setting}\label{sec:setting}

\paragraph{Items and subjects.}
Items are described by a finite set $\X$ of valid encodings. Every $x\in\X$ has a \emph{subject} $s(x)\in\Sub$: the categorical attribute a correction is about, such as a manufacturer or a part number. Subjects partition the domain into blocks $\X=\bigsqcup_{k\in\Sub}\Xk$, with $\Xk=\{x\in\X: s(x)=k\}$. The subject is encoded as a one-hot block of inputs; the other inputs lie on a lattice $\tfrac1q\Z$. Because $\X$ is finite, every claim ``for every valid input'' can be, and in our benches is, checked exhaustively. No guarantee is given for inputs outside $\X$, so a deployment must validate inputs before evaluating the model.

\paragraph{Rule, truth, parties, records.}
The \emph{published rule} $r:\X\to\{0,1\}$ (1 = accept) is public and fixed. The \emph{truth} $t:\X\to\{0,1\}$ is the rule that ought to be applied; it is unknown. Each \emph{party} $p\in\Par$ holds \emph{records} $R_p=\{(x_i,y_i)\}_{i=1}^{n_p}$ drawn from its own distribution $Q_p$, with $y_i=t(x_i)$ except for label noise at rate~$\eta$. In the simplest case each party owns one subject and observes only that block. Real data add two complications: a party may hold records on subjects it does not own (a test house), and two parties may hold records on the same subject.

\paragraph{The shared model and the acceptance body.}
All parties use one shared ReLU network $f_\theta:\X\to\R$ with parameters $\theta\in\R^{\Loc}$, where $\Loc$ is the finite set of parameter locations, and decision $\dec_\theta(x)=\ind{f_\theta(x)\ge\tfrac12}$. It starts from a state $\theta_0$ that anyone can recompute, and it changes only through artefacts that the \emph{acceptance body} admits. The acceptance body is trusted to run a stated, deterministic check. It is not trusted with any party's records, except in a dispute and then only as far as the dispute requires.

\paragraph{Running examples.}
Two benches recur. Section~\ref{sec:experiments} gives their protocols.
\begin{itemize}[nosep]
  \item \emph{Synthetic bench.} An item carries an achieved and a required level of a first quantity ($d_r,d_m\in\{0,\dots,15\}$), the same for a second quantity ($c_r,c_m\in\{0,\dots,8\}$), a class in $\{0,1,2\}$, a binary flag $h$, and a subject in $\{\mathrm{A},\mathrm{B},\mathrm{C}\}$, each subject also being a party. Each subject block has $16\cdot16\cdot9\cdot9\cdot3\cdot2=124{,}416$ valid items. The published rule is $r(x)=\ind{d_r-d_m\ge 2-h+\ind{\text{class}=0}}\wedge\ind{c_r\ge c_m}$. In the application that motivated the work, the subject is a component manufacturer, the two quantities are radiation dose and cold limit, the class is a quality grade and $h$ marks flight heritage; nothing formal depends on that reading.
  \item \emph{Real-record bench.} The rule is the drain--source voltage line of a published derating standard for power transistors: accept when the applied voltage is at most 80\% of the rating and at most a safe-operating limit that depends on gate bias, a limit that stays empty until a test fills it. Subjects are part numbers; declarations were written by a fixed procedure from public single-event test reports. Here two parties can declare on the same part.
\end{itemize}

\subsection{Part 1: the compiled prior}\label{sec:compiled-prior}

\begin{definition}[Patch, support, application]\label{def:patch}
A \emph{patch} is a finite partial map $P:\Loc\rightharpoonup\R$ with the hash of the model it was written against. Its \emph{support} is $\supp(P)=\operatorname{dom}(P)$. Applying it \emph{assigns} values:
$(\theta\oplusP P)_\lambda=P(\lambda)$ for $\lambda\in\supp(P)$ and $\theta_\lambda$ otherwise.
Application is refused if the base hash does not match, a location repeats, a value is not finite, or the patch exceeds the edit limits $(e_{\max},v_{\max},\Delta_{\max})$: at most $e_{\max}$ locations, $|P(\lambda)|\le v_{\max}$ and $|P(\lambda)-\theta_\lambda|\le\Delta_{\max}$.
\end{definition}

Assignment, not addition, is deliberate: it makes withdrawal exact in floating point, where $(\theta+\tau)-\tau\neq\theta$ in general. The model's hash is a cryptographic hash of a canonical serialisation in which equal reals print identically.

\begin{definition}[Threshold--conjunction rule]\label{def:tc-rule}
An \emph{atom} is a predicate $\ind{w\cdot x\ge c}$ with $w\in\Z^n$ and $c\in\tfrac1q\Z$. A \emph{threshold--conjunction rule} has the form
\begin{equation}\label{eq:tc-rule}
  r(x)=\Bigl(\textstyle\bigwedge_{i=1}^{m}\alpha_i(x)\Bigr)\wedge\neg\textstyle\bigvee_{j=1}^{J}\Bigl(\bigwedge_{l=1}^{L_j}\alpha_{jl}(x)\Bigr),
\end{equation}
with every $\alpha$ an atom. The first conjunction is the base rule; each inner conjunction is an \emph{exception}, a region the rule rejects.
\end{definition}

Membership of a subject or a class is an atom on a one-hot input, and ``at most the lower of two limits'' is a conjunction of two atoms, so both benches' rules have this form. We use two constructions, each exact on $\X$ (Proposition~\ref{prop:exact}).

\paragraph{Construction M (margin form, one hidden layer).}
With $q=1$, each base atom gets a hidden unit computing its \emph{violation} $c_i-w_i\cdot x$, an integer that is at most $0$ exactly when the atom holds, and the output is $f(x)=1-\sum_i\ReLU(c_i-w_i\cdot x)$. A single-atom exception $\ind{\ell(x)\ge1}$ restricted to subject $k$ and a scope $\sigma$ is one further unit with output weight $-1$ and pre-activation $\ell(x)-G(1-x_{s,k})-G(1-\sigma(x))$, where the \emph{gate weight} $G\ge\max(1,\max_{\X}\ell)$ silences the unit outside its subject and scope. The synthetic bench uses this form: 11 inputs, 6 hidden units and 1 linear output, 79 parameters. Units 0 and 1 carry the two base atoms; units 2, 3 and 4 are spare and \emph{reserved}, one per subject.

\paragraph{Construction I (indicator form, two hidden layers).}
On the lattice, $\ind{w\cdot x\ge c}=q\bigl(\ReLU(w\cdot x-c+\tfrac1q)-\ReLU(w\cdot x-c)\bigr)$, two first-layer units per atom. A conjunction of indicators $b_1,\dots,b_L$ is one second-layer unit $\ReLU(\sum_l b_l-(L-1))$, and the rule~\eqref{eq:tc-rule} is the output $\ReLU(b_0-\sum_j e_j)$. The real-record bench uses this form with $q=256$: 21{,}213 parameters, 48 reserved exception slots, and 33 million valid inputs, on every one of which the compiled network agreed with the rule (0 disagreements).

\subsection{Part 2: typed declarations}\label{sec:typed}

\begin{definition}[Declaration, typed space, compile map]\label{def:decl}
A \emph{declaration type} is a tuple of \emph{slots}, each with a finite domain; one slot is always the subject. A \emph{declaration} is either the empty declaration $\none$ or a well-formed assignment of values to the slots. Party $p$'s \emph{typed space} $\Hyp_p$ is the finite set of declarations it may submit; it always contains $\none$, and it names only subjects $p$ is authorised for. The \emph{semantics} of $d$ is an exception region $\sem{d}\subseteq\Xk$, where $k$ is $d$'s subject and $\sem{\none}=\emptyset$; the refined rule is $r_d=r\wedge\neg\sem{d}$. Every declaration here \emph{tightens}. The \emph{compile map} $\kap$ sends $d$ to a patch writing $\sem{d}$ into the units reserved for $d$'s slot; it is a deterministic public function of $d$ and the architecture, and it is undefined when the patch would exceed the edit limits.
\end{definition}

\paragraph{Synthetic bench: three nested spaces.}
A declaration names the party's own subject, one of six \emph{scopes} (all items, one of three classes, $h=0$, $h=1$) and a linear score $\ell_d$ with integer coefficients; it rejects in-scope items with $\ell_d(x)\ge1$. The narrow spaces fix the score to a named \emph{form} with a \emph{size}: a first-quantity margin raised by $\Delta\in\{1,\dots,8\}$, a second-quantity margin $\Gamma\in\{1,\dots,4\}$, or a second-quantity threshold $\tau\in\{0,\dots,8\}$. This gives spaces nested by syntax: $\Hyp^{(1)}$ (margins only, $|\Hyp^{(1)}|=73$), $\Hyp^{(2)}$ (all three forms, $|\Hyp^{(2)}|=127$) and $\Hyp^{(3)}$ (any score with coefficients in $\{-2,\dots,2\}$, flag and class terms in $\{-1,0,1\}$ and offset in $\{-16,\dots,16\}$; $|\Hyp^{(3)}|=1{,}113{,}751$). The confirmatory runs use $\Hyp^{(2)}$ for the primary arm and $\Hyp^{(3)}$ as the ``free-form linear'' ablation. A narrow declaration discloses four values: subject, scope, form and size. Its patch writes only the reserved unit $u_k$: the score's coefficients as incoming weights, the gate weight on subject $k$'s input and on the scope input, the bias, and output weight $-1$. Patches for different subjects therefore have disjoint support.

\paragraph{Real-record bench.}
A declaration is a list of slots $(k,g_{\mathrm{lo}},g_{\mathrm{hi}},L)$: reject part $k$ when $g_{\mathrm{lo}}\le g\le g_{\mathrm{hi}}$ and the voltage exceeds $L$. Each slot is one exception of Construction~I, written into its own reserved index (six first-layer units, one conjunction unit gated by the part's pass-through unit, output weight $-1$). In the run on public test reports, 35 slots from 15 report rows needed 19 to 82 edits per declaration.

\subsection{Part 3: the learner and the evidence gate}\label{sec:learner}

The learner maps a party's records to one declaration in its typed space. Choosing \emph{which} declaration (a tie-break) is kept separate from deciding \emph{whether} to change anything (the gate).

\paragraph{Search.}
For records $R$ let $\err(d;R)=\sum_{(x,y)\in R}\ind{r_d(x)\neq y}$. Because $\Hyp_p$ is finite, the learner enumerates it, finds the least record error $E^\ast$, and picks from the tie set $\{d:\err(d;R_p)=E^\ast\}$ by a fixed strict order~$\prec$. Two orders are used. The \emph{pre-stated} order $\prec_{\mathrm{pre}}$ prefers fewer nonzero fields, then smaller coefficients, then smaller offset, with $\none$ first. The \emph{minimal-change} order $\prec_{\mathrm{mc}}$ prefers the declaration that flips the fewest of the subject's published decisions, then $\prec_{\mathrm{pre}}$; it is the minimal-revision principle of theory revision. It uses no truth and no records beyond $R_p$. Minimal change was introduced after the pilot and is the primary order in both confirmatory runs; the pre-stated order is reported beside it.

\paragraph{The evidence gate, as run.}\label{sec:gate}
For the chosen declaration $\hat d$, let $a$ be the number of records it decides correctly and the published rule incorrectly, and $b$ the reverse, so that $T=a-b=\err(\none;R_p)-\err(\hat d;R_p)$ is the improvement in record error. The gate passes when both
\begin{equation}\label{eq:gate}
  T\ge m \qquad\text{and}\qquad \hat p\le\alpha,
\end{equation}
where $\hat p$ is a \emph{selection-aware} Monte Carlo $p$-value. It is computed as follows. Set $\hat\varepsilon=\err(\none;R_p)/n$. Draw $K=199$ replicate label vectors in which each record's label is the published rule's decision, flipped independently with probability $\hat\varepsilon$. On each replicate, rerun the \emph{whole} search and record the best improvement $T^\ast_k$ over the published rule. Then $\hat p=(1+\#\{k:T^\ast_k\ge T\})/(K+1)$, and $\hat p=1$ when $T=0$. Because each replicate repeats the search, the $p$-value accounts for choosing the best of many candidates. Two settings were registered: the \emph{judgement} setting $(m,\alpha)=(3,0.05)$, used in the first confirmatory run, and the \emph{tuned} setting $(1,0.2)$, chosen on separate development data as the most accurate setting whose false-proposal rate stayed under 5\%, and used in the second. The nominal $\alpha$ is a tuning parameter, not the realised false-proposal rate; that rate is measured (Section~\ref{sec:guarantees}).

\paragraph{An analysable variant.}
If the noise rate $\eta$ is stated in advance, the gate can instead use the fixed threshold $c_\alpha(n,\eta)=\min\{c\ge1:\Pr[\mathrm{Bin}(n,\eta)\ge c]\le\alpha\}$ and pass when $T\ge c_\alpha(n,\eta)$. Proposition~\ref{prop:gate} bounds the false-update rate of this variant, and of the Monte Carlo gate when its replicates use the true noise rate. The experiments ran the Monte Carlo gate with the estimated rate $\hat\varepsilon$, for which the bound is not exact.

\begin{algorithm}[t]
\caption{Typed learner with evidence gate (party $p$, run locally on $R_p$)}\label{alg:learn}
\begin{algorithmic}[1]
\Require published rule $r$; typed space $\Hyp_p\ni\none$; order $\prec$; gate setting $(m,\alpha)$; replicates $K$; records $R_p$, $|R_p|=n$
\State $\hat d\gets\min_\prec\arg\min_{d\in\Hyp_p}\err(d;R_p)$ \Comment{exhaustive enumeration}
\State $T\gets\err(\none;R_p)-\err(\hat d;R_p)$;\quad $\hat\varepsilon\gets\err(\none;R_p)/n$
\For{$k=1,\dots,K$} \Comment{selection-aware null}
  \State $\tilde y_i\gets r(x_i)\oplus\mathrm{Bernoulli}(\hat\varepsilon)$ for each record
  \State $T^\ast_k\gets\err(\none;\tilde R)-\min_{d\in\Hyp_p}\err(d;\tilde R)$ \Comment{rerun the whole search}
\EndFor
\State $\hat p\gets(1+\#\{k:T^\ast_k\ge T\})/(K+1)$ \textbf{if} $T>0$ \textbf{else} $1$
\If{$T\ge m$ \textbf{and} $\hat p\le\alpha$} \Comment{evidence gate}
  \State $\mathit{com}\gets\textsc{Commit}(R_p;\text{salt})$ \Comment{opened to the acceptance body only in a dispute}
  \State \Return $(\hat d, T, \mathit{com})$
\Else
  \State \Return $\none$
\EndIf
\end{algorithmic}
\end{algorithm}

\subsection{Part 4: the acceptance test, composition and withdrawal}\label{sec:acceptance}

\begin{definition}[Composition and withdrawal]\label{def:compose}
After admitting $d_1,\dots,d_K$ in order, the shared state is $\theta_K=\theta_0\oplusP\kap(d_1)\oplusP\cdots\oplusP\kap(d_K)$; the admitted declarations not yet withdrawn form the \emph{live set}. The inverse of a patch $P$ restores its support to the initial values, $P^{-1}:\lambda\mapsto(\theta_0)_\lambda$ for $\lambda\in\supp(P)$. Withdrawing $d_i$ applies $\kap(d_i)^{-1}$. The inverse depends only on $\theta_0$ and $\supp(P)$, so the acceptance body can withdraw a declaration without the withdrawing party's records or cooperation.
\end{definition}

\paragraph{Re-derivation and canonical form.}
Parties submit declarations, never weights. The acceptance body discards any weights that arrive with a declaration and computes the patch itself as $\kap(d)$, after replacing $d$ by the least-index declaration in $\Hyp_p$ with the same behaviour on $\X$ (computable by enumeration, since both sets are finite). Re-derivation removes every freedom below the level of the declaration, such as number spelling, edit order, idle edits and null-space directions; the canonical representative removes the freedom that remains among declarations that behave identically. This is what makes the hidden-capacity result of Section~\ref{sec:guarantees} hold. Because the declaration \emph{is} the stated intent and the patch is computed from it, a separate check that ``the edit matches its author's intent'' is not needed for typed declarations. It is needed only for free-form weight edits, which we use as an ablation.

\begin{algorithm}[t]
\caption{Acceptance test and withdrawal (acceptance body)}\label{alg:accept}
\begin{algorithmic}[1]
\Require state $\theta$, live set $\mathcal{D}$; submission $(d,T,\mathit{com})$ signed by $p$; reference set $R_{\mathrm{ref}}$; margin $\mu\ge0$
\State \textbf{if} $d\notin\Hyp_p$ \textbf{then reject} \Comment{type and authorisation}
\State $d\gets\mathrm{can}(d)$;\quad $P\gets\kap(d)$;\quad \textbf{if} $P$ undefined \textbf{then reject} \Comment{re-derive; submitted weights ignored}
\State \textbf{if} $\supp(P)\cap\supp(\kap(d'))\neq\emptyset$ for some $d'\in\mathcal{D}$ \textbf{then reject} \Comment{slot taken}
\State $\theta'\gets\theta\oplusP P$;\quad $k\gets$ subject of $d$
\State \textbf{if} $\exists x\in\Xk:\dec_{\theta'}(x)\neq(\dec_\theta(x)\wedge\neg\sem{d}(x))$ \textbf{then reject} \Comment{meaning, exhaustive}
\State \textbf{if} $\exists x\in\X\setminus\Xk:\dec_{\theta'}(x)\neq\dec_\theta(x)$ \textbf{then reject} \Comment{no spill, exhaustive}
\State \textbf{if} $\mathrm{hash}(\theta'\oplusP P^{-1})\neq\mathrm{hash}(\theta)$ \textbf{then reject} \Comment{withdrawal}
\State \textbf{if} the gate statistic $T$ fails~\eqref{eq:gate} \textbf{then reject} \Comment{$\mathit{com}$ opened only in a dispute}
\State \textbf{if} $\err_{\theta'}(R_{\mathrm{ref}}\cap\Xk)>\err_{\theta}(R_{\mathrm{ref}}\cap\Xk)+\mu$ \textbf{then reject} \Comment{reference test, subject $k$}
\State admit: $\theta\gets\theta'$; $\mathcal{D}\gets\mathcal{D}\cup\{d\}$; publish $(d,p,\mathrm{hash}(\theta))$; keep $(T,\mathit{com})$ with the body
\Statex
\Procedure{Withdraw}{$d_i$}
  \State $\theta\gets\theta\oplusP\kap(d_i)^{-1}$;\quad $\mathcal{D}\gets\mathcal{D}\setminus\{d_i\}$;\quad log the withdrawal
\EndProcedure
\end{algorithmic}
\end{algorithm}

Algorithm~\ref{alg:accept} has three groups of checks: \emph{syntactic} (lines 1--3), \emph{exhaustive behavioural} (lines 5--7) and \emph{evidential} (lines 8--9). The first two need no records, and for typed declarations the exhaustive checks cannot fail once the syntactic ones pass (Propositions~\ref{prop:exact}, \ref{prop:noninterference} and~\ref{prop:withdraw}); we keep them as a check on the implementation, at the cost of one pass over $\X$. Two points limit the evidential group. The reference test checks only the declared subject's own error, because non-interference with other subjects is already settled exhaustively by line~6. And it needs a reference set, labelled items held by the acceptance body that no party chose, which does not yet exist for any application we know of (Section~\ref{sec:limitations}). Line~8 trusts the party's reported statistic unless a dispute opens the commitment.

\paragraph{Why parties do not vote.}\label{rem:no-votes}
In a pilot variant, parties could veto a proposal on the evidence of their own private records. A veto protected only the regions the vetoing party had sampled, and a party making repeated proposals could read another party's private limit from its votes: in nine probes it read a limit of 37 against a true 38. Each vote is a function of private records, so each discloses information outside the bound of Proposition~\ref{prop:disclosure-bound}. The acceptance test therefore polls no party.

\subsection{The mechanism over a trained base}\label{sec:trained-base}

Nothing in the declaration language, the compile map's support or withdrawal requires the base to be a compiled rule. Let $\theta_b$ be any ReLU network with reserved units held at zero during its training, for example one trained on shared reference records, with decision $\dec_b$. Give the reserved unit of declaration $d$ the output weight $-M_d$ in place of $-1$, with
\[
  M_d>\max_{x\in\sem{d}}f_{\theta_b}(x)-\tfrac12,
\]
computable exhaustively on a finite $\X$. Where $d$ fires the unit's activation is at least 1, so the output falls below $\tfrac12$; elsewhere the unit is silent. Then $\dec_{\theta_b\oplus\kap(d)}=\dec_b\wedge\neg\sem{d}$ on $\X$, non-interference holds unchanged because its conditions do not mention how the base was obtained, and withdrawal is exact relative to $\theta_b$ (Corollary~\ref{cor:trained}). The edit overrides the trained network's decisions on the declared region; it does not change trained weights.

This construction has now been run under a registered plan (Section~\ref{sec:experiments}). The truths there have two parts: a component shared by all three parties that lies outside the typed space (an interaction the type cannot express, or a relaxation, which no tightening can express), and an owner-specific tightening for each party. The base, an 11--17--1 network whose three reserved units are masked at zero throughout training, learns from 300 shared reference records labelled by the shared component only. In every one of 1{,}800 rows, the combination changed no output of another party (0), matched $\dec_b\wedge\neg\sem{d}$ on every valid input (0 mismatches), and withdrew exactly. Its accuracy result is the one quoted above. Section~\ref{sec:res-hybrid} reports where it does not win: an exploratory check run after unblinding found that oracle-tuned private fine-tuning and Ditto match it at 30 records per party (differences \ensuremath{-0.3} and \ensuremath{-0.1} points, intervals including zero), and that at 10 records, where its gate declines in 96\% of cases, fine-tuning from a well-trained base beats it. The claim that stands is the registered one: the combination beats each of its parts, while keeping exact non-interference and exact withdrawal and disclosing about 2.2 declaration values per party, against 210 to about 992 uploaded values for the pooled and personalised federated comparators.

\subsection{What is guaranteed by construction}\label{sec:guarantees}\label{sec:properties}

The results below are properties of the construction. Once it is fixed they are theorems, so in experiments they are \emph{acceptance checks}: a failure would indicate a fault in the implementation, not a refutation of a hypothesis. Most are elementary. We state them because parties who do not trust one another should be given the guarantees in writing, with the points where they stop. Proofs are in Appendix~\ref{app:proofs}. Two standing assumptions apply.

\begin{assumption}[Valid encoding]\label{ass:encoding}
Every input in $\X$ has exactly one subject bit set, at most one bit set in each other one-hot block, and every other coordinate on the lattice $\tfrac1q\Z$ within stated bounds; $\X$ is exactly the set of such inputs.
\end{assumption}

\begin{assumption}[Exact arithmetic]\label{ass:float}
Weights are integers; biases and inputs are integer multiples of $2^{-e}$; and every pre-activation, product and partial sum in the forward pass has magnitude below $2^{52-e}$ on $\X$. Then every intermediate value is exactly representable in IEEE-754 binary64, and the computed forward pass equals the exact one in any order of summation.
\end{assumption}

In the synthetic bench all values are small integers ($e=0$); in the real-record bench weights are small integers or $\pm256$ and inputs are multiples of $2^{-8}$. Both are far inside the bound.

\begin{proposition}[Exact compilation]\label{prop:exact}
For a threshold--conjunction rule $r$ and $\theta_0=\comp(r)$ built by Construction~I, or by Construction~M when $q=1$ and every exception is a gated single atom, $\dec_{\theta_0}(x)=r(x)$ for every $x\in\X$ under Assumption~\ref{ass:encoding}, and the binary64 forward pass agrees under Assumption~\ref{ass:float}.
\end{proposition}

\begin{proposition}[Gated support gives non-interference]\label{prop:noninterference}
Let $P$ be a patch concerning subject $k$, writing only the incoming weights, biases and outgoing weights of a set $W$ of units that are spare in $\theta$ and disjoint from the units of every other live patch. If every unit of $W$ that feeds a unit outside $W$ is silent on every $x\in\X\setminus\Xk$ under $\theta\oplusP P$, then $f_{\theta\oplusP P}(x)=f_\theta(x)$ on $\X\setminus\Xk$, so no decision about another subject changes. The compile maps of both benches satisfy the condition for every declaration they compile. Patches with disjoint supports also commute, so the shared state does not depend on the order of admission.
\end{proposition}

\begin{corollary}[Each subject decides as if it alone had declared]\label{cor:alone}
With at most one live declaration $d_k$ per subject, $\dec_{\theta}(x)=r(x)\wedge\neg\sem{d_k}(x)$ for every $x\in\Xk$, in any order of admission. (In the real-record bench, $\sem{d_k}$ is replaced by the union of the live slots on part $k$; there a shared subject's limit is the minimum of the parties' declared limits, which is order-free but is not non-interference between those two parties.)
\end{corollary}

\begin{corollary}[Over a trained base]\label{cor:trained}
With the output weight $-M_d$ of Section~\ref{sec:trained-base}, $\dec_{\theta_b\oplus\kap(d)}=\dec_b\wedge\neg\sem{d}$ on $\X$, and Propositions~\ref{prop:noninterference} and~\ref{prop:withdraw} hold with $\theta_b$ in place of $\theta_0$.
\end{corollary}

\begin{proposition}[Exact withdrawal]\label{prop:withdraw}
For patches $P_1,\dots,P_K$ with pairwise disjoint supports, admitted in any order onto $\theta_0$ to give $\theta_K$, $\theta_K\oplusP P_i^{-1}=\theta_0\oplusP\bigoplus_{j\neq i}P_j$ as real vectors, the two sides have equal hashes, and the equality survives any later admission or withdrawal of disjoint patches. No retraining is involved.
\end{proposition}

These three properties do not distinguish the typed learner from every alternative. A gradient learner confined to the same reserved unit, which we run as a baseline, also has zero spill and exact withdrawal. They distinguish it from unconfined, pooled and federated training. The next result says why confinement must be imposed rather than learned.

\begin{proposition}[A party's own records cannot identify confinement]\label{prop:identify}
If party $k$'s records lie in $\Xk$, any two hypotheses that agree on $\Xk$ have equal empirical loss on them, however they differ elsewhere. In particular, gradient descent on one party's records moves a unit's subject weight and its bias by equal amounts at every step (under conditions stated in Appendix~\ref{app:proofs}), so it shifts other subjects' pre-activations as well.
\end{proposition}

\begin{proposition}[Occam bound for a finite typed space]\label{prop:occam}
If party $k$'s records are $n$ independent noise-free draws from $Q_k$, and the truth on $\Xk$ equals $r_{d^\ast}$ for some $d^\ast\in\Hyp_k$ (\emph{realisability}), then a learner returning a declaration with zero record error satisfies, with probability at least $1-\delta$,
$\Pr_{x\sim Q_k}[r_{\hat d}(x)\neq t(x)]\le(\ln|\Hyp_k|+\ln(1/\delta))/n$.
\end{proposition}

The bound explains the ordering of the spaces ($\ln|\Hyp|$ is $4.29$, $4.84$ and $13.92$ nats for the three synthetic-bench spaces; Table~\ref{tab:occam} gives the bound), not the observed magnitudes: at 30 records it is weaker than the published rule's own error. Realisability holds in our synthetic benches because the typed space was built to contain the owner-specific truths; Appendix~\ref{app:proofs} discusses the agnostic case and a comparison with a parametric class.

\begin{proposition}[False updates under the evidence gate]\label{prop:gate}
Suppose the published rule is correct for party $p$ on the support of $Q_p$, and each record's label is flipped independently with probability at most $\eta$. Then, for any typed space and any tie-break,
\begin{enumerate}[nosep,label=(\roman*)]
  \item the fixed-threshold gate proposes some $\hat d\neq\none$ with probability at most $\Pr[\mathrm{Bin}(n,\eta)\ge c_\alpha(n,\eta)]\le\alpha$;
  \item if each label is flipped with probability exactly $\eta$, the Monte Carlo gate~\eqref{eq:gate} with replicates drawn at $\eta$ in place of $\hat\varepsilon$ proposes with probability at most $\lfloor\alpha(K+1)\rfloor/(K+1)\le\alpha$, for any $m\ge0$.
\end{enumerate}
\end{proposition}

Both parts hold uniformly over the search: no correction for the number of candidates is needed, because every candidate's improvement is bounded by the number of flipped labels, and because each replicate repeats the search. Neither part covers the gate as run, whose replicates use the estimated rate $\hat\varepsilon$. Its false-proposal rate is therefore a measured quantity, not a guarantee. On a fresh null run (labels equal to the published rule's decisions with 10\% flipped, so every proposal is false), the tuned setting proposed in 4.3\% of party-draws pooled over 10, 30 and 100 records, rising to 6.2\% at 100 records, above the 5\% cap at which it was chosen; the judgement setting proposed in 0.2\%. That run was written after the second confirmatory run and is exploratory. The gate is not designed to detect records that do not fit its null at all, and it does not: on shuffled outcomes it proposed in 46\% (judgement) and 76\% (tuned) of party-draws at 30 records, against a pre-stated acceptable rate of 10\%. That pre-stated criterion failed in both runs (Section~\ref{sec:failures}).

\begin{proposition}[Upper bound on disclosure to peers]\label{prop:disclosure-bound}
Let $W$ be any private quantity of party $p$, with $(W,R_p)$ jointly distributed arbitrarily, and let the learner be a public function of $R_p$ (randomised independently of $W$). Let $A_p$ be what peers see: the declaration, its patch, the resulting hashes and public metadata $M$, with the gate statistic and commitment held by the acceptance body. Then
$I(W;A_p\mid M)\le I(R_p;A_p\mid M)\le H(d\mid M)\le\log_2|\Hyp_p|$ bits per declaration, summed over a sequence of declarations and withdrawals.
\end{proposition}

For the primary screening space the bound is $\log_2 127\approx6.99$ bits per party, whether or not the party proposes. The declared values themselves are disclosed in full, by design, including the subject; the bound limits what \emph{else} peers can learn. It is not differential privacy. For gradient, federated and pooled baselines we can state only an upper bound of the same kind (at most the information in the records), so the ratio of the two bounds that we report is a ratio of upper bounds, not of information disclosed.

\begin{definition}[Capacities of an edit language]\label{def:hidden}
An edit language has messages $\mathcal{M}$, an acceptance predicate, a normaliser $\nu$ applied by the acceptance body (re-derivation, canonical form, or the identity) that preserves behaviour on $\X$, and a realisation whose behaviour on $\X$ we write $\beta(m)$. With $V$ the set of legitimate facts and $\mathcal{M}_{\mathrm{acc}}$ the accepted messages, the \emph{channel capacity} is $C=\log_2|\nu(\mathcal{M}_{\mathrm{acc}})|$ and the \emph{hidden capacity} is $\HC=\max_b\log_2|\{\nu(m):m\in\mathcal{M}_{\mathrm{acc}},\beta(m)=b\}|$: the bits carried by the choice among artefacts that behave identically, which no behavioural test can detect.
\end{definition}

\begin{proposition}[Typed declarations with re-derivation attain the floor]\label{prop:hidden-floor}
Every edit language able to express each legitimate fact has $C\ge\log_2|V|$, all of it available to a dishonest party. For typed declarations with re-derivation and the least-index canonical form, $\HC=0$ and $C=\log_2|\Hyp_p/{\equiv_\X}|$, which equals the floor $\log_2|V|$ exactly when the type's behaviour classes correspond one to one with the agreed facts.
\end{proposition}

The result needs re-derivation: if the body accepted submitted weights, number spelling alone would be an unbounded channel. For free-form weight patches we can only bound hidden capacity from below, by counting: in a ranking bench, one patch under the default edit limits hid 7,989 bits with no change in behaviour on any valid input. The typed slot for the same bench carries its floor of 3.46 bits, the declared fact itself. Channels outside the artefact, such as when a party submits or withdraws, are not covered.

\paragraph{Where checks were exhaustive.}
In the synthetic benches each declaration's compiled patch was applied alone and checked on every other party's full grid of valid items, and its withdrawal was checked against the composition of the other parties' patches; in every draw of both confirmatory runs the primary arm had zero spill and exact withdrawal. In the real-record bench each declaration's patch was checked on a spot-check subset of ratings and withdrawn exactly; the composed model was checked on all 33 million valid inputs, where it agreed with the rule plus declared limits (0 disagreements) and gave the same model in either order of admission.

\subsection{What is tested: the registered hypothesis}\label{sec:hypothesis}\label{sec:guarantees-vs-hypothesis}

The guarantees above are not part of the hypothesis. The hypothesis consists only of claims that could fail empirically, each with a decision rule fixed before the data were drawn. The analysis plans define the rules as follows.

\paragraph{Primary endpoint (two confirmatory runs).}
At the primary cell (owner-specific truths inside the typed space, no label noise, 30 records per party), the typed learner with the minimal-change tie-break and the evidence gate is \emph{non-inferior}, with a margin of 2 accuracy points, to each of seven baselines, every one tuned by cross-validation on its own records: a gradient learner confined to the same reserved unit; the personalised federated methods Ditto and FedRep; federated averaging; a pooled network of the same 79-parameter size; a pooled network sized to the data; and pooled gradient-boosted trees. The unit of inference is the hidden truth: 20 truths sampled at random from a registered generator, each the mean of 5 record draws. For each baseline $B$, the test takes the mean truth-level paired difference (typed minus $B$) and its two-sided 95\% $t$-interval with 19 degrees of freedom; non-inferiority to $B$ requires the interval's lower bound to exceed $-2$ points. The endpoint passes only if this holds for all seven baselines. This is an intersection--union test~\cite{berger-hsu-iut}, which needs no multiplicity correction because every component must pass at its own level.

The plans also list typed spill above zero, or an inexact withdrawal, in any draw as falsifying the hypothesis. By Section~\ref{sec:guarantees} such a failure could only come from a fault in the implementation, so we read those clauses as implementation checks written into the plan; none was triggered.

\paragraph{Secondary comparisons.}
Superiority over each baseline, the effect of the gate, the effect of the tie-break and (second run only) the tuned against the judgement gate form a secondary family: ten comparisons in the first run and eleven in the second. Each uses a sign-flip randomisation test on the truth-level differences (200{,}000 resamples), with Holm's step-down correction across the family at $\alpha=0.05$ and a Wilcoxon signed-rank test as a check.

\paragraph{Trained-base endpoint.}
A separate registered plan tests the hybrid of Section~\ref{sec:trained-base} at 300 shared and 30 private records per party: it must be more accurate than the trained base alone and than gated typed declarations over the published rule, on 40 sampled truths (two families of 20; $t$-intervals with 39 degrees of freedom), with zero spill and exact withdrawal in every row. Six secondary comparisons are Holm-corrected on sign-flip $p$-values.

\paragraph{Pre-stated criteria that are not endpoints.}
The plans also fix acceptable rates for the gate: at most 10\% of party-draws proposing on shuffled outcomes, and at most 5\% harmful proposals at 10 records. Truth classes outside the typed space (a component shared by all parties, an interaction the type lacks, a relaxation) are reported with the same prominence as the primary cell, but as exploratory comparisons. The reader that would turn prose test reports into typed records has not been evaluated; no claim is made for it.

\paragraph{The language-model comparison.}
A frontier language model given a subject's records and the published rule, with no tools, was asked to decide held-out items directly. That arm was run once, on one draw, before the registered plans, and is reported separately from the seven baselines: it decided 347 of 360 probe items correctly, against 348 of 360 for the patched network (Section~\ref{sec:fail-llm}).

\section{Where the mechanism might apply}\label{sec:applications}

Everything in this section is \emph{prospective}. It rests on desk research into published standards, public datasets and public statements. We contacted no organisation, and we claim no customer, partner, demand or market. The one application with any run behind it is the first, and even there the run shows compilation and typed declarations on public records, not learning (Section~\ref{sec:fail-real}).

\paragraph{What an application needs.}
The mechanism earns its place only where four conditions hold together.
\begin{enumerate}[nosep,label=(\roman*)]
  \item \emph{A small, explicit public rule known to be imperfect.} Without one there is nothing to compile. If the rule already captures what the parties know, a declaration adds nothing.
  \item \emph{Private, owner-specific evidence held by parties who will not pool it.} If the evidence is public, or the holders already share it, pooling is the right answer.
  \item \emph{A declaration that is tolerable to reveal.} A declaration discloses its values and its signer. If those values are themselves the secret, the mechanism leaks exactly what it was meant to protect.
  \item \emph{Scarce evidence per holder, and value in attribution or withdrawal.} If each holder has abundant data, a party can learn alone; if nobody cares who made a correction or whether it can be retracted, the owned patch is overhead.
\end{enumerate}
For each candidate below we give the public rule, who holds the private exceptions and why they will not pool, what a declaration would reveal, who could act as acceptance body, and what would make it fail. Table~\ref{tab:applications} summarises them.

\begin{table}[t]
\centering
\small
\renewcommand{\arraystretch}{1.15}
\setlength{\tabcolsep}{3pt}
\begin{tabularx}{\textwidth}{@{}*{6}{>{\raggedright\arraybackslash}X}@{}}
\toprule
Application & Public rule & Private holders; why no pooling & A declaration reveals & Possible acceptance body & Most likely failure \\
\midrule
Component derating and qualification (worked example) & Per-family load ratios with a test-derived safe-operating limit~\cite{ecss-q-st-30-11c-rev2} & Test houses, integrators' parts engineers, suppliers; test cost, export control, procurement intent & Part, slot values, signer; hence which part the signer is qualifying & The standard's specification authority for qualified parts; open for commercial parts & Slot too small for real records (energy, lot); no body willing to carry liability \\
Earth-observation calibration and data products & Published cross-sensor adjustment coefficients; published cloud-screening thresholds & Operators of commercial sensors, each owning one sensor's calibration evidence & Gain and offset for the signer's own sensor & A third-party mission-quality assessment programme & Corrections are continuous; for masks, truth is public and labels plentiful \\
In-situ sensor-network quality control & Published plausibility flags for soil-moisture records & Network operators holding site calibration and maintenance logs & A site-specific limit, e.g.\ where a site saturates & The network data centre or its quality service & Operators already share data, so the case for not pooling is weak \\
Space-weather anomaly screening & Public alert thresholds on the particle environment & Satellite operators holding anomaly logs; customer confidence, security & That a bus family is susceptible above some level: a vulnerability & A national space-weather centre, or a trusted operators' association & What is declared is itself sensitive; anonymity preferred to attribution \\
Other sectors (unassessed) & Varies & Varies & Varies & Varies & Not examined \\
\bottomrule
\end{tabularx}
\caption{Prospective applications. None is evidenced as a use of the mechanism; the entries summarise desk research into public rules, data and institutions. The first row is the only one with a run behind it, and that run shows compilation and typed declarations on public records, not learning.}
\label{tab:applications}
\end{table}

\subsection{Civil space: component derating and qualification}\label{sec:app-derating}

\paragraph{The public rule.}
The European derating standard sets load ratios per component family: voltage, current, power and junction temperature as fractions of the rating~\cite{ecss-q-st-30-11c-rev2}. For power transistors and some diodes, the voltage limit is the lower of a fixed ratio and a safe-operating limit measured under radiation, as in the introduction. The rule is small enough to compile exactly; Section~\ref{sec:exp-real} describes the compiled form. The standard states that its ratios ``were derived from information available at the time of writing'' and that part strength ``varies from manufacturer to manufacturer, from type to type, and from lot to lot''. A companion standard on radiation hardness assurance sets tolerance limits whose statistical factor grows as the sample shrinks~\cite{ecss-q-st-60-15c-rev1}. That factor covers dose drift only, not destructive pass or fail, so it is not our evidence gate; it shows only that the sector already prices small samples.

\paragraph{Who holds the exceptions, and why they will not pool.}
Radiation test houses, integrators' parts engineers and component suppliers hold lot-level test results. Single-event tests on commercial parts, run at a few values of linear energy transfer (LET), typically use very few samples, for example 5+5+1 samples for dose tests and 2 samples at 2 LETs for destructive single-event tests in one European test programme, so each holder sees little. Test time at a beam facility is scarce and paid for privately. The agency presentation quoted in the introduction contrasts high demand for getting commercial-parts data with low willingness to provide it~\cite{esa-esccon-cots}. A United States government effort to pool radiation test data across agencies aims to ``Eliminate duplicative or wasteful utilization of limited access at radiation test facilities''~\cite{yarbrough-srhec}, through a vetted, government-only repository rather than an industry pool. A United States audit of how defence agencies report suspect counterfeit parts to the government--industry data exchange found likely under-reporting and ``no standardized process for establishing how much evidence is needed before reporting''~\cite{gao-16-236}. That last finding is the gap an evidence gate addresses: a stated, checkable threshold for when a correction may be submitted.

\paragraph{An institution that already exists.}
The standard already admits exceptions. A type- or manufacturer-specific derating rule in an approved detail specification ``For a particular type or manufacturer, when a specific derating rule is defined in the appendix of the approved ESCC detail specification issued by the ESCC Executive, it shall take precedence over the generic requirement''. Users must ``shall clearly identify in the Parts Stress Analysis document the list of the ESCC Derating exceptions taken into consideration''~\cite{ecss-q-st-30-11c-rev2}. The public register of such deviations\footnote{\url{https://escies.org/webdocument/showArticle?id=825&groupid=6}} lists 14 entries covering 19 component types, and records the component, family code and specification, with no field for owner, value, evidence or withdrawal. A declaration supplies exactly those fields, and the hash of the live declaration set is a natural thing for the stress analysis to cite. This is our inference about fit, not a statement that the institution wants it.

\paragraph{What a declaration would reveal.}
In the real-record bench, each declaration names a part and gives a few numbers per slot (gate bias, limit, evidence count), and it carries the signer. For the SFF6661 record of the introduction, the declaration has 3 slots, one for each of three gate biases, and discloses nine numbers; it compiles to 59 weight edits and changes the decision for 0 other parts (Figure~\ref{fig:worked}). The values are engineering data about a part, not price or capacity. The signer is more sensitive: it reveals which part a party is qualifying, which matters to competitors and to manufacturers. A deployment may need declarations signed by the acceptance body on a party's behalf, which trades attribution for discretion.

\paragraph{Who could act as acceptance body.}
For parts with an approved detail specification, the natural candidate is the specification authority the standard already names, which issues the exceptions today; that is our inference from the standard's text. For commercial parts, which are the growth area and have no route into the deviation register, we found no body that approves exceptions, so the role is open. Test houses author the reports and sell the testing, which makes them parties rather than neutral judges, though one could act as the acceptance body's laboratory for re-checking disputed records.

\paragraph{What was run.}
The public transistor rule compiles exactly, checked over thirty-three million inputs with 0 disagreements. Of 30 rows extracted, fifteen became 35 typed slots, compiled into patches of 19 to 82 weight edits each. The composed model disagreed with the refined rule on 0 inputs and was the same model in either order, and every withdrawal was exact. Of the rows that state a rating, 6 tighten the rule at a tested bias. The source data are public use~\cite{gsfc-compendium-2014,gsfc-compendium-2015,gsfc-compendium-2016,gsfc-compendium-2017,gsfc-compendium-2019,gsfc-compendium-2023,gsfc-compendium-2024,gsfc-compendium-2025}.

\paragraph{What would make it fail.}
The real rows have already shown four failure routes (Section~\ref{sec:fail-real}). Destructive outcomes depend on particle energy and on the lot, which the slot did not record, so the two parties' declarations on the same part could not be compared. Several readings, 12 of 49 bias points, say ``no failure up to'' a voltage, and the rule's ``whichever is lower'' discards them, so the public rule cannot express leniency. By our own binomial arithmetic, supporting a relaxation by survival alone would need 22 clean parts, against a largest lot of 15 in these rows. Beyond the data, the deployment can fail on institutions: the specification authority may decline the liability of approving a learned screen; the standard may be quoted, but reproducing it needs the publisher's consent, so publishing the compiled weights would need permission; and integrators may refuse to cite any exception their customer has not approved. The screen informs engineering judgement; tailoring and approval stay with the customer.

\subsection{Earth observation: calibration and data-product consortia}\label{sec:app-eo}

\paragraph{The public rule.}
Earth-observation products rest on published thresholds and coefficients. Two kinds matter here. Cloud screening uses threshold tests on reflectance and temperature; the first pass of the widely used Fmask algorithm is a conjunction of such tests\footnote{Zhu and Woodcock, \emph{Remote Sensing of Environment}, \url{https://doi.org/10.1016/j.rse.2011.10.028}.}. Harmonised products combine sensors through published per-band adjustment coefficients\footnote{\url{https://hls.gsfc.nasa.gov/algorithms/bandpass-adjustment/}}. We have compiled the first pass of the published cloud-screening rule into a network of 3,707 parameters, which agreed with the rule on about 4.0 billion checked inputs, with 0 disagreements. That is a compilation result only; we report no learning result for this rule.

\paragraph{Why cloud screening is a test bed, not an application.}
Cloud masks fail the second condition. Where the rule errs, it errs by surface and region, not by owner, and the truth is visible in public imagery. Labelled reference scenes are public and plentiful, and operational products are moving from fixed thresholds to contextual and learned methods. The rule is a good test of the mechanics on continuous, real data, and a poor case for the mechanism.

\paragraph{Where an owner structure exists: calibration.}
Cross-sensor calibration has the structure the mechanism needs. Each commercial sensor has one owner, which mirrors the part-and-manufacturer structure of derating. The owner holds vicarious calibration and crossover evidence that refines the published adjustment for its own sensor, and calibration sites and crossovers are few. Recalibrations are versioned, so exact withdrawal has a use. A declaration would reveal a gain and offset for the signer's own sensor, which operators may already partly publish in product metadata; we have not checked how much.

\paragraph{Who could act as acceptance body.}
A European programme already performs early data-quality assessments of national and commercial Earth-observation missions\footnote{\url{https://telespazio.co.uk/en/ground-systems/data-quality/earthnet-data-assessment-pilot-edap}}. A body of that kind, with access to reference sites, is a natural candidate. We have not asked whether such a programme would take the role.

\paragraph{What would make it fail.}
Calibration corrections are continuous, and a typed slot quantises them; whether a coarse grid of shifts is accurate enough is untested. Open data contain no private party, so no public experiment can show the owner-specific effect. Operators may prefer bilateral assessment to any shared model. The domain needs calibration and validation expertise that a deployment would have to bring in.

\subsection{In-situ sensor-network quality control}\label{sec:app-insitu}

\paragraph{The public rule.}
The international soil-moisture network that collects in-situ records for satellite validation publishes plausibility flags: values outside the physical range, values above a saturation point derived from a global soil database, rises without rainfall, spikes, breaks and plateaus\footnote{The flags are implemented openly: \url{https://flagit.readthedocs.io/}.}. Each flag is a threshold test, so the rule compiles in the same way as the derating rule.

\paragraph{Who holds the exceptions.}
Network operators hold site calibration, maintenance logs and sensor histories. A site's true saturation point, or a sensor known to read high in clay soil, is owner-specific knowledge, held per station and per sensor. A declaration such as ``this site saturates at this level'' is harmless to reveal.

\paragraph{Who could act as acceptance body.}
The network's data centre, or the quality-assessment service built on it, could re-derive declarations and test them against collocated satellite or reference measurements.

\paragraph{What would make it fail.}
The case for not pooling is weak. Operators already contribute data voluntarily under the network's terms of use\footnote{\url{https://ismn.bafg.de/terms-and-conditions}}, so a pooled correction, or simply a better flag, may serve them better than owned declarations. The mechanism's remaining argument here is attribution and exact withdrawal of station-specific corrections, not secrecy. That may not be worth the machinery.

\subsection{Space-weather anomaly screening}\label{sec:app-spaceweather}

\paragraph{The public rule.}
National space-weather services publish alert thresholds on the particle environment, for example on high-energy electron flux at geostationary orbit, above which deep-dielectric discharges on spacecraft have historically become more likely. An operator's screen of the form ``treat an anomaly as environment-caused, and act, when flux exceeds the public threshold'' is small and explicit.

\paragraph{Who holds the exceptions, and why they will not pool.}
Operators hold anomaly logs. Susceptibility depends on the bus design and its shielding, so the bus family is a natural subject, and anomalies are rare for any one operator. Public reports from a policy research organisation and from the United States weather agency to an international coordination group describe commercial operators as reluctant to share anomaly details, for reasons of customer confidence and proprietary exposure. We read those reports only in summary and do not cite them as evidence.

\paragraph{What a declaration would reveal, and who could accept it.}
A declaration would say that a bus family is susceptible above some level. That is vulnerability information, sensitive for security and for reputation. A national space-weather centre could check declarations against the environment record, and an operators' association acting as a trusted third party could hold the reference set; both are our guesses.

\paragraph{What would make it fail.}
The payload is the sensitive fact itself, which breaks the third condition, and operators' existing sharing arrangements favour anonymity over attribution. Of all the candidates, this is the closest structural twin of derating, and the one where an owned, signed declaration is least welcome.

\subsection{Other sectors, unassessed}\label{sec:app-other}

The following have a public rule and private exceptions. We have not assessed any of them beyond a first reading.
\begin{itemize}[nosep]
  \item \emph{Non-conformance and parts alerts.} Agencies run alert systems for failed parts and lots. A declaration here would be the alert itself, naming who failed, so it suits an anonymised owner, perhaps as the withdrawal and exclusion channel of the derating case.
  \item \emph{Nuclear commercial-grade dedication.} Utilities already share supplier-qualification evidence through joint audits that a regulator relies on. That is an institutional precedent for an acceptance body, not a data source.
  \item \emph{Index-based insurance triggers.} A public trigger on a satellite vegetation index decides payouts per area; insurers hold loss records showing where it misfires. Owners' books overlap by area, which is the conflict case the mechanism handles least well.
  \item \emph{Defensive cyber-control scoring} for space systems, \emph{avionics parts} and \emph{automotive qualification}: each has public tables and private evidence; in the last, data are plentiful per holder and pooled training may simply win.
\end{itemize}
Two candidates we examined fail outright. A shared ranker for forming bidding consortia works technically, but in the pilot a reader recovered every company's booked days and nothing about customers or rates: current capacity, the kind of exchange competition guidance lists among those treated as restrictions by object in earlier decisions~\cite{cma184}. Conjunction screening between satellite operators fails the first condition: the private knowledge is orbital state, not a rule, and secure computation already addresses it~\cite{hemenway-mpc-conjunction,veskus-sharesat}.

\section{The protocol around the mechanism}\label{sec:protocol}

The mechanism of Section~\ref{sec:mechanism} says what a declaration is, how it compiles and what the acceptance body checks. A deployment between competing organisations also needs to say who signed what, in what order things happened, how a contribution is retracted, and what a party may learn by asking. This section sets out those pieces. None of them is evaluated in this paper; they are requirements, and the last part says how far the company's existing systems go towards meeting them.

\paragraph{Declarations travel in signed envelopes.}
A submission is the declaration $d$, the party's gate statistic $T$ and a salted commitment $\mathit{com}$ to the records behind it (Algorithm~\ref{alg:accept}). The envelope adds the party's public key, a signature, the hash of the shared model the declaration was written against, and a short statement of intent in words. Weights are never part of the envelope: if a party attaches any, the acceptance body discards them. The envelope should be an existing, openly specified signed-event format that anyone can verify with standard libraries, not a new cryptographic format. We use a Nostr event~\cite{nip-01}: the party's identity is its secp256k1 public key, written as the decentralised identifier \texttt{did:nostr:} followed by the 64-character hexadecimal key~\cite{w3c-did-core,did-nostr-cg}, and the event carries a BIP-340 Schnorr signature over a hash of its contents~\cite{bip-340}. The identifier and the key are the same 32 bytes, so naming a party adds no key material and no new trust assumption. A signature binds a party to a claim. It does not show that the claim is true, that the stated learner produced it, or that the party's records stayed on its premises.

\paragraph{An append-only ledger of edits.}
On admission the acceptance body publishes $(d, p, \mathrm{hash}(\theta))$ and keeps $T$ and $\mathit{com}$ to itself (line~10 of Algorithm~\ref{alg:accept}). Each published entry also carries the hash of the previous entry, so the ledger is a hash chain: no entry can be altered or reordered without breaking every later link. Because admitted patches have disjoint support and commute, the current model is a pure function of the compiled prior and the live set. Anyone can therefore recompute it from the public rule and the ledger and compare hashes. The ledger is also what an engineering document can cite: in the derating case, the hash of the live set names exactly which exceptions an analysis relied on (Section~\ref{sec:app-derating}).

\paragraph{Exact withdrawal.}
A party, or the acceptance body after a dispute, withdraws a declaration by appending a withdrawal entry that names it. The model after withdrawal is the model the remaining declarations would have produced had the withdrawn one never been admitted, bit for bit, and no party retrains anything. Withdrawal does not erase what the declaration disclosed while it was live. Its values were published, and a ledger cannot recall them. Retraction of the record is exact; erasure of the disclosure is not available.

\paragraph{An acceptance body holding a reference set.}
The acceptance body re-derives every patch, runs the exhaustive checks and applies a reference test on labelled items it holds and no party chose (Section~\ref{sec:acceptance}). It sees a party's records only in a dispute, when the commitment is opened to it alone, and the number of records opened is reported as disclosure to the body. The reference set does not yet exist for any application. Building one is the hardest institutional step, because the regions that declarations change are by nature the ones where public labels are scarce.

\paragraph{Rationed probing.}
A reference set leaks if parties can query it freely. A party that submits many near-identical declarations and watches which are admitted learns the body's labels, just as parties in the pilot learnt each other's limits from votes. Votes are therefore not part of the acceptance test (Section~\ref{sec:fail-voting} gives the figures). The reference test needs a ration of its own: each party gets a fixed budget of submissions per period; a rejection says that the reference test failed, not on which items; and the budget a party spends is counted as disclosure of the body's labels to that party.

\paragraph{Canonical re-derivation removes hidden capacity.}
Because the acceptance body computes $\kap(d)$ itself from the canonical representative of $d$ (Section~\ref{sec:acceptance}), two parties who mean the same thing produce the same patch, and nothing about the way a patch was written can carry a message. Hidden capacity is then zero by construction (Appendix~\ref{app:proofs}). The comparison matters: in the pilot's ranking bench, counting showed that a free-form weight patch could hide at least thousands of bits in choices that no behavioural test sees (Section~\ref{sec:fail-hidden}). A deployment that accepted submitted weights, even with behavioural checks, would forgo this guarantee.

\subsection{What exists today}\label{sec:protocol-tiers}

DreamLab AI Consulting Ltd, where this work was done, builds two open systems that a deployment could stand on: VisionClaw\footnote{\url{https://github.com/DreamLab-AI/VisionClaw}}, a knowledge-graph platform with self-sovereign identity and an append-only provenance history, and agentbox\footnote{\url{https://github.com/DreamLab-AI/agentbox}}, a container for running agents under self-sovereign identities. Neither contains the compiler, the typed learner or the acceptance test, which exist only in the experimental code behind this paper. Table~\ref{tab:tiers} sorts the relevant components into three tiers. Where a configuration flag and a design record disagree about a component's status, we use the lower tier.

\begin{table}[t]
\centering
\small
\renewcommand{\arraystretch}{1.15}
\begin{tabular}{@{}p{3.4cm}p{11.6cm}@{}}
\toprule
Tier & Components relevant to the protocol \\
\midrule
Running today &
Self-sovereign identities as \texttt{did:nostr} identifiers over secp256k1 keys, with DID documents in the form of the Nostr Community Group's draft method~\cite{did-nostr-cg}, and BIP-340 Schnorr signatures from established, audited libraries; verification of signed HTTP requests (NIP-98 events~\cite{nip-98}) at the inbound boundary, failing closed; a per-identity data store with deny-by-default access control; an append-only, bi-temporal history of assertions that can be queried at any past time (its writes are not yet transactional, so a partial record is possible). \\
Built and tested, not switched on &
A signing service that offers a closed list of named operations and refuses generic signing; a library for salted-hash commitments with disclosure and revocation receipts signed as Nostr events of provisional kinds; a durable, crash-safe store for review cases from proposal to publication; an independent re-check of a proposed change by a pinned external checker that is not a build dependency of the system it checks; a timeline view that shows what was added and retracted between two times. \\
Designed only &
Separate \texttt{did:nostr} keys per party and for the acceptance body; a Nostr event kind for declarations and a hash-chained ledger of declarations, optionally anchored externally; a sealed reference set with rationed probing; a recorder, beyond the reach of the learner, of the model and prompt behind a declaration; a confined configuration in which a party's records provably stay on its premises; erasure of a disclosed value. \\
\bottomrule
\end{tabular}
\caption{The company's existing components, by maturity. Only the first row is in use. Recorded tests were found for the second row's components except the signing service, whose code was read but whose tests were not inspected. The mechanism itself (compiler, typed learner, acceptance test) is in neither system.}
\label{tab:tiers}
\end{table}

Two limits follow from the table. First, in today's default configuration of the agent container, agent context can leave the container through hosted model calls and session hooks, so nothing in it could yet support a claim that a party's records stay inside. That claim needs a local model, outbound traffic restricted to the model and the ledger, and no route by which the learner can rewrite its own provenance hooks. Second, signatures and hash chains establish who claimed what and that nothing changed afterwards. Re-derivation and the exhaustive checks are the only parts of the protocol that an outsider can verify deterministically, and they are the parts this paper tests.

\section{Experimental design}\label{sec:experiments}

This section describes the experiments as they were registered. Each analysis plan was written and hashed before the run it governs. The SHA-256 of the plan is written into every result row, and the plan is quoted by commit in the analysis that reports it. Table~\ref{tab:plans} lists every plan, what it fixed, and every departure from it. All accuracy evidence is synthetic (Section~\ref{sec:threat-synthetic}). The one bench on real records (Section~\ref{sec:exp-real}) tests compilation and the operations, not accuracy.

\subsection{The synthetic bench}\label{sec:exp-bench}

\paragraph{Items and the public rule.}
The bench is an invented qualification screen with three parties. An item is a part with four integer quantities (rated dose, required dose, rated cold limit, mission cold limit), a class (COTS for commercial off-the-shelf, automotive or space), a heritage flag and the party that holds it, encoded as eleven inputs. Write $s$ for rated dose minus required dose and $\mathrm{pm} = 2 - \text{heritage} + \text{COTS}$ for the public margin. The public rule accepts a part when $s \ge \mathrm{pm}$ and its rated cold limit is at least the mission cold limit. It is compiled exactly into the shared model (Section~\ref{sec:compiled-prior}). Each party has a fixed deployment set of 20{,}000 parts and an exhaustive grid of 124{,}416 parts. Accuracy is measured on the deployment set; spill is counted on the grids of the other parties.

\paragraph{Typed space.}
The narrow typed space $\Hyp$ has 127 declarations, ``no change'' included: a dose margin of $+1$ to $+8$, a cold margin of $+1$ to $+4$, or a cold threshold of $0$ to $8$, each on one of six scopes (all parts, COTS, automotive, space, no heritage, heritage). Every declaration tightens the rule. A free-form space of 1{,}113{,}751 linear declarations on the same six scopes is used as an ablation.

\paragraph{Truth classes.}
Hidden truths are sampled, not chosen, and the truth is the unit of inference. Each class has 20 truths, drawn from a stated distribution with a fixed seed stream. A sampled rule is admissible only if it changes between 2\% and 25\% of the party's deployment decisions relative to the public rule; otherwise it is redrawn from the same stream.
\begin{itemize}[nosep]
  \item \emph{Owner-specific, in the typed space} (\texttt{inspace-owner}, the primary population). Each party independently receives one non-empty declaration, uniform over the 126. Its truth is the public rule tightened by that declaration.
  \item \emph{Shared, in the typed space} (\texttt{inspace-shared}). One declaration, drawn the same way, applies to all three parties.
  \item \emph{Interaction, outside the typed space} (\texttt{heldout-interaction}). Each party rejects parts in a scope whose mission cold limit is at least $t$ and whose $s < \mathrm{pm} + d$, with $t \in \{2,\dots,6\}$ and $d \in \{1,\dots,5\}$. A rule that matches some declaration on the party's grid is redrawn.
  \item \emph{Relaxation, outside the typed space} (\texttt{heldout-lenient}). Each party loosens the rule on one scope, either by dose ($s \ge \mathrm{pm} - L$, $L \in \{1,2,3\}$) or by cold ($L \in \{1,2\}$). The typed space only tightens, so the best it can do is decline.
\end{itemize}
The two held-out families are reported with the same prominence as the in-space classes. The plans stated beforehand that the typed learner was expected to lose on relaxations and on shared truths, and to fall short of free-form or pooled learners on interactions at larger record counts.

\paragraph{Records, draws and seeds.}
Each truth has five independent record draws, so a cell holds 100 draws. A draw's accuracy for an arm is the mean over the three parties. A truth-level value is the mean over its five draws, and inference is across truths (19 degrees of freedom), never across draws. Records per party are the first $n$ of one seeded stream. Label noise is 0\% or 10\%; at 10\%, exactly $\lfloor n/10 + \tfrac12 \rfloor$ labels are flipped. Draw indices are disjoint across runs, so no run reuses data seen by another: 0--29 for the fixed-truth replication (Table~\ref{tab:app-rep}), 1000--1099 for confirmatory run~1, 2000--2099 for the gate development study, 3000--3099 for confirmatory run~2, 4000--4199 for the null check and 5000--5199 for the trained-base experiment. Truth streams are likewise disjoint (\texttt{0x7417}, \texttt{0x7418}, \texttt{0x7419}, \texttt{0x741A}, \texttt{0x4B1D}).

\paragraph{Grid of cells.}
Plan~v2 registered 70 cells. Before any cell beyond the primary had been analysed, the grid was reduced for time to nine (Table~\ref{tab:plans}), and plan~v3 registered the same nine:
\begin{itemize}[nosep]
  \item the primary cell, \texttt{inspace-owner} with 0\% noise and $n = 30$;
  \item \texttt{inspace-owner} at $n = 10$ and $n = 100$, and with 10\% noise at $n = 30$;
  \item \texttt{inspace-shared}, \texttt{heldout-interaction} and \texttt{heldout-lenient} at $n = 30$;
  \item two negative controls at $n = 30$: outcomes shuffled within each party's records, and each party learning from the next party's records.
\end{itemize}
The other 61 cells are deferred and listed in the analysis. No claim in this paper rests on them.

\subsection{Arms}\label{sec:exp-arms}

\paragraph{Primary arm.}
The typed learner (Algorithm~\ref{alg:learn}) searches the 127 declarations exhaustively on the party's own records. It breaks ties by fewest record errors, then by fewest of the party's own grid decisions changed from the public rule (the minimal-change tie-break, $\prec_{\mathrm{mc}}$), then by fewest slots, smallest coefficients and smallest offset. It then applies the evidence gate (Section~\ref{sec:gate}). The minimal-change tie-break was introduced after the first pilot and is therefore post hoc in origin. Plan~v2 made it primary before any v2 run, and the earlier pre-stated tie-break ($\prec_{\mathrm{pre}}$) is reported beside it.

\paragraph{The seven baselines.}
These are named exactly as in confirmatory run~2.
\begin{enumerate}[nosep]
  \item \emph{confined gradient}: gradient descent confined to the party's reserved unit (ten trainable values), with ``no change'' in its grid;
  \item \emph{Ditto pFL}: personalised federated learning, a FedAvg global model plus a proximal personal step per party;
  \item \emph{FedRep pFL}: a shared body averaged each round, with a personal head;
  \item \emph{FedAvg}: federated averaging;
  \item \emph{pooled 79-param net}: the shared network's architecture trained by SGD on all parties' records pooled;
  \item \emph{pooled large net}: a pooled network with one hidden layer of 32 or 128 units, given the eleven raw inputs plus three public-knowledge features ($s$, the cold slack, and the public decision);
  \item \emph{pooled gradient-boosted trees}: pooled boosted trees on fourteen integer features, the strong baseline that is not a neural network.
\end{enumerate}
Federated arms run ten rounds. The pooled arms break the setting's no-pooling constraint and are included as an accuracy reference only.

\paragraph{Ablations and references.}
The ablations are the same learner without the gate; the free-form linear space with and without the gate; and a learner with no compiled prior, which composes up to three declarations greedily from an absolute 151-element space, starting from ``accept everything''. The public rule is the reference and the truth is the ceiling.

\paragraph{The language-model arm.}
The language-model arm is not part of the registered grid and is reported separately (Section~\ref{sec:fail-llm}). In the pilot, a frontier model, given a party's records and the public rule but no tools, decided 60 held-out boundary parts per party directly. Its stated rule was also written out and scored on every valid input. Local open-weight models were run as readers of prose test reports, as authors of free-form weight edits, and as the model-alone arm on one prompt.

\paragraph{Honest tuning, and oracle tuning as sensitivity only.}
Every baseline is tuned only on data it is allowed to see. Per-party arms use 5-fold cross-validation (CV) on the party's own records, with record $i$ in fold $i \bmod 5$. Federated arms use federated cross-validation, and pooled arms use CV on the pooled records. Every grid includes ``no change'' first. The CV-tuned figures are the ones tested. ``Oracle'' variants choose from the same grid using the held-out deployment truth, which no real party has. They appear only in sensitivity tables, as upper bounds on what better tuning could achieve, and never in a test.

\subsection{Metrics}\label{sec:exp-metrics}

\begin{itemize}[nosep]
  \item \emph{Accuracy}: the share of a party's 20{,}000 deployment parts decided as the truth decides them, averaged over parties. Two error rates are reported with it: \emph{bad accepted}, the share of truth-rejected parts that are accepted, and \emph{good rejected}, the share of truth-accepted parts that are rejected.
  \item \emph{Spill}: the number of decisions changed on the other parties' exhaustive grids. For typed and confined arms it is measured with the compiled weight patches. For federated and pooled arms it is a leave-one-party-out retrain of the CV-chosen configuration.
  \item \emph{Exact withdrawal}: compose every party's patch, invert one, and compare model hashes. Federated and pooled arms have no withdrawal operation and need retraining.
  \item \emph{Values disclosed and disclosure bound}. A typed party discloses one element of $\Hyp$, so its bound is $\log_2 |\Hyp|$ whether or not it proposes. Other arms are bounded by $\min(64 \times \text{values sent},\, n \times c)$, under two record caps $c$: the log of the number of encodable records, and the generator's exact entropy per record. The ratio of two upper bounds is not a lower bound on the true ratio (Remark~\ref{rem:disclosure-not}).
  \item \emph{Gate statistics}: the proposal rate per party-draw; proposals less accurate than the public rule on that party's deployment set; and exact recovery of the sampled declaration.
\end{itemize}

\subsection{Decision rule}\label{sec:exp-decision}

\paragraph{Primary endpoint.}
The primary endpoint is set at the primary cell. For each CV-tuned baseline $B$, let $D_B$ be the truth-level paired difference (typed learner minus $B$). Non-inferiority to $B$ holds if the lower end of the two-sided 95\% $t$-interval of the mean of $D_B$ (df $= 19$) lies above $-2$ points. The endpoint requires non-inferiority to all seven baselines. This is an intersection--union test: ``no worse than the best honestly tuned baseline'' is tested without choosing the best baseline from the data, and it needs no multiplicity correction. Plan~v2 added a second clause: the disclosure-bound ratio, Ditto pFL over the typed learner, must be at least 10 in every draw. Plan~v3 dropped that clause from its endpoint (Table~\ref{tab:plans}).

\paragraph{Secondary family.}
The secondary comparisons are tested together with Holm correction at $\alpha = 0.05$, all two-sided:
\begin{itemize}[nosep]
  \item S1--S7: superiority over each baseline;
  \item S8: the typed learner minus the public rule;
  \item S9: gated minus ungated;
  \item S10: minimal-change minus pre-stated tie-break;
  \item S11 (plan~v3 only): tuned gate minus judgement gate.
\end{itemize}
The Holm step uses a two-sided sign-flip randomisation test (200{,}000 resamples). A Wilcoxon signed-rank test is reported as a check. Each comparison also reports a two-stage bootstrap interval (resampling truths, then draws within truths), the minimum detectable difference at 80\% power, and the number of truths won, lost and tied. Rates carry Wilson intervals. These are labelled as clustered, because party-draws within a truth are not independent.

\paragraph{Falsification.}
The hypothesis is falsified at the primary cell if the endpoint fails, if any typed patch spills in any draw, or if any typed withdrawal is inexact. Spill and withdrawal are acceptance checks guaranteed by construction (Section~\ref{sec:guarantees-vs-hypothesis}). They are reported as counts, not as tests.

\subsection{The evidence gate: development, then confirmation}\label{sec:exp-gate}

Plans~v1 and~v2 set the gate by judgement: a declaration must fix at least three more records than it breaks ($m \ge 3$), and the selection-aware Monte-Carlo $p$ must be at most 0.05 (199 replicates of the full search on records whose public-rule decisions are flipped at the public rule's own error rate). In run~1 this setting cost accuracy on every truth, so the gate was tuned. The tuning had to avoid using the confirmatory data.

\paragraph{Development study.}
The gate tuning plan was committed before the sweep ran. It fixed sixteen settings, $m \in \{1,2,3,4\}$ crossed with $p \in \{0.05, 0.1, 0.2, \text{none}\}$, on fresh truths and draws used for nothing else. The selection criterion was fixed in advance:
\begin{enumerate}[nosep]
  \item A setting is \emph{feasible} if its false-proposal rate on a public-rule null (the public rule's own decisions as labels, with 10\% flipped at random) is at most 5\%, and its harmful proposals at $n = 10$ are at most 5\% of party-draws.
  \item Among feasible settings within a quarter of a point of the best accuracy at $n = 30$, choose the strictest.
\end{enumerate}
The held-out families were reported but did not enter the criterion. The chosen setting, $m \ge 1$ and $p \le 0.2$, then went into plan~v3 and was confirmed on fresh truths and draws.

\paragraph{Null check.}
After run~2, an exploratory check was run on unused draws: the tuned gate on a fresh public-rule null. Its note was hashed before it ran, but it was written after unblinding and is not pre-registered.

\subsection{The trained-base experiment}\label{sec:exp-hybrid}

A separate plan, committed before any comparison, tested declarations laid over a \emph{trained} base rather than the compiled public rule (Section~\ref{sec:trained-base}). It used 40 truths. Each truth combines a shared component outside the typed space, which is an interaction or a relaxation applied to all parties, with an owner component, which is one typed declaration per party. The base is a 17-unit network. It is trained on shared records labelled by the shared component only, while three reserved units are held at exactly zero. Each party's gated declaration compiles into its own reserved unit.

The primary cell is 30 private records per party and 300 shared records. The endpoint is an intersection--union test of two superiorities: the combination over the trained base alone, and the combination over typed declarations on the public rule. Each is tested across the 40 truths: the lower end of the 95\% $t$-interval (df 39) must lie above zero. Alongside them, exact non-interference (checked on output values, not only decisions), fidelity and exact withdrawal must hold in every draw.

Personalised FL (Ditto pFL), per-party fine-tuning of the base and pooled training on the same network form the secondary family. After unblinding, an exploratory sensitivity check added stronger comparators, at the operator's request:
\begin{itemize}[nosep]
  \item a better-trained base;
  \item a pooled 128-unit network;
  \item boosted trees, pooled and per party;
  \item oracle-tuned fine-tuning and Ditto pFL, from both bases.
\end{itemize}

\subsection{The real-record bench}\label{sec:exp-real}

The real-record bench compiles one published derating line into a ReLU network: ECSS-Q-ST-30-11C Rev.~2, Table~6-36, drain--source voltage at most 80\% of rated. Declarations are read mechanically from public single-event-effect test reports for power transistors, in NASA compendia from 2014 to 2025. The three parties are year bands of the compendia. Each declaration sets a voltage limit for one part over a range of gate bias, and it can only tighten the published line. The bench checks exact compilation, composition, non-interference and withdrawal on every valid input. It has no accuracy evaluation, because no held-out truth exists for real parts. Section~\ref{sec:res-real} gives a worked example.

\subsection{Plans and deviations}\label{sec:exp-plans}

Table~\ref{tab:plans} lists every plan. Commits are in the experimental repository, and the order of commits was checked from its history. A plan described as ``committed before its run'' has a commit earlier than the commit of its result rows and of any analysis of them.

\begin{table}[p]
\centering
\caption{Every analysis plan, what it fixed and every departure from it. ``Pre-registered'' means committed (or, for v1, hashed) before the run it governs. The exploratory analyses at the foot were written after the registered results had been seen.}
\label{tab:plans}
\footnotesize
\begin{tabular}{@{}p{1.9cm}p{1.5cm}p{3.7cm}p{7.6cm}@{}}
\toprule
Plan & Commit & What it fixed & Deviations, as recorded \\
\midrule
v1: fixed-truth replication (exploratory) & hashed \texttt{0edf77ae}; first committed in \texttt{86bb5fb} & Three fixed truths; 30 draws; seven record counts; pre-stated tie-break primary; judgement gate. & Hashed but not committed before its run, and draws 0--9 had been seen by an earlier pilot, so all of it is exploratory. Four departures: (1) an ungated no-prior arm was added, because the gate's null degenerates when the reference rule is worse than a constant rule; (2) the confined-gradient patch was checked under a 16-edit limit; (3) federated values counted changed floats rather than transmitted floats, which undercounts; (4) the pooled large net's step size, which the plan left open, was fixed before the run. \\
v2: confirmatory run 1 & \texttt{abfc9c8} & 20 sampled truths per class; truth as the unit of inference; minimal-change tie-break primary; seven CV-tuned baselines; endpoint of disclosure ratio $\ge 10$ plus accuracy non-inferiority; judgement gate unchanged. & (1) \emph{The endpoint failed as first run, in one draw}. The code counted the federated floats a client \emph{changed}, not the floats it \emph{transmitted} as the plan defines. In one draw every record already agreed with the public rule with margin, so the federated updates were exactly zero and Ditto pFL's bound was counted as 0 bits. The disclosure clause therefore failed in that draw, and the endpoint as first run fails. With uploads counted as the plan defines them and the cell re-run, the endpoint passes; accuracy and spill are byte-identical, and the as-run rows are kept. (2) \emph{The grid was reduced} from 70 cells to 9 before any non-primary cell was analysed. (3) The boosted-tree leaf regularisation, which the plan left open, was set to 1. \\
Gate tuning (development) & \texttt{3a52b7e} & Sixteen gate settings on fresh data; feasibility bounds; selection rule fixed before the sweep. & None in the selection. The chosen setting was selected on development data, so its false-proposal rate there is biased low. A column for wrong-party proposals was added to the report after the sweep, computed from the committed rows; the criterion was unchanged. \\
v3: confirmatory run 2 & \texttt{78c9304} & Tuned gate ($m \ge 1$, $p \le 0.2$); fresh truths and draws; accuracy-only endpoint; S11; the nine-cell grid. & None in the design, gate, data or grid. The disclosure clause was removed from the endpoint: once uploads are counted as transmitted it is close to automatic, and in v2 it had added brittleness without evidence. The analysis routines for v3 were written after the plan and before any v3 row existed; the reconstruction of the judgement gate was validated on v2's rows first. \\
Trained base (hybrid) & \texttt{8a5f6a5} & 40 truths with shared and owner components; endpoint of two superiorities plus guarantees; secondary family S1--S6. & One confined-gradient row exceeded the edit limit and is marked partial (slot ablation only). Pooled training was expected to be an upper bound and was not. \\
\midrule
Null check (exploratory) & \texttt{6a38b4f} & The tuned gate on a fresh public-rule null, at three record counts. & Not pre-registered: written after run 2. Its pooled false-proposal rate's interval reaches above the 5\% bound, and the rate itself exceeds the bound at 100 records (Section~\ref{sec:res-gate}). \\
Trained-base sensitivity (exploratory) & \texttt{7352944} & Stronger and oracle-tuned comparators on the registered truths and draws. & Post-unblinding, at the operator's request; plan and rows were committed together. \\
Loop pilot (exploratory) & \texttt{7352944} & A local model in a propose--check--revise loop on held-out truths. & Plan and results were committed together. Only one truth at one record count ran (Section~\ref{sec:fail-loop}). \\
\bottomrule
\end{tabular}
\end{table}

\section{Results}\label{sec:results}

The results come in the order of the registered claims: the confirmatory runs, then the acceptance checks, disclosure, the gate, the real-record bench and the trained base. Section~\ref{sec:failures} reports, with the same weight, every condition in which the mechanism lost, tied or failed. Appendix~\ref{app:tables} gives every cell and every arm with intervals.

\subsection{Confirmatory run 2: the primary endpoint}\label{sec:res-primary}

\begin{figure}[tbp]
\centering
\includegraphics[width=0.85\linewidth]{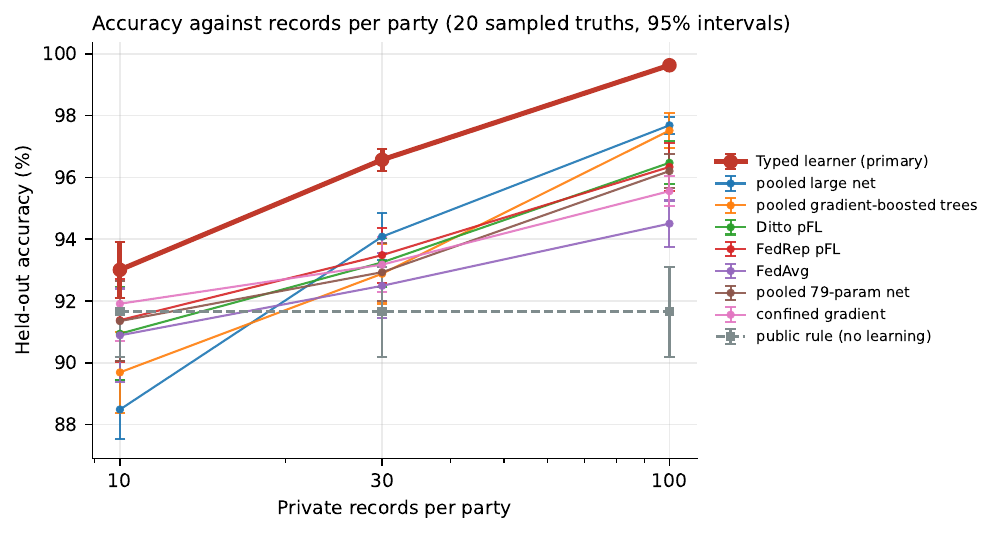}
\caption{Accuracy against the number of private records per party, confirmatory run 2. Owner-specific truths in the typed space, no label noise. Each point is the mean over 20 sampled truths, each truth averaged over five record draws. Bars are 95\% $t$-intervals across truths. Red: the typed learner (minimal-change tie-break, tuned gate). Dashed grey: the public rule, which learns nothing. The other lines are the seven CV-tuned baselines. Only the 30-record cell is confirmatory; 10 and 100 records are exploratory. The truths here have the shape the typed space was built to hold (Section~\ref{sec:threat-tailored}).}
\label{fig:accuracy}
\end{figure}

At the primary cell (owner-specific truths in the typed space, no label noise, 30 records per party), the typed learner with the tuned gate was non-inferior to all seven CV-tuned baselines. Figure~\ref{fig:accuracy} shows accuracy against the number of records for the typed learner, the seven baselines and the public rule; only the 30-record cell is confirmatory. The primary endpoint passes, and the hypothesis is not falsified at this cell. Table~\ref{tab:primary} gives the seven paired differences. Each lower bound lies above zero, not only above the margin of $-2$ points. Each superiority comparison, S1 to S7, has a Holm-adjusted $p$ below 0.05 (Table~\ref{tab:app-secondary}); the comparison with the pooled large net has $p$ <0.0001. The typed learner gained \ensuremath{+4.9} points [\ensuremath{+3.5}, \ensuremath{+6.3}] over the public rule (S8). Oracle tuning, which chooses each baseline's configuration on the deployment truth, is reported only as a sensitivity check. Even so tuned, the best baseline, Ditto pFL, reached 95.6\% and the pooled large net 94.8\%, against 96.6\% for the typed learner (Table~\ref{tab:app-oracle-ev3}). These are point estimates; no oracle arm enters a test. In run~1 the oracle ordering was reversed: oracle-tuned Ditto pFL reached 95.9\%, the pooled large net 95.5\% and FedRep pFL 95.3\%, against 95.2\% for the typed learner (Table~\ref{tab:app-oracle-ev}). The two runs differ in both the gate setting and the sample of truths, so the run-2 oracle ordering should not be read as robust.

\begin{table}[htbp]
\centering
\caption{Primary endpoint, confirmatory run 2 (plan v3). The typed learner (minimal-change tie-break, tuned gate) minus each CV-tuned baseline, at 30 records per party, owner-specific truths, no label noise. Mean of 20 truth-level paired differences, with the 95\% $t$-interval (df 19). Non-inferiority requires every lower bound to exceed $-2$ points.}
\label{tab:primary}
\small
\begin{tabular}{@{}lll@{}}
\toprule
Baseline & Difference & 95\% interval \\
\midrule
confined gradient & \ensuremath{+3.4} & [\ensuremath{+2.5}, \ensuremath{+4.3}] \\
Ditto pFL & \ensuremath{+3.3} & [\ensuremath{+2.6}, \ensuremath{+4.1}] \\
FedRep pFL & \ensuremath{+3.1} & [\ensuremath{+2.2}, \ensuremath{+4.0}] \\
FedAvg & \ensuremath{+4.1} & [\ensuremath{+3.1}, \ensuremath{+5.1}] \\
pooled 79-param net & \ensuremath{+3.6} & [\ensuremath{+2.7}, \ensuremath{+4.5}] \\
pooled large net & \ensuremath{+2.5} & [\ensuremath{+1.7}, \ensuremath{+3.3}] \\
pooled gradient-boosted trees & \ensuremath{+3.7} & [\ensuremath{+2.8}, \ensuremath{+4.6}] \\
\bottomrule
\end{tabular}
\end{table}

\begin{figure}[tbp]
\centering
\includegraphics[width=\linewidth]{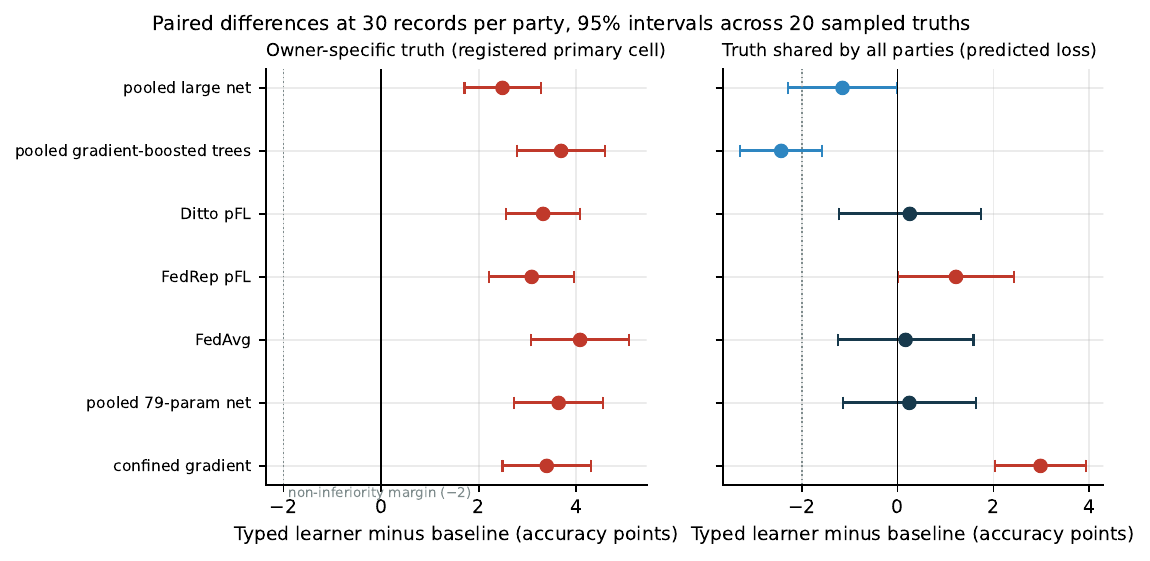}
\caption{Paired differences at 30 records per party, confirmatory run 2: typed learner minus each CV-tuned baseline, in accuracy points, with 95\% intervals across 20 sampled truths. Left: owner-specific truths, the registered primary cell. Right: one truth shared by all parties, where pooling triples the relevant data and the plan predicted a loss. Red: the typed learner is ahead. Blue: the baseline is ahead. Dark: no difference shown. Dotted line: the non-inferiority margin of $-2$ points.}
\label{fig:paired}
\end{figure}

\paragraph{Run 1, and how much of run 2's margin comes from the sampled truths.}
Confirmatory run 1 (plan v2, judgement gate, a different sample of 20 truths) also passed its accuracy endpoint, after the counting correction described in Table~\ref{tab:plans}. Its margins were smaller (Table~\ref{tab:app-arms-v2}, with the differences in Table~\ref{tab:app-diff-v2}). After Holm correction it was \emph{not shown superior} to FedRep pFL (Holm $p$ 0.0618; \ensuremath{+1.4} points [\ensuremath{+0.1}, \ensuremath{+2.7}]) or to the pooled large net (Holm $p$ 0.2620; \ensuremath{+0.4} points [\ensuremath{-0.3}, \ensuremath{+1.1}]). Against the pooled large net, run 1 was level.

Run 2 differs from run 1 in two ways: the gate setting and the truth sample. The design lets the two be separated, because the judgement gate can be reconstructed exactly on run 2's draws from each party's recorded evidence. The tuned gate was \ensuremath{+1.3} [\ensuremath{+0.9}, \ensuremath{+1.6}] points more accurate than the judgement gate on the same draws (S11). Run 2's margin over the pooled large net was \ensuremath{+2.5} points. With the judgement gate on the same draws it would have been \ensuremath{+1.2} points [\ensuremath{+0.4}, \ensuremath{+2.0}], and against FedRep pFL \ensuremath{+1.8} points [\ensuremath{+0.9}, \ensuremath{+2.7}]. \textbf{So about half of run 2's margin over the pooled network comes from the sample of rules; the other half comes from the tuned gate.} Even with the judgement gate, the run-2 truths separate the typed learner from both baselines that run 1 could not separate. This exploratory decomposition is not part of either registered test.

\subsection{Acceptance checks}\label{sec:res-guarantees}

Non-interference and exact withdrawal are guaranteed by construction (Section~\ref{sec:properties}) and checked in every draw. In run 2, typed patches spilled in 0 draws across all nine cells, and every withdrawal restored the model hash exactly. The confined gradient, which writes only its own reserved unit, spilled in 0 draws.

Every federated and pooled baseline changed other parties' decisions in most draws. In run 2 the counts of draws with spill were:
\begin{itemize}[nosep]
  \item Ditto pFL, 554;
  \item FedRep pFL, 498;
  \item FedAvg, 490;
  \item pooled 79-param net, 515;
  \item pooled large net, 699;
  \item pooled gradient-boosted trees, 633.
\end{itemize}
None of these baselines can withdraw a party without retraining.

\subsection{Disclosure}\label{sec:res-disclosure}

\begin{figure}[tbp]
\centering
\includegraphics[width=0.75\linewidth]{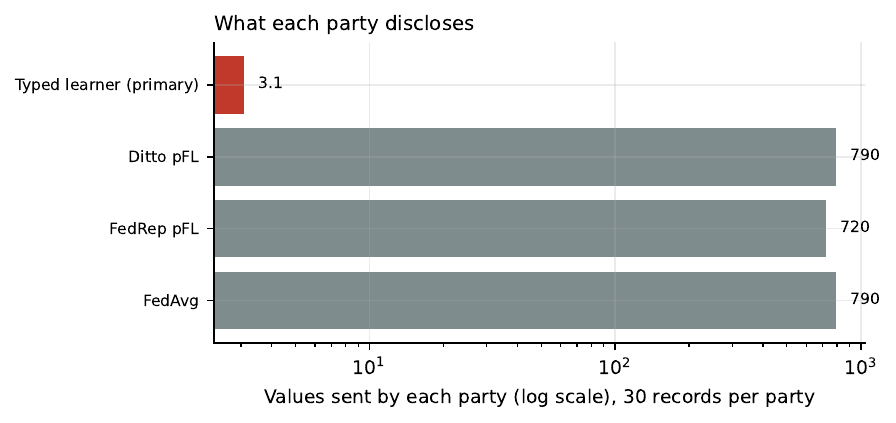}
\caption{Values each party sends at 30 records per party, confirmatory run 2, on a log scale: the typed learner's declaration slots against the parameters uploaded by the federated baselines in their final training run. Uploads made during federated cross-validation are excluded, which favours the federated arms. The information bound is the smaller of $64$ bits per value and a cap on the information in the records themselves. A value count is not an information measure (Section~\ref{sec:res-disclosure}).}
\label{fig:disclosure}
\end{figure}

A typed party discloses one element of a 127-element space, so at most $\log_2 127 \approx 7$ bits per declaration whether it proposes or not. Figure~\ref{fig:disclosure} shows the values each party sends. At 30 records, the smallest ratio over draws of Ditto pFL's disclosure bound to the typed learner's was 76.9$\times$ under the encoding cap and 63.6$\times$ under the generator-entropy cap. At 10 records it was 25.6$\times$, and at 100 records 256.5$\times$. These are ratios of two upper bounds. They carry no sampling error and are not lower bounds on the true ratio, since a federated upload may reveal less than its bound (Remark~\ref{rem:disclosure-not}).

\subsection{The evidence gate}\label{sec:res-gate}

In neither confirmatory run, nor over the trained base, did the gate raise accuracy at the reported precision. The judgement setting used in run 1 cost accuracy. In run 1, gated minus ungated was \ensuremath{-1.6} points [\ensuremath{-1.9}, \ensuremath{-1.2}]. The tuned setting removed most of that cost: in run 2, gated minus ungated was \ensuremath{-0.1} points [\ensuremath{-0.1}, \ensuremath{0.0}] (S9). In every other cell of both runs the ungated learner was level with the gated one or ahead of it at the printed precision (Tables~\ref{tab:app-arms-v3} and~\ref{tab:app-arms-v2}). Interaction truths in run~2 are the only cell in which the gated learner was ahead before rounding; both print as 95.3\% gated and 95.3\% ungated. Over the trained base the gate cost accuracy in a registered comparison (Section~\ref{sec:res-hybrid}).

What the gate buys is fewer harmful proposals. In no cell of run~2 was a larger share of its proposals worse than the public rule than without it; with 10\% label noise that share was 17.8\% gated against 28.0\% ungated, and on interaction truths 19.9\% against 26.1\% (Table~\ref{tab:app-gate}).

\begin{figure}[tbp]
\centering
\includegraphics[width=0.75\linewidth]{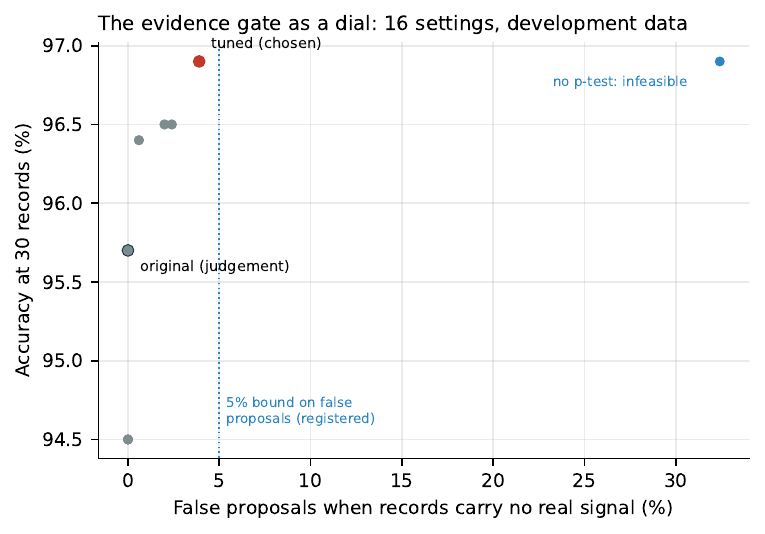}
\caption{The evidence gate as a dial, on development data used for nothing else. Each point is one of 16 settings (minimum net evidence $m \in \{1,\dots,4\}$ crossed with the selection-aware $p$ threshold in $\{0.05, 0.1, 0.2, \text{none}\}$). Horizontal axis: the false-proposal rate when the public rule is correct and record labels are 10\% noise. Vertical axis: accuracy at 30 records on owner-specific truths. The selection rule, fixed before the sweep, chose the strictest feasible setting within a quarter of a point of the best accuracy: the ``tuned'' point. The ``original'' point is the judgement setting of runs v1 and v2. With a minimum net evidence of one, dropping the $p$ test breaks the 5\% bound. On fresh data the tuned setting's false-proposal rate exceeds the bound at 100 records (Section~\ref{sec:res-gate}).}
\label{fig:gate}
\end{figure}

The gate is a dial rather than a fixed component, and Figure~\ref{fig:gate} shows its sixteen settings on the development data. Compared with the judgement setting, the tuned setting proposes more often, and that is where its accuracy gain comes from. On the development data its false-proposal rate on the public-rule null was 3.9\%, against 0.0\% for the judgement setting. That figure is biased low, because the setting was chosen on the same data.

On fresh data (the exploratory null check), the tuned gate's pooled false-proposal rate was 4.3\% [3.5\%, 5.4\%]. Its interval reaches above the 5\% bound. The rate rose with the number of records:
\begin{itemize}[nosep]
  \item 10 records: 3.0\%;
  \item 30 records: 3.8\%;
  \item 100 records: 6.2\%, which \textbf{exceeds the bound}.
\end{itemize}
The judgement gate stayed below 1\% at every record count (0.2\% pooled). The tuned setting is best read as a small-sample setting. With more records, small chance improvements clear a one-record margin more often.

\subsection{A worked example on real records}\label{sec:res-real}

\begin{figure}[tbp]
\centering
\includegraphics[width=\linewidth]{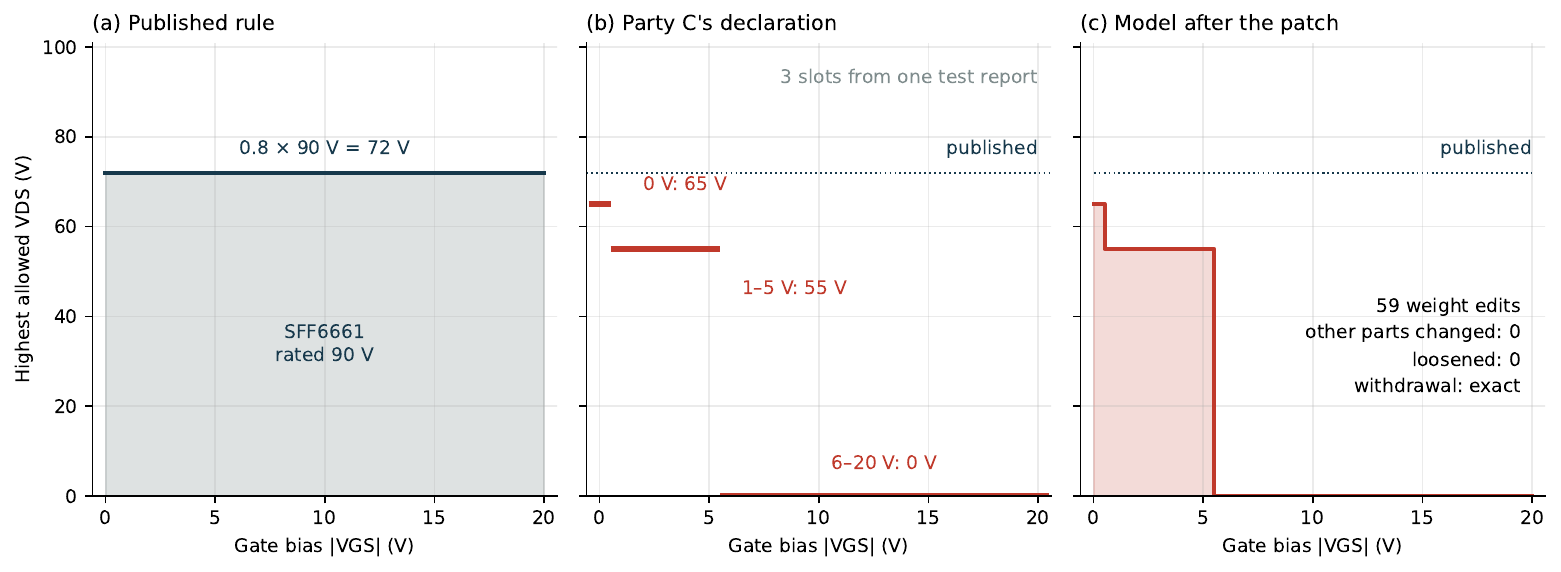}
\caption{From a published derating rule to one declaration to the compiled edit, on public records. (a) The published rule (ECSS-Q-ST-30-11C Rev.~2, Table~6-36: drain--source voltage at most 80\% of rated), compiled exactly into the shared model, for one transistor rated at 90~V; grey is accepted. (b) Party C's declaration, read mechanically from one public single-event test report on one lot. Each slot sets the highest voltage at which the lot survived, over a range of gate bias. (c) The shared model after the declaration's weight patch: the published rule tightened by the declaration, for this part only. The patch's edit count, the number of other parts' decisions it changed (zero) and the number of decisions it loosened (zero) are read from the run log. Withdrawal restores the model exactly. The tested range ends at 10~V of bias; reading the region above it as zero is a policy choice for the acceptance body (Section~\ref{sec:fail-real}).}
\label{fig:worked}
\end{figure}

Figure~\ref{fig:worked} follows one public test report through the mechanism. The published line, drain--source voltage at most 80\% of rated, compiles into a network of 21,213 parameters. That network agrees with the line on all 32,975,040 valid inputs, with 0 disagreements.

One report on one transistor lot shows destructive failures below the published limit, at a gate bias that depends on the test conditions. Read mechanically, it becomes a declaration with three slots. The first gives a limit of 65 V at zero gate bias, against the published 72 V. The others give lower limits at larger bias. The declaration compiles to a weight patch that tightens the decision for that part alone.

Across the bench, 15 of 30 report rows compiled, using 35 slots and patches of 19 to 82 edits. Composition gave the same model in either order. The composed model disagreed with the plain rule (the ratio plus every declared limit) on 0 inputs. A blind language-model reader matched every declared limit within the tested range on 15 of 15 rows. This bench shows that the published rule compiles exactly and that the operations behave as proved. It shows nothing about accuracy (Section~\ref{sec:threat-real}).

\subsection{Declarations over a trained base}\label{sec:res-hybrid}

The trained-base experiment tests the setting the mechanism is meant for. Part of the shared model is learned from shared records, and each party's private exception is laid over it as a typed weight edit.

\paragraph{The registered result.}
At 30 private and 300 shared records per party, the primary endpoint passed. The combination beat both of its parts:
\begin{itemize}[nosep]
  \item the trained base alone, by \ensuremath{+5.4} points;
  \item typed declarations over the compiled public rule, by \ensuremath{+3.1} points.
\end{itemize}
Both lower bounds lay above zero (Table~\ref{tab:app-hybrid}); Table~\ref{tab:app-hybrid-cells} gives the registered comparisons in every cell. In every draw of every cell, the combination changed none of the other parties' outputs (0), had 0 fidelity mismatches, and withdrew exactly in 1,800 rows. Each party disclosed 2.2 values on average, against 991.9 for Ditto pFL.

In the secondary family, the combination was more accurate than the registered Ditto pFL (\ensuremath{+1.9}), private fine-tuning (\ensuremath{+1.9}) and pooled training (\ensuremath{+4.8}). All three share one small network and trainer, and that is a weakness of the registered design. The registered comparison with the same combination without its gate (S5 of this plan) went the other way: gated minus ungated was \ensuremath{-1.3} points [\ensuremath{-1.5}, \ensuremath{-1.0}], significant after Holm correction, with the ungated combination at 95.6\% against 94.3\% (Table~\ref{tab:app-hybrid}). As in the confirmatory runs, the gate did not raise accuracy here; Section~\ref{sec:res-gate} reports what it buys.

\begin{figure}[tbp]
\centering
\includegraphics[width=\linewidth]{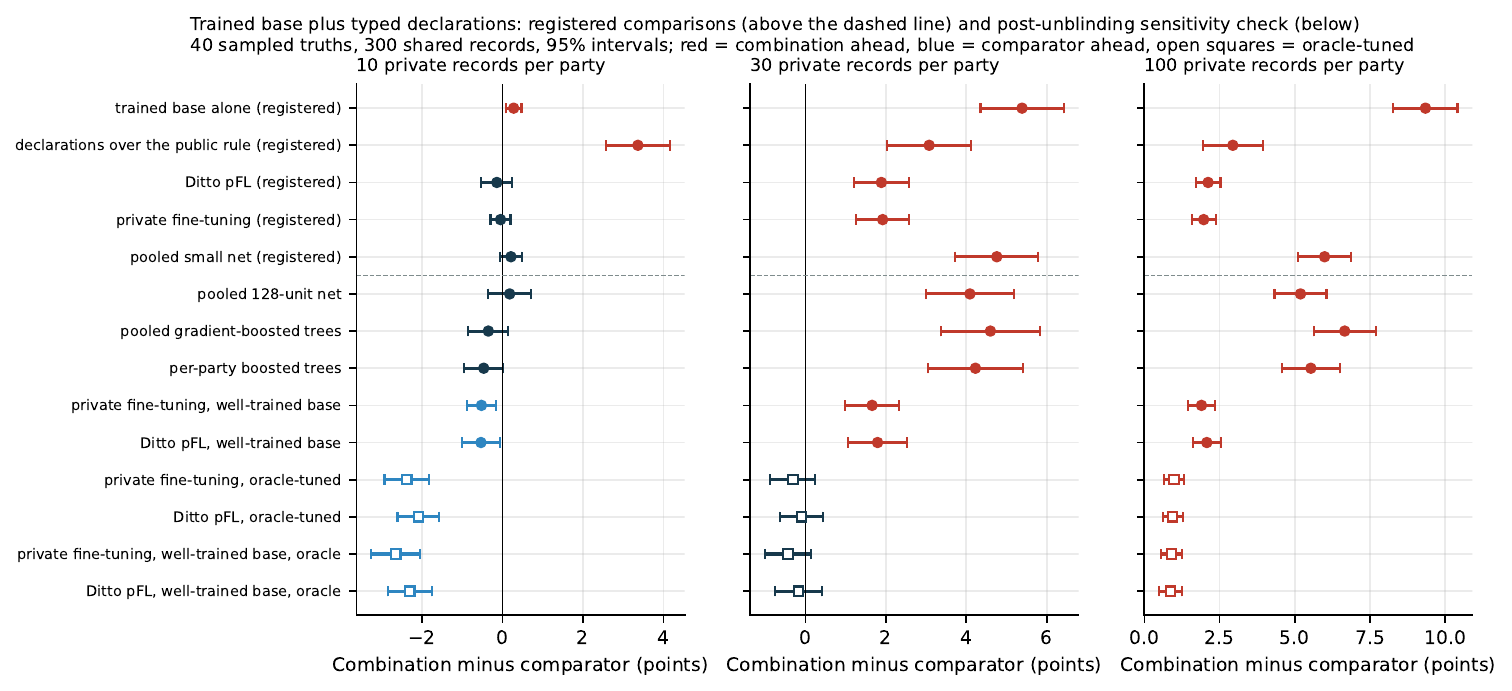}
\caption{Typed declarations over a trained base: the combination minus each comparator, in accuracy points, with 95\% intervals across 40 sampled truths (interaction and relaxation families), 300 shared records, and 10, 30 or 100 private records per party. Above the dashed line: the registered comparisons (plan committed before any comparison). Below: the post-unblinding sensitivity check with stronger comparators. Open squares are oracle-tuned, an advantage no real party has, so they bound what better tuning could achieve. Red: the combination is ahead. Blue: the comparator is ahead. Dark: no difference shown. At 30 records, oracle-tuned private fine-tuning and Ditto pFL match the combination; at 10 records, fine-tuning and Ditto pFL from a better-trained base (labelled ``well-trained base'' in the figure) beat it even when CV-tuned.}
\label{fig:trained-base}
\end{figure}

\paragraph{The sensitivity check.}
The post-unblinding check (Figure~\ref{fig:trained-base}) shows the registered base was competent. It scored 98.3\% against the shared component alone, and a better-trained base scored 98.6\%. So the registered margin over the base is not an artefact of underfitting.

At 30 records, no CV-tuned comparator matched the combination. That includes fine-tuning and Ditto pFL from the better base, a 128-unit pooled network, and boosted trees, pooled or per party. \textbf{Well-tuned private fine-tuning, however, matches the trained-base combination at 30 records.} With its configuration chosen on deployment truth (oracle tuning), the combination minus private fine-tuning was \ensuremath{-0.3} points [\ensuremath{-0.9}, \ensuremath{+0.2}]. Oracle-tuned Ditto pFL also matched it (\ensuremath{-0.1} [\ensuremath{-0.6}, \ensuremath{+0.4}]; Table~\ref{tab:app-hybrid-sens}). Against these, the combination's accuracy advantage at the primary cell is not established. What remains is structural: one shared model, exact non-interference, exact withdrawal, and a few disclosed values per party rather than hundreds.

At 10 records the combination lost (Section~\ref{sec:fail-ten}). At 100 records it beat every comparator, oracle-tuned ones included; the only exception is its own variant over the better base. For example, against oracle-tuned Ditto pFL from the better base the difference was \ensuremath{+0.9} [\ensuremath{+0.5}, \ensuremath{+1.2}] (Table~\ref{tab:app-hybrid-sens}).

\section{Where it fails}\label{sec:failures}

This section carries the same weight as Section~\ref{sec:results}. It lists every condition in which the typed learner lost, tied, proposed wrongly or could not be tested, together with the mechanisms the evidence argues against. Most of these failures were predicted in the plans. The rest are stated as found. Section~\ref{sec:threats} addresses the objections that follow from them.

\subsection{Truths shared by all parties}\label{sec:fail-shared}

When one exception applies to every party, pooling triples the relevant data, and pooled learners win. In run 2 at 30 records, the typed learner was behind two baselines (typed learner minus baseline, in points):
\begin{itemize}[nosep]
  \item the pooled large net, \ensuremath{-1.2} [\ensuremath{-2.3}, \ensuremath{0.0}];
  \item pooled gradient-boosted trees, \ensuremath{-2.4} [\ensuremath{-3.3}, \ensuremath{-1.6}].
\end{itemize}
Figure~\ref{fig:paired}, right, shows both. Run 1 lost to the same two baselines (Table~\ref{tab:app-diff-v2}). The mechanism makes each party learn alone from its own records. When the truth is common, that discards two-thirds of the evidence. A shared exception belongs in the public rule, or in a trained base, not in per-party declarations.

\subsection{Truths that relax the rule}\label{sec:fail-lenient}

Every declaration tightens the rule. When a party's truth is more lenient than the published rule, the typed learner can only decline. In run 2 it proposed in 0.0\% of party-draws, and its accuracy equalled the public rule's (93.5\% against 93.5\%). FedRep pFL, which can loosen, was ahead: the typed learner minus FedRep pFL was \ensuremath{-1.6} [\ensuremath{-2.6}, \ensuremath{-0.7}].

Declining is the safe failure, because nothing is made worse. But a community that adopts the mechanism gives up learning leniency from records. This is deliberate in the qualification setting: the min() in the published derating rule discards any limit above it. The real-record bench found test reports where survival lay above 80\% of rated, and the rule could not use them (Section~\ref{sec:threat-real}).

On interaction truths, which no declaration can express, the typed learner was still ahead of every CV-tuned baseline at 30 records (Table~\ref{tab:app-diff-v3}). It got there by declaring the nearest expressible tightening, and 19.9\% of its proposals were less accurate than the public rule on the party's own deployment set (Table~\ref{tab:app-gate}). The plans expected free-form and pooled learners to overtake it at larger record counts. Those cells were deferred and remain untested.

\subsection{Ten records}\label{sec:fail-ten}

With ten records per party there is rarely enough evidence to pass the gate. In run 2 the typed learner proposed in 22.3\% of party-draws at 10 records (Table~\ref{tab:app-gate}). Its accuracy (93.0\%) was close to the public rule's (91.7\%), though still ahead of every baseline (Table~\ref{tab:app-diff-v3}).

The trained-base experiment shows the cost more sharply. At 10 records the gate declined almost always (96\% of party-draws), so the combination was the trained base. Fine-tuning and Ditto pFL from a better-trained base beat it even when CV-tuned (combination minus comparator, in points):
\begin{itemize}[nosep]
  \item fine-tuning: \ensuremath{-0.5} [\ensuremath{-0.9}, \ensuremath{-0.2}];
  \item Ditto pFL: \ensuremath{-0.5} [\ensuremath{-1.0}, \ensuremath{-0.1}].
\end{itemize}
Oracle-tuned fine-tuning beat it by more (\ensuremath{-2.4} [\ensuremath{-2.9}, \ensuremath{-1.8}]). Table~\ref{tab:app-hybrid-sens} gives every comparator at every record count.

Real records are small in a different way. A single failure proves a tightening for that lot and test condition. Claiming survival at 90\% probability with 90\% confidence takes 22 clean parts, but the largest lot in the compiled reports was 15, and most per-point sample sizes are unstated.

\subsection{Contaminated and wrong-party records, and the gate}\label{sec:fail-gate}

The gate tests whether a declaration fits a party's records better than chance. It does not test whether the records are about that party.

\paragraph{Shuffled outcomes.}
With outcomes permuted within each party's records, the typed learner proposed in 46.3\% of party-draws under the judgement gate (run 1) and 76.0\% under the tuned gate (run 2). The registered acceptable rate was at most 10\%, so by that criterion the gate is miscalibrated in both runs. Part of this is not a gate error. Once outcomes are independent of the inputs, failures among accepted parts far exceed what the public rule implies, so a tightening does fit shuffled records better. That is why the gate tuning plan used the public-rule null instead. On that null the tuned gate is close to its bound, and it exceeds the bound at 100 records (Section~\ref{sec:res-gate}). Either way, the proposals made on shuffled records lowered accuracy, by \ensuremath{-7.8} points [\ensuremath{-9.2}, \ensuremath{-6.4}] against the public rule in run 2.

\paragraph{Another party's records.}
When each party learned from the next party's records, the tuned gate proposed in 77.7\% of party-draws (233/300), and 82.8\% of those proposals were worse than the public rule. Accuracy changed by \ensuremath{-2.1} points [\ensuremath{-2.9}, \ensuremath{-1.3}]. The plan stated in advance that no gate would catch this. Records that are internally consistent but describe another population pass any test on the records themselves. Only an acceptance body holding a reference set that no party chose could catch it, and that reference set does not yet exist (Section~\ref{sec:threat-reference}).

\paragraph{Label noise.}
With 10\% of labels flipped, 17.8\% of the typed learner's proposals were worse than the public rule, against 28.0\% without the gate (Table~\ref{tab:app-gate}). The gate limited the harm, but it did not remove it, and it did not raise accuracy: the ungated learner scored 94.8\% against 94.5\% gated. Both exceeded every federated and pooled baseline (Table~\ref{tab:app-diff-v3}), so that advantage did not come from the gate.

\subsection{Party voting as acceptance}\label{sec:fail-voting}

An obvious alternative to an acceptance body is to let the parties vote. The conflict bench tried unanimity, majority and veto on a shared screen with overlapping private knowledge. Each failed in a way the typed acceptance test is designed to avoid.
\begin{itemize}[nosep]
  \item \emph{Majority passes hostile edits.} Under a two-of-three majority, voting let a hostile patch through: an edit that harms one party passed with the votes of the two it did not affect. Two parties with opposite corrections each passed in turn, so the model oscillated and the last writer won.
  \item \emph{Unanimity lets a wrong party block a right fix,} and invites creep. When two parties' regimes conflicted, the only limit all three accepted was a creep into one party's sampling gap. The creep passed unanimously and broke a real case: a silent regression that broke as many inputs as it fixed. With noisy labels, a wrong party vetoed the correct fix in 32 of 200 random ledgers.
  \item \emph{Unanimity accepts memorised exceptions.} A point fit that flags one reading only at one temperature passed every vote and repaired only its own slice. It encodes nothing true. Only truth on the whole grid, which no consortium holds, exposes it.
  \item \emph{The veto is an oracle.} By probing with proposals and watching one party's vote, another party read that party's private limit as 37 against a true 38, in nine probes. Revealing votes only in aggregate did not stop this when the third party was blind to the probed regime.
  \item \emph{Misreporting passes a no-regression veto.} In the covert-channel bench, a party that declared a false value for its own exception passed every company's private veto, because the false value was still better than no change. It got 823,770 (0.65\%) of its own inputs wrong, a figure nobody else could see.
\end{itemize}
A rule that let a veto stand only if a test house re-measured the disputed cases resolved these conflicts, disclosing 1 to 13 records. But the bench used the truth as the re-measurement, so that rule works by construction. This is why the acceptance test uses no votes (Section~\ref{sec:acceptance}, ``Why parties do not vote'').

\subsection{Hidden capacity: free-form edits against typed declarations}\label{sec:fail-hidden}

A free-form weight patch can carry information in choices that change no decision. In the ranking bench, the edits were exhaustively checked to have zero behavioural effect on every valid row, after composition in every order, after withdrawal and through serialisation. Within that constraint:
\begin{itemize}[nosep]
  \item one patch under the slice's edit limits carried about 440 bits;
  \item under the library's default limits it carried 7,989 bits.
\end{itemize}
Defences closed channels one at a time:
\begin{itemize}[nosep]
  \item forbidding idle edits;
  \item integer values only;
  \item a canonicaliser, which reduces hidden capacity in the weight channels to 0 bits;
  \item confining each party to its own column, which still left the near-zero-value channel open (about 440 bits).
\end{itemize}
An identity null direction survived the first two defences (about 11 bits per four edits).

For typed declarations with re-derivation, hidden capacity is zero by construction (Proposition~\ref{prop:hidden-floor}). The 3.46 bits a typed slot carries is the floor: the legitimate disclosure itself, which a party can also misreport.

A blind auditor reading patches without tools caught every random payload. When payloads were written as plausible engineering choices, its score fell to 3 of 4 payloads caught with 2 of 4 false alarms, and with mechanical aids 3 of 4 payloads caught with 1 of 4 false alarms. A margin of $-6$ against $-10$ cannot be told apart from the patch alone. These are lower bounds from a search, not proofs. The smuggler and colluder agents were not run: the API's safety classifier stopped the smuggler sub-agents before they replied (all three attempts), and no substitute was written.

\subsection{The language-model arm}\label{sec:fail-llm}

The pilot's language-model result went against the hypothesis, and Section~\ref{sec:threat-llm} gives the argument. The numbers are these. Given each party's records and the public rule, a frontier model deciding held-out boundary parts directly scored 347 of 360, against 348 of 360 for the patched network (six calls, one draw). Its stated rules, scored on every valid input and composed, reached 96.9\% at 30 records. Two calls on identical input gave identical answers on the probe, but their stated rules differed on 450 valid items.

Local open-weight models did well as readers and badly as authors:
\begin{itemize}[nosep]
  \item \emph{Reading prose test reports into fields}, each of three models was exact on 30 of 30, 30 of 30 and 30 of 30 reports respectively. The typed search on their extractions recovered the planted rule, changing 0 of 248,832 of the other parties' decisions. With one injected error, the rule was still uniquely recovered in 318 of 337 cases.
  \item \emph{Writing free-form weight edits}, two independent models made the same mistake: each dropped the heritage term, leaving 1,575 errors on the party's grid. Their edits fitted every record, so no check on the records could catch it.
  \item \emph{Deciding alone}, two local models scored 58 of 60 and 58 of 60 on one prompt.
\end{itemize}
The local runs are single calls on one draw. The pilot's model-alone arm has not been repeated across draws or model families on the registered grid.

\subsection{The exploratory loop pilot}\label{sec:fail-loop}

The loop pilot asked whether a local model, in a propose--check--revise loop with a mechanical checker, could write a declaration in a small grammar that recovers a truth outside the typed space. The planned run had 16 truths at two record counts. Only one ran: an interaction truth at 30 records, on the cost-measuring pilot.

The model's first proposal fitted the party's records perfectly (0 record errors, against 2 for the public rule). Because it fitted, the checker had nothing to feed back and the loop stopped after one round. On the party's full grid, however, the declaration made 2,517 errors, against 2,394 for the public rule. It overfitted its records. The instruction to prefer the smallest change produced the same declaration. On the held-out probe, the loop and the model deciding alone scored the same (40 of 60).

One truth supports no conclusion. The pilot shows only that a checker which feeds back record errors cannot detect overfitting, and that is the gate's job.

\subsection{What the real-record bench could not do}\label{sec:fail-real}

The real-record bench has no accuracy evaluation. Of 30 candidate report rows:
\begin{itemize}[nosep]
  \item 9 were excluded, for degradation rather than failure, ambiguity, protons or inconsistency;
  \item 6 of 30 were withheld by their source.
\end{itemize}
The typed slot omits the particle's linear energy transfer and the lot. So the one part reported by two parties could not be compared across them. Above the highest tested gate bias, the mechanical reading sets the limit to zero, whereas a blind reader left it uncovered. That choice is a policy decision for the acceptance body, not a fact in the records.

\section{Related work}\label{sec:related}

\paragraph{Rule compilation and theory revision.}
Knowledge-based artificial neural networks compiled propositional rules into sigmoid networks with fixed link weights, so that the network initially makes the same classifications as the rules, then refined every weight by backpropagation; they reached good accuracy from few examples, and by their authors' account the refined concepts became incomprehensible~\cite{towell-kbann}. The later recap of the line names spare zero-weighted units for missing rules~\cite{mooney-shavlik-recap}. Tracr and RASP compile programs into transformers and reserve a subspace for each variable~\cite{lindner-tracr,weiss-rasp}. Symbolic theory revision~\cite{ourston-either,richards-forte} and inductive logic programming~\cite{muggleton-ilp} revise rules from examples, and minimal change is their classic preference; language models have recently been used as hypothesis proposers under symbolic checks~\cite{wang-hypothesis-search,qiu-phenomenal,llm-theory-induction}. Our compiled prior is the established compile-and-refine idea made exact on a finite domain. What differs is the refinement: not gradients, which dissolve the rule, but typed declarations from several parties, each confined to its own units, so the compiled rule stays readable after every edit.

\paragraph{Model editing and weight deltas.}
ROME writes one fact into one layer by a rank-one update~\cite{meng-rome}; MEMIT, MEND and AlphaEdit scale, learn and protect such edits~\cite{meng-memit,mitchell-mend,fang-alphaedit}. GRACE keeps edits in a removable codebook outside the weights~\cite{hartvigsen-grace}. Task vectors compose by addition and are reduced by negation, only partially~\cite{ilharco-task-arithmetic}; merging and adapters manage interference between dense deltas~\cite{yadav-ties,wortsman-soups,hu-lora}. Two works edit collaboratively: CollabEdit merges several parties' edits under an explicit assumption that the parties are trustworthy~\cite{zheng-collabedit}, and FLEKE shares edit vectors in a simulated federation~\cite{fleke-fededit}. Neither gives an edit an owner, confines it, or withdraws it. Surveys cover the wider field~\cite{yao-editing-survey}.

\paragraph{Language models that read and write weights.}
Weight-space programming has a general-purpose language model read and write a network's weights directly~\cite{takehi-wsp-blog}; follow-up studies report large gains from very local edits on small models and small gains on larger ones~\cite{hu-wsp-scope}, a blind audit of small clinical-screening networks in which testing against the written protocol caught every planted flaw~\cite{nusrat-wsp-audit}, and extensions to imaging, biosignals, games, model checking and deployed predictors~\cite{lindholz-wsp-imaging,varam-wsp-biosignals,garrido-merchan-wsp-games,gross-wsp-model-checking,li-weightloop}. Each of these studies has one editor acting for one owner. FedReason brings a reading model into federated learning: every ten rounds a language model inspects the client models through code run in a server-side sandbox and writes the aggregation program the server then applies~\cite{ruzbahani-fedreason}. The model chooses how dense client updates are combined, so no update has an owner who can confine or withdraw it. Our setting has several parties who do not trust one another, so edits are typed, owned and withdrawable, and we measure what an edit discloses and what it could hide. Our free-form ablation and the hidden-capacity count are the direct comparison with this line. Our code builds on the open-source ruv-FANN implementation of weight patches~\cite{ruv-fann} (see Availability, interests and disclosures).

\paragraph{Federated learning, personalisation and withdrawal.}
Federated averaging learns a shared model from local updates~\cite{mcmahan-fedavg,kairouz-open-problems}; personalised methods keep per-client heads or layers beside a shared part~\cite{collins-fedrep,arivazhagan-fedper,li-ditto}, and every client's data still shapes that shared part. SISA makes unlearning exact by retraining only an isolated shard~\cite{bourtoule-sisa}; one-shot federated editing makes adding or withdrawing a client exact in distribution, given a trusted coordinator and full-model uploads~\cite{tao-fsosa}; surveys cover federated and general unlearning~\cite{cao-unlearning,liu-federaser,khalil-not,liu-fu-survey,romandini-fu-survey}. Zero interference and exact withdrawal follow from disjoint support, as in SISA's shards, and we present them as consequences we verify, not as discoveries. Ditto, FedRep and federated averaging are three of our seven baselines.

\paragraph{Learning from few records.}
For a finite class under realisability a consistent learner needs a number of examples that grows with the logarithm of the class size~\cite{blumer-occam,valiant-pac,shalev-shwartz-uml}, and strong structured priors give similar effects in concept learning~\cite{lake-bpl}. A small typed space learns from few records for that known reason; we claim no new learning-theoretic effect.

\paragraph{Contribution, leakage, hidden capacity and verification.}
Data Shapley and influence methods value contributions~\cite{ghorbani-data-shapley,jia-shapley,koh-influence,pruthi-tracin,wang-fl-shapley,song-fedsv}; with disjoint edits, attribution is direct. Shared gradients and updates leak data, membership and properties~\cite{zhu-dlg,geiping-inverting,shokri-mia,nasr-privacy,melis-leakage}, and differential privacy is the standard bound~\cite{dwork-dp}; our disclosure figure is a count and an upper bound, not a differential-privacy guarantee. Weights can carry undeclared content: covert channels~\cite{lampson-confinement}, memorised data~\cite{song-remember}, hidden payloads~\cite{liu-stegonet,wang-evilmodel} and backdoors that pass validation~\cite{gu-badnets} or cannot be detected~\cite{goldwasser-backdoors}. Reading weights therefore cannot certify an edit. Small ReLU networks can be verified exactly by SMT and mixed-integer methods~\cite{katz-reluplex,katz-marabou,tjeng-mip} and repaired with guarantees~\cite{sotoudeh-prdnn}.

\paragraph{Sector setting.}
The published rule in our real-record bench is a derating standard~\cite{ecss-q-st-30-11c-rev2}, beside a radiation assurance standard~\cite{ecss-q-st-60-15c-rev1} and a United States baseline~\cite{nasa-eee-inst-002}; records come from public NASA Goddard test compendia~\cite{gsfc-compendium-2014,gsfc-compendium-2015,gsfc-compendium-2016,gsfc-compendium-2017,gsfc-compendium-2019,gsfc-compendium-2023,gsfc-compendium-2024,gsfc-compendium-2025}. ESA has reported more demand to obtain commercial-parts data than willingness to provide it~\cite{esa-esccon-cots}, and a United States effort proposed a vetted multi-agency radiation data repository~\cite{yarbrough-srhec}. Multi-party computation already serves conjunction assessment between satellite operators, where the private input is state rather than a rule~\cite{hemenway-mpc-conjunction,veskus-sharesat}. UK competition guidance treats exchanges of current capacity as high risk~\cite{cma184}, which bears on what a declaration may name.

\paragraph{What appears to be new.}
A search of arXiv abstracts for weight-space programming, federated learning with knowledge editing, and multi-party model editing with withdrawal or attribution, with the closest papers read in full, found no prior work in which several parties improve a shared, exactly compiled rule network through typed declarations, each compiled to an owned weight edit with exact withdrawal and an exhaustive acceptance test; and none that counts each party's disclosure in values and bits or measures the hidden capacity of the edit language. The protocol and these measurements are the contribution; the individual properties are not.

\section{Limitations and the case against}\label{sec:limitations}\label{sec:threats}

\paragraph{What stands.}
Under two analyses registered before they ran, on hidden truths sampled at random from the typed space and against seven baselines each tuned on its own records, a learner that searches a small typed space around a compiled prior, under an evidence gate, was non-inferior to every baseline at 30 records per party, and superior to most of them after Holm correction. A descriptive ablation without the prior, which also changed the space and the search, lost far more accuracy than any baseline gap. Over a trained base, typed declarations beat both the base alone and the declarations over the published rule, in a third registered test. Throughout, each declaration changed no decision about another subject, withdrew exactly, and disclosed a handful of values. The rest of this section argues against that record as a hostile reviewer would, and says where the reviewer is right.

\subsection{The typed space was built to contain the owner-specific truths}\label{sec:lim-tailored}\label{sec:threat-tailored}
The primary cell samples truths from the typed space itself, so realisability (Proposition~\ref{prop:occam}) holds by design: the experiment shows that a learner told the shape of the answer can find its size from few records. The free-form linear space, about a million declarations, measures the cost of a space wide enough not to be tailored, but the spaces are nested (Section~\ref{sec:typed}), so it still contains every planted truth: it measures the cost of a larger space, not of a misspecified one. It too exceeded every CV-tuned baseline (95.7\% with the gate; Table~\ref{tab:app-arms-v3}), which places most of the margin on the prior and on search over a space that contains the truth, not on the narrowness of the type.

Two things mitigate this. Truth classes outside the space, a component shared by all parties, an interaction the type lacks and a relaxation, are reported with the same prominence (Section~\ref{sec:results}), and there the typed learner loses or declines. And the claim is narrowed accordingly: where a community can agree in advance on the \emph{kinds} of exception it will accept, as a published standard often already does, a learner confined to those kinds needs few records and leaves an artefact with the stated properties. We do not claim it discovers exceptions of unknown shape. Whether real exceptions fall into a type that a standards body would agree in advance is open, and real records already showed two kinds that the type cannot hold (Section~\ref{sec:lim-real}).

\subsection{The accuracy evidence is synthetic}\label{sec:lim-synthetic}\label{sec:threat-synthetic}
Every accuracy figure comes from an invented screening world with three parties, a generator that concentrates items near the decision boundary, and deployment items drawn from the same generator as the records. Real test records come in small lots, with censored outcomes and conditions the encoding omits. The real-record bench establishes only that a published rule compiles exactly and that the operations behave as Section~\ref{sec:guarantees} says on a realistic network; it has no accuracy evaluation, because there is no held-out truth for real parts. Its declarations were written from the reports by a fixed procedure: no learner searched and no evidence gate ran on real records. We report orderings and paired differences, but we make no claim that either the magnitudes or the orderings transfer to real data: within the synthetic bench the ordering already reverses on shared truths and on relaxations (Sections~\ref{sec:fail-shared} and~\ref{sec:fail-lenient}).

\subsection{A language model given the records decides about as accurately}\label{sec:lim-llm}\label{sec:threat-llm}
In a pilot comparison, run once on one draw before the registered plans, a frontier language model given a subject's records and the published rule decided 347 of 360 held-out probe items correctly, against 348 of 360 for the patched network. We accept that finding and do not rest the case on accuracy against a language model. The typed learner uses no language model; the comparison is between two processes. The model-alone process needs the raw records in front of the model at every decision, so its disclosure is the records themselves, every time; it leaves no artefact to check, attribute or withdraw; and its stated rule is not stable: two identical calls gave identical answers on the probe but stated rules that differ on 450 valid items. The probe was weak (rules differing on thousands of items scored alike on sixty), and the repeat rests on one pair of calls. Whether these process properties are worth their cost is a question for the parties.

\subsection{Non-interference, withdrawal and bounded disclosure follow from the construction}\label{sec:lim-construction}\label{sec:threat-construction}
They do, and the paper is organised around the point: they appear as propositions with assumptions, and the experiments treat them as acceptance checks. They do not separate the typed learner from a confined gradient learner, which has both. Against that learner the typed learner must win on accuracy or on disclosure. In the registered runs it was more accurate at the primary cell after correction (S1 in run~2, \ensuremath{+3.4} points [\ensuremath{+2.5}, \ensuremath{+4.3}]). No disclosure figure is reported for the confined gradient learner at the primary cell, so the separation from it rests on accuracy. The disclosure bound that was 76.9 times smaller at minimum is against Ditto pFL, a federated baseline, and is a ratio of two upper bounds (Section~\ref{sec:guarantees}). Over the trained base the confined gradient learner disclosed 3.9 values per party against 2.2 for the combination (Table~\ref{tab:app-hybrid}), the same order of magnitude.

\subsection{The acceptance test depends on a reference set that does not exist}\label{sec:lim-reference}\label{sec:threat-reference}
Typing bounds what a declaration can carry; only evidence can say whether it is true. The only check on truth that does not rely on the declaring party is the reference test, and it needs labelled items that no party chose. No such set exists for any application we know of. Without it the acceptance test reduces to a type check, exhaustive behavioural checks and trust in the party's reported gate statistic, backed by a commitment opened in a dispute. Who would hold a reference set, how it would cover the rare regions declarations change, and how it would avoid leaking the body's labels to parties who probe it are open. Votes are not a substitute: in the pilot they protected only sampled regions and leaked private limits.

\subsection{This is a rule table with an owner column}\label{sec:lim-table}\label{sec:threat-table}
On a rule-only base the objection is correct. A table of typed exceptions run beside the published rule makes the same decisions (Corollary~\ref{cor:alone}), discloses the same values, is free of interference by scoping, withdraws by deleting a row, is easier to read, and has the same hidden capacity under re-derivation, since Definition~\ref{def:hidden} applies to any message language. The typed space, the gate, the acceptance test and the measurement of disclosure and hidden capacity would all carry over to a table unchanged.

The registered trained-base result changes the case but does not overturn it. It shows that trained weights can carry a shared component that no typed exception expresses, and that the combination is more accurate than either part: \ensuremath{+5.4} points over the base alone and \ensuremath{+3.1} over declarations on the published rule. It does not show that holding the exceptions as weights beats holding them as a table run beside the trained network, which would make the same decisions $\dec_b\wedge\neg\sem{d}$. What the weight form adds is one artefact: the trained component and every owned exception hashed, checked exhaustively and withdrawn by one procedure.

Its comparators also have limits. The private fine-tuning, Ditto and pooled arms share one small network and trainer, and the pooled arm mostly chose very short training under cross-validation; a check run after unblinding found that, oracle-tuned, private fine-tuning and Ditto match the combination at 30 records, and that at 10 records, where the gate almost always declines, fine-tuning from a well-trained base beats it. Against those learners the combination's advantage is in disclosure and withdrawal, not accuracy. The experiment also assumes 300 shared records labelled by the shared component alone, with no owner exceptions in their labels. A clean shared corpus of that kind is a resource of the same sort as the reference set of Section~\ref{sec:lim-reference}, which does not exist for any application we know of; with shared labels contaminated by private exceptions the result is untested.

\subsection{The evidence gate missed a pre-stated criterion}\label{sec:lim-gate}
The plans fixed an acceptable rate of at most 10\% proposals on shuffled outcomes. The gate proposed far more often in both runs (Section~\ref{sec:guarantees}); that criterion failed. Two further weaknesses were not pre-stated criteria. The gate does not detect another party's records, which the plans did not expect it to: proposals made from a wrong party's records were mostly worse than the published rule. And in an exploratory null check written after the second run, the tuned setting's false-proposal rate was close to its development estimate at 10 and 30 records but exceeded 5\% at 100. The gate is a dial, not a guarantee: the tuned setting gained \ensuremath{+1.3} points over the judgement setting at the primary cell and proposed more often on contaminated records, moving more of the burden of rejecting wrong declarations onto the acceptance test. Part of the second run's wider margin over the baselines comes from the tuned gate and part from its fresh sample of truths. Separately, the first run's primary endpoint \emph{as run} fails: its disclosure clause counted federated values changed rather than values uploaded, giving one draw a zero bound. The code was corrected to the plan's definition and the cell re-run with accuracy and spill byte-identical; both readings are recorded. The second plan made accuracy the only endpoint.

\subsection{Real records showed what one slot cannot hold}\label{sec:lim-real}\label{sec:threat-real}
On public single-event test reports the declaration type was too small in four ways. Destructive outcomes depend on particle energy and manufacturing lot, which the slot omits. 12 of 49 gate-bias points were censored passes (``no failure up to $X$''), evidence the rule's ``whichever is lower'' clause discards, so the published rule cannot express leniency and no tightening-only type can. 8 rows were tested at several gate biases and need several slots per part. And while one failure justifies a tightening, a supported relaxation needs many clean parts: about 22 with zero failures for 90\% reliability at 90\% confidence, against a largest compiled sample of 15. Adding a slot with $v$ values costs at most $\log_2v$ bits of disclosure and $\ln v$ in the Occam numerator per declaration, so the cost of a richer type is stated, not hidden; whether a type rich enough for real data stays small enough for the guarantees to bite is the central open question for any application.

\subsection{Further objections}\label{sec:lim-further}
\begin{description}[style=nextline,leftmargin=1.5em,itemsep=2pt]
  \item[This is theory revision with enumeration.] Largely, yes. Our claim is about the protocol between parties: the typed artefact, its gate and acceptance test, exact withdrawal, and disclosure and hidden capacity measured as properties of the artefact language. It is not a new induction algorithm.
  \item[The tie-break was chosen after the pilot.] Minimal change was introduced after the pilot and registered as primary in both confirmatory runs; the pre-stated order is reported beside it, and the registered comparison between them showed no significant difference in either run.
  \item[The baselines are small.] The federated and small pooled baselines use the 79-parameter architecture of the shared model. The plans therefore added a pooled network sized to the data and pooled gradient-boosted trees, and tuned every baseline by cross-validation on its own records, with oracle-tuned figures reported separately as upper bounds. In the first run the typed learner was not shown superior to the larger pooled network or to FedRep.
  \item[Ten records are not enough.] At 10 records the gate usually declines and exact recovery of the truth is rare; there the learner's value is that it does no harm, not that it learns.
  \item[No adversary was run.] No adversarial party tried to pass a false declaration or a covert payload through the typed acceptance test. Attempts to run adversarial authors against free-form edits did not complete (all three attempts were stopped before replying), and nothing was substituted. Hidden capacity for typed declarations is a theorem and needs no adversary; whether false declarations get through does, and it is open. Collusion and one party acting under several identities are not addressed.
  \item[The same team built the benches, the spaces and the truths.] Researcher degrees of freedom are real. The plans fixed generators, typed spaces, orders, gate settings, baselines and analysis code before any confirmatory draw, every draw is reported, and the departures from plan are recorded with their reasons.
  \item[Exhaustive checking does not scale.] The propositions do not need it; the acceptance test as written does, and beyond domains of a few hundred million inputs each patch's region of decision change would need a symbolic check. That is untested.
  \item[The record reader.] Turning prose test reports into typed records would use a language model. When local models were asked to \emph{author} weight patches directly in the pilot, they dropped a term of the published rule, an error the records could not reveal, which is one reason the reader is confined to extraction. Its accuracy on real reports is not established.
\end{description}

\section*{Availability, interests and disclosures}\label{sec:availability}
\addcontentsline{toc}{section}{Availability, interests and disclosures}

\paragraph{Code and data.}
The experimental code (rule compiler, declaration compiler, typed learner, acceptance checks, baselines and every bench), together with the results ledger that maps each value printed in this paper to the file and line that produced it and the command that regenerates it, is held today in a private repository of DreamLab AI Consulting Ltd. It is available from the author on request, at the revision that produced this manuscript; a public release will be announced in a later version of this paper. The synthetic benches are generated by that code from seeds named in it. The real-record bench uses single-event test rows from NASA Goddard compendia~\cite{gsfc-compendium-2014,gsfc-compendium-2015,gsfc-compendium-2016,gsfc-compendium-2017,gsfc-compendium-2019,gsfc-compendium-2023,gsfc-compendium-2024,gsfc-compendium-2025}, which are marked Public Use Permitted. The derating standard~\cite{ecss-q-st-30-11c-rev2} may be quoted; reproduction needs the publisher's consent. The compiled weights of that standard's rule are not distributed without the publisher's consent.

\paragraph{Third-party software.}
Parts of the experimental code are a Rust port of the patch, evaluation, workspace and workbench logic of the npm package \texttt{ruv-fann-wsp}~\cite{ruv-fann}\footnote{\url{https://github.com/ruvnet/ruv-FANN/tree/main/packages/wsp}}, published under the MIT licence; that licence and its copyright notice are reproduced in the repository. The forward pass is provided by the \texttt{ruv-fann} crate (MIT or Apache-2.0), used unmodified and pinned to version 0.2.2. The two interchange schemas for dense layers and weight patches are the upstream project's formats, used for interoperability.

\paragraph{Competing interests.}
The author is director of DreamLab AI Consulting Ltd, which develops the systems described in Section~\ref{sec:protocol-tiers}. The company is applying for UK public funding to develop the mechanism further and may offer services based on it. No patent application has been filed on the mechanism.

\paragraph{Acknowledgements.}
We thank Reuven Cohen (ruvnet) and the contributors to \texttt{ruv-FANN}~\cite{ruv-fann} for the weight-patch package this work builds on, and the NASA Goddard radiation effects group for publishing its test compendia for public use.

\bibliographystyle{plainnat}
\bibliography{refs}

\appendix
\section{Proofs and supporting analysis}\label{app:proofs}

This appendix proves the results stated in Section~\ref{sec:guarantees} and records the analysis that the main text summarises. Assumptions~\ref{ass:encoding} and~\ref{ass:float} hold throughout unless stated otherwise.

\subsection{Exact compilation}

\begin{proof}[Proof of Proposition~\ref{prop:exact}]
\emph{Construction M.} Each violation $v_i(x)=c_i-w_i\cdot x$ is an integer, and atom $i$ holds if and only if $v_i(x)\le0$. If every base atom holds and no gated exception fires, every $\ReLU(v_i)=0$ and $f(x)=1\ge\tfrac12$. If some atom fails, $v_i(x)\ge1$, every other term is non-positive, and $f(x)\le0<\tfrac12$. For a gated exception unit the pre-activation is $\ell(x)-G(1-x_{s,k})-G(1-\sigma(x))$. On an item of subject $k$ in scope it equals the integer $\ell(x)$, which contributes $-\ReLU(\ell(x))\le-1$ exactly when $\ell(x)\ge1$. Otherwise one bracket equals 1 and the pre-activation is at most $\ell(x)-G\le0$, because $G\ge\max_\X\ell$; the unit is silent.

\emph{Construction I.} For $a\in\tfrac1q\Z$, $q(\ReLU(a+\tfrac1q)-\ReLU(a))$ equals 1 if $a\ge0$ and 0 if $a\le-\tfrac1q$, and no other values of $a$ occur on the lattice. A conjunction unit $\ReLU(\sum_{l=1}^L b_l-(L-1))$ with $b_l\in\{0,1\}$ equals 1 when all $b_l=1$ and 0 otherwise, and the output $\ReLU(b_0-\sum_j e_j)$ with $b_0,e_j\in\{0,1\}$ equals $b_0\wedge\neg\bigvee_j e_j$.

\emph{Floating point.} Under Assumption~\ref{ass:float} every operation maps exactly representable operands to an exactly representable result, so no rounding occurs. Construction~M uses integer weights; Construction~I uses integer first-layer weights and second-layer weights $\pm q$, integers when $q$ is.
\end{proof}

\begin{remark}[Known in another guise]
Writing a propositional rule into network units is the starting point of knowledge-based artificial neural networks~\cite{towell-kbann}, and compiling programs into transformer weights is the same idea at a larger scale~\cite{weiss-rasp,lindner-tracr}. What the lattice and floating-point conditions add is that ``exact'' becomes a property checked on every input rather than one that holds approximately. Off the lattice the indicator interpolates linearly and the network no longer computes a decision rule.
\end{remark}

\subsection{Composition, non-interference and withdrawal}

\begin{lemma}[Disjoint patches commute]\label{lem:commute}
If $\supp(P)\cap\supp(Q)=\emptyset$ then $\theta\oplusP P\oplusP Q=\theta\oplusP Q\oplusP P$, so any order of pairwise disjoint patches gives the same parameters and the same hash.
\end{lemma}
\begin{proof}
Assignments to distinct coordinates commute, and each coordinate is written by at most one patch.
\end{proof}

\begin{proof}[Proof of Proposition~\ref{prop:noninterference}]
Order the units topologically. For a unit outside $W$, the incoming weights and bias are unchanged by $P$ except possibly weights from units of $W$. Fix $x\notin\Xk$. By hypothesis, every unit of $W$ that feeds a unit outside $W$ is silent on $x$, so by induction over layers every unit outside $W$ has the same activation under $\theta\oplusP P$ as under $\theta$, and so does the output. Under $\theta$ the units of $W$ contributed nothing, because they were spare. Commutation is Lemma~\ref{lem:commute}.

\emph{Screening compile map.} $W=\{u_k\}$, whose only outgoing weight goes to the output. Off $\Xk$ the input for subject $k$ is 0, so the pre-activation of $u_k$ is $\ell_d(x)+\beta'-G_d$ with $\beta'\le0$ collecting the scope terms; since $G_d\ge\max_\X\ell_d$, it is at most 0.

\emph{Published-rule compile map.} $W$ is six first-layer units and one conjunction unit, and only the conjunction unit feeds outside $W$. Its inputs include the part's pass-through unit, which equals $x_{s,k}=0$ off $\Xk$; a conjunction of four indicators one of which is 0 is 0.
\end{proof}

\begin{proof}[Proof of Corollary~\ref{cor:alone}]
By Proposition~\ref{prop:noninterference} applied to each other subject's patch, only $u_k$ and the public units are active on $x\in\Xk$. Then $f(x)=1-\ReLU(v_0(x))-\ReLU(v_1(x))-\ReLU(\ell_{d_k}(x))$ on in-scope inputs, and the integer argument of Proposition~\ref{prop:exact} applies. Order is irrelevant by Lemma~\ref{lem:commute}. For two parties declaring on one part, the live slots on that part form a union of rejected regions; union is commutative and monotone, so the effective limit is the minimum.
\end{proof}

\begin{proof}[Proof of Corollary~\ref{cor:trained}]
Off $\sem{d}$ (including every item of another subject, by the gate weight) the reserved unit's pre-activation is at most 0, so the output equals $f_{\theta_b}$. On $\sem{d}$ the pre-activation is an integer at least 1, so the output is at most $f_{\theta_b}(x)-M_d<\tfrac12$. Hence $\dec_{\theta_b\oplus\kap(d)}=\dec_b\wedge\neg\sem{d}$. The proofs of Propositions~\ref{prop:noninterference} and~\ref{prop:withdraw} never use how the base was obtained, only that the reserved units are spare in it.
\end{proof}

\begin{proof}[Proof of Proposition~\ref{prop:withdraw}]
Compare coordinates. For $\lambda\in\supp(P_i)$ the left side holds $(\theta_0)_\lambda$ by definition of $P_i^{-1}$, and so does the right, because no other patch writes $\lambda$. For $\lambda\in\supp(P_j)$, $j\neq i$, both sides hold $P_j(\lambda)$. Elsewhere both hold $(\theta_0)_\lambda$. Assignment copies values, so no arithmetic and no rounding is involved, and the canonical serialisation prints equal reals identically, so the hashes agree. Later disjoint admissions and withdrawals act on other coordinates and preserve the equality.
\end{proof}

\begin{remark}[Overlapping support and additive semantics]
If two live patches write the same location, the later write wins: composition then depends on order, and withdrawing the earlier patch restores initial values over the later one's edits. A pilot bench with overlapping knowledge showed both effects, which is why Algorithm~\ref{alg:accept} refuses a patch whose support meets a live patch, and why the number of live declarations is capped by the number of reserved slots. Under additive semantics, $\theta+\tau-\tau$ need not equal $\theta$ in binary64, and withdrawal would be exact only for values satisfying Assumption~\ref{ass:float}. That disjoint, gated changes do not interfere and can be removed is known in several forms: negation of task vectors, approximate because the vectors overlap~\cite{ilharco-task-arithmetic}; keyed adapters in model editing~\cite{hartvigsen-grace}; per-client heads in personalised federated learning~\cite{collins-fedrep,li-ditto}; and exact unlearning by sharding and retraining~\cite{bourtoule-sisa}.
\end{remark}

\begin{remark}[What the guarantee is relative to]
Non-interference is a statement about $\X$. The gate weight is the maximum of $\ell_d$ over $\X$, so an input outside the stated bounds, or with two subject bits set, can defeat it. ``No spill'' in our tables means no spill on $\X$, checked exhaustively in the synthetic benches; for the real-record bench see Section~\ref{sec:guarantees}.
\end{remark}

\subsection{Identifiability of confinement}

\begin{proof}[Proof of Proposition~\ref{prop:identify}]
The empirical loss is a function of $(f(x_i))_i$ with every $x_i\in\Xk$ only, so hypotheses that agree on $\Xk$ have equal loss.

For the gradient statement, take a hidden unit with pre-activation $z(x)=w\cdot\phi(x)+\sum_{j\in\Sub}u_jx_{s,j}+b$, where $\phi$ collects the non-subject inputs, trained on $R_k$ alone with the same step-size schedule for $u_k$ and $b$, an optimiser whose update for a coordinate depends only on that coordinate's gradient history, and an objective that depends on the parameters only through the predictions on $R_k$ (or a regulariser treating $u_k$ and $b$ identically from equal starting values). On every record $x_{s,k}=1$ and $x_{s,j}=0$ for $j\neq k$, so $\partial z/\partial u_k=\partial z/\partial b=1$ and $\partial z/\partial u_j=0$. By the chain rule $u_k$ and $b$ receive identical gradients at every step, hence identical updates, and $u_j$ for $j\neq k$ does not move. The change of $z$ on an item of another subject is therefore $\Delta b+\Delta w\cdot\phi(x)$. A regulariser that treats $u_k$ and $b$ differently, such as weight decay on weights but not biases, breaks the equality, and the conclusion then holds for the data-driven part of the update only.
\end{proof}

\begin{remark}[What this does and does not explain]
The proposition explains why a party fine-tuning the shared network on its own records will in general move other subjects' decisions, and why every confined arm we test, the typed learner and the confined gradient learner alike, is confined by construction: the gate weight on the subject input is fixed, not learned. It does not explain the accuracy differences in Section~\ref{sec:results}. In pooled and federated training the subject input varies across the data, so the argument does not apply there; their shortfall at small record counts is a matter of sample size and capacity. Nor does it separate the typed learner from the confined gradient learner.
\end{remark}

\subsection{Sample complexity}

\begin{proof}[Proof of Proposition~\ref{prop:occam}]
This is the standard bound for a consistent learner over a finite class~\cite{blumer-occam,valiant-pac,shalev-shwartz-uml}. A declaration with true error above $\varepsilon$ is consistent with $n$ independent records with probability at most $(1-\varepsilon)^n\le e^{-\varepsilon n}$; a union bound over at most $|\Hyp_k|$ such declarations, set equal to $\delta$, gives the result. Algorithm~\ref{alg:learn} without its gate returns a zero-error declaration whenever one exists, under either order; with the gate, the bound applies on the event that the gate passes, and otherwise the output is the published rule.
\end{proof}

\begin{table}[h]
\centering
\caption{Occam bound of Proposition~\ref{prop:occam} at $\delta=0.05$ ($\ln20=2.996$). These are definitional constants of the typed spaces, not results. The last column is $\lceil(\ln|\Hyp|+\ln20)/0.1\rceil$.}
\label{tab:occam}
\small
\begin{tabular}{@{}lrrrrrr@{}}
\toprule
space & $|\Hyp|$ & $\log_2|\Hyp|$ & $\ln|\Hyp|$ & $\varepsilon$, $n{=}30$ & $\varepsilon$, $n{=}100$ & $n$ for $\varepsilon{=}0.1$\\
\midrule
$\Hyp^{(1)}$ margins only & 73 & 6.19 & 4.29 & 0.243 & 0.073 & 73\\
$\Hyp^{(2)}$ all three forms & 127 & 6.99 & 4.84 & 0.261 & 0.078 & 79\\
$\Hyp^{(3)}$ linear & 1{,}113{,}751 & 20.09 & 13.92 & 0.564 & 0.169 & 170\\
\bottomrule
\end{tabular}
\end{table}

\paragraph{Against a parametric class.}
The natural parametric comparison is the confined gradient learner: ten trainable values on the same reserved unit (eight incoming weights on non-subject inputs, a bias and an output weight), with the subject gate fixed. Its tightening sub-class, $\{x\mapsto r(x)\wedge\neg\ind{w\cdot\phi(x)+b>\tau}\}$, removes a halfspace from the published rule. The class one-hots sum to one, so $\phi(\Xk)$ lies in a 7-dimensional affine subspace, where halfspaces have VC dimension 8. By the Sauer--Shelah lemma the sub-class induces at most $\sum_{i=0}^8\binom{N}{i}\le(eN/8)^8$ labellings of the $N=124{,}416$ items of $\Xk$, so $\ln(\#\text{labellings})\le8(1+\ln(N/8))\approx85.2$ nats against $\ln127\approx4.84$: the class-size term is about seventeen times larger. With a positive output weight the slot can also relax the rule, which no typed declaration can, so the full class is larger still.

\begin{remark}[What the sample-complexity analysis does not show]\leavevmode
\begin{enumerate}[nosep]
  \item \emph{Upper bounds against upper bounds.} No lower bound is proved for the parametric class on this distribution, so the comparison shows that the typed bound is smaller, not that the parametric learner needs more data. In distribution-free terms $\Hyp^{(2)}$ (VC dimension below 7) and a single gated halfspace (VC dimension 8) are comparable; the typed space's advantage is its small cardinality on the finite domain and, above all, realisability.
  \item \emph{Realisability is assumed.} It holds for owner-specific truths in our benches because the typed space was built to contain them. Without it the bound is replaced by an agnostic one with excess-risk term of order $\sqrt{(\ln|\Hyp|+\ln(2/\delta))/(2n)}$, about $0.38$ at $n=30$ and $|\Hyp|=127$, which says nothing useful.
  \item \emph{Magnitudes.} At $n\le30$ the bound is weaker than the published rule's own error. It explains the ordering of the spaces, not the accuracies observed.
  \item \emph{Gradient learners} with cross-validated stopping are not empirical risk minimisers, so the bound does not apply to them as stated. With label noise the relevant analysis is the agnostic one together with the gate.
\end{enumerate}
\end{remark}

\subsection{The evidence gate}

\begin{lemma}[Every improvement is bounded by the flips]\label{lem:flips}
If $t=r$ on the support of $Q_p$ and $F$ is the number of flipped labels in $R_p$, then $T(d)=\err(\none;R_p)-\err(d;R_p)\le F$ for every declaration $d$, and $\err(\none;R_p)=F$.
\end{lemma}
\begin{proof}
A record's label disagrees with the published rule exactly when it was flipped, so $\err(\none;R_p)=F$. Writing $T(d)=n_+(d)-n_-(d)$, where $n_+$ counts records $d$ decides correctly and the rule incorrectly, every record counted by $n_+$ is a flipped record, so $T(d)\le n_+(d)\le F$.
\end{proof}

\begin{proof}[Proof of Proposition~\ref{prop:gate}]
(i) By Lemma~\ref{lem:flips}, $\max_dT(d)\le F$, and $F$ is stochastically dominated by $\mathrm{Bin}(n,\eta)$. The gate passes only if $\max_dT(d)\ge c_\alpha(n,\eta)$, hence only if $F\ge c_\alpha$, an event of probability at most $\Pr[\mathrm{Bin}(n,\eta)\ge c_\alpha]\le\alpha$ by definition of $c_\alpha$.

(ii) Condition on the items $x_1,\dots,x_n$. Under the hypothesis, the observed labels are $y_i=r(x_i)\oplus\xi_i$ with $\xi_i$ independent Bernoulli$(\eta)$, and each replicate is $\tilde y_i=r(x_i)\oplus\xi^\ast_i$ with fresh independent Bernoulli$(\eta)$ flips. The statistic $T$ is the same deterministic function (the best improvement over the published rule found by the full search) of the observed labels as $T^\ast_k$ is of the replicate labels. So $T,T^\ast_1,\dots,T^\ast_K$ are independent and identically distributed, hence exchangeable. For a Monte Carlo test with $\hat p=(1+\#\{k:T^\ast_k\ge T\})/(K+1)$, exchangeability gives $\Pr[\hat p\le j/(K+1)]\le j/(K+1)$ for each integer $j$, since ties are counted against the observation; taking $j=\lfloor\alpha(K+1)\rfloor$ gives the bound. The additional requirement $T\ge m$, and the convention $\hat p=1$ when $T=0$, only shrink the set on which the gate passes.
\end{proof}

\begin{remark}[The gate as run]
In the experiments the replicates flip the published rule's decisions at $\hat\varepsilon=\err(\none;R_p)/n=F/n$, the observed flip fraction, rather than at a stated rate. The observed and replicate statistics are then no longer identically distributed, and part~(ii) does not apply. The statistic is still selection-aware, because every replicate reruns the search. Two failure modes of this choice are visible in our data. First, the realised false-proposal rate rises with the number of records (Section~\ref{sec:guarantees}), because with more records small chance improvements clear a one-record margin more often; the tuned setting is best read as a small-sample setting. Second, when the reference rule is worse than a constant rule, as for the empty rule in the no-prior ablation, $\hat\varepsilon$ exceeds one half and a constant declaration improves the fit under the null almost as much as a real one, so the gate cannot fire. We do not justify any gate threshold by analogy with sample-size tolerance factors in radiation-hardness standards: the factor we checked covers dose drift only, not destructive pass or fail.
\end{remark}

\subsection{Disclosure and hidden capacity}

\begin{proof}[Proof of Proposition~\ref{prop:disclosure-bound}]
Given $M$, $W\to R_p\to d\to A_p$ is a Markov chain, because the patch and hashes are deterministic functions of $d$ and public data and the learner's randomness is independent of $W$. The data-processing inequality gives the first two inequalities, and the third holds because $d$ takes at most $|\Hyp_p|$ values. Sums over events follow from the chain rule. If the gate statistic were published, $\log_2(n+1)$ bits per declaration would be added. Publishing a salted commitment would replace the information-theoretic bound by a computational one, which is why the commitment stays with the acceptance body.
\end{proof}

\begin{remark}[What the disclosure bound does not give]\label{rem:disclosure-not}\leavevmode
\begin{enumerate}[nosep]
  \item It is not a privacy guarantee for the declared fact, which is disclosed in full, nor differential privacy~\cite{dwork-dp}: there is no noise and no guarantee about any single record.
  \item The subject is always disclosed, and naming which component a party is qualifying may itself be commercially sensitive.
  \item No ratio against baselines is proved. A gradient or federated upload $U$ is a function of the records, so $I(R_p;U)\le H(R_p)$, an upper bound. A ratio of disclosures would need a lower bound on the baseline, which we do not have. The per-record caps used in the experiments, $\log_2248{,}832\approx17.9$ bits for an encodable labelled record and the generator's exact entropy, are upper bounds.
  \item Votes and probing are excluded by assumption: the acceptance test polls no party.
\end{enumerate}
\end{remark}

\paragraph{Full statement of the capacities.}
With the notation of Definition~\ref{def:hidden}, let $e:V\to\mathcal{M}$ be the honest encoding, with $\beta\circ e$ injective, and define the excess behavioural capacity $\mathrm{XC}=\log_2|\beta(\mathcal{M}_{\mathrm{acc}})|-\log_2|V|$. Partitioning accepted artefacts by behaviour gives $C\le\log_2|V|+\mathrm{XC}+\HC$. Hidden capacity is invisible to any behavioural test on $\X$; excess capacity is detectable, but only by a test that knows $V$.

\begin{proof}[Proof of Proposition~\ref{prop:hidden-floor}]
\emph{Floor.} $\nu$ preserves behaviour and $\beta\circ e$ is injective, so distinct facts give distinct normalised artefacts and $|\nu(\mathcal{M}_{\mathrm{acc}})|\ge|V|$. A dishonest party can submit $e(v')$ for any $v'\in V$, and no test that does not consult evidence can tell a false declared fact from a true one.

\emph{Typed language.} Take $\mathcal{M}=\Hyp_p$ with canonical parsing and serialisation, realisation $\kap$, and $\nu=\mathrm{can}$, the least-index representative of each class of $\equiv_\X$. Then $\nu$ maps each behaviour class to a single message, so every set in the definition of $\HC$ is a singleton and $\HC=0$; the image of $\nu$ has one element per class, which gives $C=\log_2|\Hyp_p/{\equiv_\X}|$ and $\mathrm{XC}=C-\log_2|V|$ for any fact space $V$ agreed independently of the type. The floor is attained when the classes correspond one to one with $V$.
\end{proof}

\begin{remark}[Scope of the hidden-capacity result]
Part~(ii) is informative only when $V$ is fixed independently of the type; in the ranking bench it is (the legitimate fact is one integer in $\{0,\dots,10\}$, so the floor is $\log_211\approx3.46$ bits). For free-form patches, $\HC$ is bounded below by counting accepted patches equivalent to an honest one; defences such as forbidding idle edits, requiring integer values or canonicalising free-form patches reduce it channel by channel, and these figures are lower bounds from a search, not proofs. A true canonical form of a free-form patch requires a search for a minimal equivalent patch whose cost grows with the network, one reason to type the artefact instead. Hiding payloads in weights without changing accuracy is known~\cite{liu-stegonet,wang-evilmodel}, as is confinement of information flow as a design problem~\cite{lampson-confinement}. What we add, so far as we know, is treating capacity as a property of the \emph{edit language}, and the observation that typing plus re-derivation reaches the information-theoretic floor.
\end{remark}

\subsection{Scale}
The propositions do not depend on exhaustive checking, but Algorithm~\ref{alg:accept} as written does. Exhaustive checks were feasible here: a few hundred thousand items per subject in the synthetic bench, 33 million inputs for the published rule, and 380{,}633{,}088 inputs for the largest rule compiled in a pilot bench. For larger domains each patch's region of decision change would need a symbolic check~\cite{katz-reluplex,katz-marabou,tjeng-mip}, and repair-style verification of a patched network~\cite{sotoudeh-prdnn} is the nearest existing tool. That is untested here.

\section{Full tables}\label{app:tables}

Every cell is a ledger entry: the value, with its 95\% interval in brackets. Accuracy intervals are $t$-intervals across sampled truths (df 19 for the confirmatory runs, 39 for the trained base); rates carry Wilson intervals, clustered within truths. Differences are typed learner minus comparator (trained base: combination minus comparator). An en dash marks a cell that was not run or has no ledger entry. Cells are labelled class, label noise, records per party.

\begin{table}[htbp]
\centering
\caption{Confirmatory run 2 (plan v3, tuned gate): accuracy of every arm in every cell, CV-tuned baselines.}
\label{tab:app-arms-v3}
\tiny
\setlength{\tabcolsep}{3pt}
\begin{tabular}{@{}p{2.4cm}p{1.65cm}p{1.65cm}p{1.65cm}p{1.65cm}p{1.65cm}p{1.65cm}p{1.65cm}@{}}
\toprule
\raggedright arm & \raggedright owner, 0\%, 10 & \raggedright owner, 0\%, 30 & \raggedright owner, 0\%, 100 & \raggedright owner, 10\%, 30 & \raggedright shared, 0\%, 30 & \raggedright interaction, 0\%, 30 & \raggedright relaxation, 0\%, 30 \tabularnewline
\midrule
\raggedright public rule & 91.7\%\newline\mbox{\tiny[90.2\%, 93.1\%]} & 91.7\%\newline\mbox{\tiny[90.2\%, 93.1\%]} & 91.7\%\newline\mbox{\tiny[90.2\%, 93.1\%]} & 91.7\%\newline\mbox{\tiny[90.2\%, 93.1\%]} & 91.4\%\newline\mbox{\tiny[89.9\%, 92.8\%]} & 93.4\%\newline\mbox{\tiny[92.2\%, 94.6\%]} & 93.5\%\newline\mbox{\tiny[92.5\%, 94.5\%]} \\
\raggedright typed, minimal change, gated (primary) & 93.0\%\newline\mbox{\tiny[92.1\%, 93.9\%]} & 96.6\%\newline\mbox{\tiny[96.2\%, 96.9\%]} & 99.6\%\newline\mbox{\tiny[99.5\%, 99.8\%]} & 94.5\%\newline\mbox{\tiny[94.0\%, 95.1\%]} & 96.3\%\newline\mbox{\tiny[95.7\%, 96.9\%]} & 95.3\%\newline\mbox{\tiny[94.7\%, 96.0\%]} & 93.5\%\newline\mbox{\tiny[92.5\%, 94.5\%]} \\
\raggedright typed, pre-stated tie-break, gated & 92.1\%\newline\mbox{\tiny[91.1\%, 93.0\%]} & 96.5\%\newline\mbox{\tiny[95.8\%, 97.2\%]} & 99.6\%\newline\mbox{\tiny[99.5\%, 99.7\%]} & 94.3\%\newline\mbox{\tiny[93.8\%, 94.7\%]} & 96.4\%\newline\mbox{\tiny[95.8\%, 97.0\%]} & 95.1\%\newline\mbox{\tiny[94.5\%, 95.7\%]} & 93.5\%\newline\mbox{\tiny[92.5\%, 94.5\%]} \\
\raggedright typed, minimal change, no gate & 93.6\%\newline\mbox{\tiny[92.7\%, 94.5\%]} & 96.6\%\newline\mbox{\tiny[96.3\%, 97.0\%]} & 99.6\%\newline\mbox{\tiny[99.5\%, 99.8\%]} & 94.8\%\newline\mbox{\tiny[94.3\%, 95.3\%]} & 96.5\%\newline\mbox{\tiny[95.8\%, 97.1\%]} & 95.3\%\newline\mbox{\tiny[94.7\%, 95.9\%]} & 93.5\%\newline\mbox{\tiny[92.5\%, 94.5\%]} \\
\raggedright typed, pre-stated, no gate & -- & 96.5\%\newline\mbox{\tiny[95.8\%, 97.2\%]} & -- & -- & -- & -- & -- \\
\raggedright free-form linear, gated & 91.6\%\newline\mbox{\tiny[90.4\%, 92.9\%]} & 95.7\%\newline\mbox{\tiny[95.1\%, 96.2\%]} & 99.3\%\newline\mbox{\tiny[99.1\%, 99.6\%]} & 92.9\%\newline\mbox{\tiny[92.2\%, 93.6\%]} & 95.8\%\newline\mbox{\tiny[95.0\%, 96.6\%]} & 94.7\%\newline\mbox{\tiny[94.2\%, 95.3\%]} & 93.5\%\newline\mbox{\tiny[92.5\%, 94.5\%]} \\
\raggedright free-form linear, no gate & -- & 95.6\%\newline\mbox{\tiny[95.0\%, 96.3\%]} & -- & -- & -- & -- & -- \\
\raggedright no compiled prior, gated & -- & 43.7\%\newline\mbox{\tiny[39.0\%, 48.3\%]} & -- & -- & -- & -- & -- \\
\raggedright no compiled prior, no gate & 74.7\%\newline\mbox{\tiny[73.7\%, 75.8\%]} & 84.0\%\newline\mbox{\tiny[82.8\%, 85.1\%]} & 90.5\%\newline\mbox{\tiny[89.4\%, 91.6\%]} & 80.6\%\newline\mbox{\tiny[79.5\%, 81.6\%]} & 82.7\%\newline\mbox{\tiny[81.8\%, 83.7\%]} & 84.1\%\newline\mbox{\tiny[82.6\%, 85.6\%]} & 87.8\%\newline\mbox{\tiny[86.6\%, 89.1\%]} \\
\raggedright confined gradient & 91.9\%\newline\mbox{\tiny[90.7\%, 93.1\%]} & 93.2\%\newline\mbox{\tiny[92.3\%, 94.1\%]} & 95.6\%\newline\mbox{\tiny[95.1\%, 96.0\%]} & 91.9\%\newline\mbox{\tiny[90.7\%, 93.1\%]} & 93.3\%\newline\mbox{\tiny[92.2\%, 94.4\%]} & 93.8\%\newline\mbox{\tiny[92.8\%, 94.7\%]} & 93.5\%\newline\mbox{\tiny[92.6\%, 94.4\%]} \\
\raggedright Ditto pFL & 90.9\%\newline\mbox{\tiny[89.4\%, 92.5\%]} & 93.2\%\newline\mbox{\tiny[92.4\%, 94.1\%]} & 96.5\%\newline\mbox{\tiny[95.8\%, 97.2\%]} & 80.2\%\newline\mbox{\tiny[78.4\%, 82.1\%]} & 96.0\%\newline\mbox{\tiny[94.8\%, 97.2\%]} & 93.5\%\newline\mbox{\tiny[92.4\%, 94.5\%]} & 93.7\%\newline\mbox{\tiny[92.5\%, 95.0\%]} \\
\raggedright FedRep pFL & 91.4\%\newline\mbox{\tiny[90.0\%, 92.7\%]} & 93.5\%\newline\mbox{\tiny[92.6\%, 94.4\%]} & 96.3\%\newline\mbox{\tiny[95.5\%, 97.1\%]} & 84.3\%\newline\mbox{\tiny[83.2\%, 85.5\%]} & 95.1\%\newline\mbox{\tiny[94.0\%, 96.1\%]} & 93.7\%\newline\mbox{\tiny[92.7\%, 94.7\%]} & 95.1\%\newline\mbox{\tiny[94.4\%, 95.8\%]} \\
\raggedright FedAvg & 90.9\%\newline\mbox{\tiny[89.4\%, 92.4\%]} & 92.5\%\newline\mbox{\tiny[91.4\%, 93.5\%]} & 94.5\%\newline\mbox{\tiny[93.8\%, 95.3\%]} & 85.5\%\newline\mbox{\tiny[83.9\%, 87.0\%]} & 96.1\%\newline\mbox{\tiny[94.9\%, 97.3\%]} & 93.6\%\newline\mbox{\tiny[92.5\%, 94.6\%]} & 93.7\%\newline\mbox{\tiny[92.6\%, 94.7\%]} \\
\raggedright pooled 79-param net & 91.3\%\newline\mbox{\tiny[90.1\%, 92.6\%]} & 92.9\%\newline\mbox{\tiny[92.0\%, 93.9\%]} & 96.2\%\newline\mbox{\tiny[95.7\%, 96.8\%]} & 83.5\%\newline\mbox{\tiny[82.1\%, 84.9\%]} & 96.0\%\newline\mbox{\tiny[94.9\%, 97.2\%]} & 93.6\%\newline\mbox{\tiny[92.6\%, 94.6\%]} & 93.8\%\newline\mbox{\tiny[92.8\%, 94.8\%]} \\
\raggedright pooled large net & 88.5\%\newline\mbox{\tiny[87.5\%, 89.5\%]} & 94.1\%\newline\mbox{\tiny[93.3\%, 94.8\%]} & 97.7\%\newline\mbox{\tiny[97.4\%, 98.0\%]} & 90.7\%\newline\mbox{\tiny[89.7\%, 91.7\%]} & 97.4\%\newline\mbox{\tiny[96.6\%, 98.3\%]} & 94.6\%\newline\mbox{\tiny[93.9\%, 95.3\%]} & 93.7\%\newline\mbox{\tiny[92.6\%, 94.7\%]} \\
\raggedright pooled gradient-boosted trees & 89.7\%\newline\mbox{\tiny[88.4\%, 91.0\%]} & 92.9\%\newline\mbox{\tiny[91.9\%, 93.8\%]} & 97.5\%\newline\mbox{\tiny[96.9\%, 98.1\%]} & 89.2\%\newline\mbox{\tiny[88.0\%, 90.4\%]} & 98.7\%\newline\mbox{\tiny[98.3\%, 99.2\%]} & 94.0\%\newline\mbox{\tiny[93.2\%, 94.9\%]} & 93.7\%\newline\mbox{\tiny[92.9\%, 94.6\%]} \\
\bottomrule
\end{tabular}
\end{table}

\begin{table}[htbp]
\centering
\caption{Confirmatory run 2 (plan v3, tuned gate): paired differences, typed learner (primary arm) minus each CV-tuned baseline, every cell. Exploratory except the owner, 0\%, 30-record cell.}
\label{tab:app-diff-v3}
\tiny
\setlength{\tabcolsep}{3pt}
\begin{tabular}{@{}p{1.8cm}p{1.7cm}p{1.7cm}p{1.7cm}p{1.7cm}p{1.7cm}p{1.7cm}p{1.7cm}@{}}
\toprule
\raggedright cell & \raggedright confined gradient & \raggedright Ditto pFL & \raggedright FedRep pFL & \raggedright FedAvg & \raggedright pooled 79-param net & \raggedright pooled large net & \raggedright pooled gradient-boosted trees \tabularnewline
\midrule
\raggedright owner, 0\%, 10 & \ensuremath{+1.1}\newline\mbox{\tiny[\ensuremath{+0.6}, \ensuremath{+1.6}]} & \ensuremath{+2.1}\newline\mbox{\tiny[\ensuremath{+1.0}, \ensuremath{+3.2}]} & \ensuremath{+1.6}\newline\mbox{\tiny[\ensuremath{+0.9}, \ensuremath{+2.3}]} & \ensuremath{+2.1}\newline\mbox{\tiny[\ensuremath{+1.1}, \ensuremath{+3.1}]} & \ensuremath{+1.7}\newline\mbox{\tiny[\ensuremath{+1.0}, \ensuremath{+2.3}]} & \ensuremath{+4.5}\newline\mbox{\tiny[\ensuremath{+3.5}, \ensuremath{+5.5}]} & \ensuremath{+3.3}\newline\mbox{\tiny[\ensuremath{+2.1}, \ensuremath{+4.5}]} \\
\raggedright owner, 0\%, 30 & \ensuremath{+3.4}\newline\mbox{\tiny[\ensuremath{+2.5}, \ensuremath{+4.3}]} & \ensuremath{+3.3}\newline\mbox{\tiny[\ensuremath{+2.6}, \ensuremath{+4.1}]} & \ensuremath{+3.1}\newline\mbox{\tiny[\ensuremath{+2.2}, \ensuremath{+4.0}]} & \ensuremath{+4.1}\newline\mbox{\tiny[\ensuremath{+3.1}, \ensuremath{+5.1}]} & \ensuremath{+3.6}\newline\mbox{\tiny[\ensuremath{+2.7}, \ensuremath{+4.5}]} & \ensuremath{+2.5}\newline\mbox{\tiny[\ensuremath{+1.7}, \ensuremath{+3.3}]} & \ensuremath{+3.7}\newline\mbox{\tiny[\ensuremath{+2.8}, \ensuremath{+4.6}]} \\
\raggedright owner, 0\%, 100 & \ensuremath{+4.1}\newline\mbox{\tiny[\ensuremath{+3.6}, \ensuremath{+4.6}]} & \ensuremath{+3.2}\newline\mbox{\tiny[\ensuremath{+2.5}, \ensuremath{+3.8}]} & \ensuremath{+3.3}\newline\mbox{\tiny[\ensuremath{+2.5}, \ensuremath{+4.1}]} & \ensuremath{+5.1}\newline\mbox{\tiny[\ensuremath{+4.4}, \ensuremath{+5.8}]} & \ensuremath{+3.4}\newline\mbox{\tiny[\ensuremath{+2.9}, \ensuremath{+4.0}]} & \ensuremath{+1.9}\newline\mbox{\tiny[\ensuremath{+1.6}, \ensuremath{+2.3}]} & \ensuremath{+2.1}\newline\mbox{\tiny[\ensuremath{+1.6}, \ensuremath{+2.7}]} \\
\raggedright owner, 10\%, 30 & \ensuremath{+2.6}\newline\mbox{\tiny[\ensuremath{+1.5}, \ensuremath{+3.8}]} & \ensuremath{+14.3}\newline\mbox{\tiny[\ensuremath{+12.6}, \ensuremath{+16.0}]} & \ensuremath{+10.2}\newline\mbox{\tiny[\ensuremath{+9.2}, \ensuremath{+11.2}]} & \ensuremath{+9.1}\newline\mbox{\tiny[\ensuremath{+7.7}, \ensuremath{+10.5}]} & \ensuremath{+11.1}\newline\mbox{\tiny[\ensuremath{+9.8}, \ensuremath{+12.4}]} & \ensuremath{+3.8}\newline\mbox{\tiny[\ensuremath{+2.9}, \ensuremath{+4.7}]} & \ensuremath{+5.3}\newline\mbox{\tiny[\ensuremath{+4.4}, \ensuremath{+6.3}]} \\
\raggedright shared, 0\%, 30 & \ensuremath{+3.0}\newline\mbox{\tiny[\ensuremath{+2.0}, \ensuremath{+3.9}]} & \ensuremath{+0.3}\newline\mbox{\tiny[\ensuremath{-1.2}, \ensuremath{+1.7}]} & \ensuremath{+1.2}\newline\mbox{\tiny[\ensuremath{0.0}, \ensuremath{+2.4}]} & \ensuremath{+0.2}\newline\mbox{\tiny[\ensuremath{-1.3}, \ensuremath{+1.6}]} & \ensuremath{+0.2}\newline\mbox{\tiny[\ensuremath{-1.1}, \ensuremath{+1.6}]} & \ensuremath{-1.2}\newline\mbox{\tiny[\ensuremath{-2.3}, \ensuremath{0.0}]} & \ensuremath{-2.4}\newline\mbox{\tiny[\ensuremath{-3.3}, \ensuremath{-1.6}]} \\
\raggedright interaction, 0\%, 30 & \ensuremath{+1.5}\newline\mbox{\tiny[\ensuremath{+1.1}, \ensuremath{+2.0}]} & \ensuremath{+1.8}\newline\mbox{\tiny[\ensuremath{+1.2}, \ensuremath{+2.5}]} & \ensuremath{+1.6}\newline\mbox{\tiny[\ensuremath{+1.2}, \ensuremath{+2.1}]} & \ensuremath{+1.7}\newline\mbox{\tiny[\ensuremath{+1.2}, \ensuremath{+2.3}]} & \ensuremath{+1.7}\newline\mbox{\tiny[\ensuremath{+1.2}, \ensuremath{+2.3}]} & \ensuremath{+0.8}\newline\mbox{\tiny[\ensuremath{+0.3}, \ensuremath{+1.2}]} & \ensuremath{+1.3}\newline\mbox{\tiny[\ensuremath{+0.7}, \ensuremath{+1.9}]} \\
\raggedright relaxation, 0\%, 30 & \ensuremath{0.0}\newline\mbox{\tiny[\ensuremath{-0.2}, \ensuremath{+0.2}]} & \ensuremath{-0.2}\newline\mbox{\tiny[\ensuremath{-1.4}, \ensuremath{+0.9}]} & \ensuremath{-1.6}\newline\mbox{\tiny[\ensuremath{-2.6}, \ensuremath{-0.7}]} & \ensuremath{-0.1}\newline\mbox{\tiny[\ensuremath{-0.6}, \ensuremath{+0.3}]} & \ensuremath{-0.3}\newline\mbox{\tiny[\ensuremath{-1.1}, \ensuremath{+0.5}]} & \ensuremath{-0.1}\newline\mbox{\tiny[\ensuremath{-0.4}, \ensuremath{+0.1}]} & \ensuremath{-0.2}\newline\mbox{\tiny[\ensuremath{-0.6}, \ensuremath{+0.2}]} \\
\bottomrule
\end{tabular}
\end{table}

\begin{table}[htbp]
\centering
\caption{Confirmatory run 1 (plan v2, judgement gate): accuracy of every arm in every cell, CV-tuned baselines.}
\label{tab:app-arms-v2}
\tiny
\setlength{\tabcolsep}{3pt}
\begin{tabular}{@{}p{2.4cm}p{1.65cm}p{1.65cm}p{1.65cm}p{1.65cm}p{1.65cm}p{1.65cm}p{1.65cm}@{}}
\toprule
\raggedright arm & \raggedright owner, 0\%, 10 & \raggedright owner, 0\%, 30 & \raggedright owner, 0\%, 100 & \raggedright owner, 10\%, 30 & \raggedright shared, 0\%, 30 & \raggedright interaction, 0\%, 30 & \raggedright relaxation, 0\%, 30 \tabularnewline
\midrule
\raggedright public rule & 90.9\%\newline\mbox{\tiny[89.7\%, 92.1\%]} & 90.9\%\newline\mbox{\tiny[89.7\%, 92.1\%]} & 90.9\%\newline\mbox{\tiny[89.7\%, 92.1\%]} & 90.9\%\newline\mbox{\tiny[89.7\%, 92.1\%]} & 89.6\%\newline\mbox{\tiny[87.9\%, 91.3\%]} & 94.1\%\newline\mbox{\tiny[93.2\%, 95.0\%]} & 93.3\%\newline\mbox{\tiny[92.5\%, 94.2\%]} \\
\raggedright typed, minimal change, gated (primary) & 91.1\%\newline\mbox{\tiny[90.0\%, 92.2\%]} & 95.2\%\newline\mbox{\tiny[94.5\%, 95.8\%]} & 99.2\%\newline\mbox{\tiny[99.0\%, 99.4\%]} & 93.4\%\newline\mbox{\tiny[92.7\%, 94.1\%]} & 95.3\%\newline\mbox{\tiny[94.8\%, 95.8\%]} & 95.3\%\newline\mbox{\tiny[94.8\%, 95.8\%]} & 93.3\%\newline\mbox{\tiny[92.5\%, 94.2\%]} \\
\raggedright typed, pre-stated tie-break, gated & 90.7\%\newline\mbox{\tiny[89.5\%, 92.0\%]} & 95.5\%\newline\mbox{\tiny[95.0\%, 96.0\%]} & 99.4\%\newline\mbox{\tiny[99.2\%, 99.6\%]} & 93.2\%\newline\mbox{\tiny[92.5\%, 93.9\%]} & 95.2\%\newline\mbox{\tiny[94.8\%, 95.7\%]} & 95.2\%\newline\mbox{\tiny[94.7\%, 95.7\%]} & 93.3\%\newline\mbox{\tiny[92.5\%, 94.2\%]} \\
\raggedright typed, minimal change, no gate & 93.2\%\newline\mbox{\tiny[92.3\%, 94.0\%]} & 96.7\%\newline\mbox{\tiny[96.2\%, 97.3\%]} & 99.5\%\newline\mbox{\tiny[99.3\%, 99.6\%]} & 94.9\%\newline\mbox{\tiny[94.3\%, 95.6\%]} & 96.7\%\newline\mbox{\tiny[96.2\%, 97.2\%]} & 95.7\%\newline\mbox{\tiny[95.2\%, 96.2\%]} & 93.3\%\newline\mbox{\tiny[92.5\%, 94.2\%]} \\
\raggedright typed, pre-stated, no gate & -- & 97.4\%\newline\mbox{\tiny[97.1\%, 97.8\%]} & -- & -- & -- & -- & -- \\
\raggedright free-form linear, gated & 90.8\%\newline\mbox{\tiny[89.6\%, 92.1\%]} & 94.8\%\newline\mbox{\tiny[94.3\%, 95.4\%]} & 99.3\%\newline\mbox{\tiny[99.0\%, 99.5\%]} & 91.7\%\newline\mbox{\tiny[90.7\%, 92.7\%]} & 94.2\%\newline\mbox{\tiny[93.4\%, 95.0\%]} & 94.9\%\newline\mbox{\tiny[94.4\%, 95.5\%]} & 93.3\%\newline\mbox{\tiny[92.5\%, 94.2\%]} \\
\raggedright free-form linear, no gate & -- & 96.5\%\newline\mbox{\tiny[95.9\%, 97.2\%]} & -- & -- & -- & -- & -- \\
\raggedright no compiled prior, gated & -- & 27.1\%\newline\mbox{\tiny[25.5\%, 28.7\%]} & -- & -- & -- & -- & -- \\
\raggedright no compiled prior, no gate & 74.7\%\newline\mbox{\tiny[73.8\%, 75.5\%]} & 84.3\%\newline\mbox{\tiny[82.8\%, 85.9\%]} & 90.2\%\newline\mbox{\tiny[89.1\%, 91.3\%]} & 81.7\%\newline\mbox{\tiny[80.6\%, 82.9\%]} & 83.2\%\newline\mbox{\tiny[82.1\%, 84.3\%]} & 85.5\%\newline\mbox{\tiny[84.1\%, 86.8\%]} & 88.1\%\newline\mbox{\tiny[86.9\%, 89.4\%]} \\
\raggedright confined gradient & 90.8\%\newline\mbox{\tiny[89.6\%, 92.0\%]} & 93.3\%\newline\mbox{\tiny[92.5\%, 94.1\%]} & 95.7\%\newline\mbox{\tiny[95.2\%, 96.2\%]} & 91.3\%\newline\mbox{\tiny[90.2\%, 92.4\%]} & 92.3\%\newline\mbox{\tiny[91.0\%, 93.7\%]} & 94.4\%\newline\mbox{\tiny[93.5\%, 95.2\%]} & 93.2\%\newline\mbox{\tiny[92.4\%, 94.0\%]} \\
\raggedright Ditto pFL & 90.0\%\newline\mbox{\tiny[88.8\%, 91.2\%]} & 93.3\%\newline\mbox{\tiny[92.6\%, 94.1\%]} & 97.6\%\newline\mbox{\tiny[97.1\%, 98.0\%]} & 81.3\%\newline\mbox{\tiny[79.5\%, 83.0\%]} & 96.3\%\newline\mbox{\tiny[95.0\%, 97.7\%]} & 94.3\%\newline\mbox{\tiny[93.5\%, 95.1\%]} & 94.1\%\newline\mbox{\tiny[93.0\%, 95.2\%]} \\
\raggedright FedRep pFL & 90.6\%\newline\mbox{\tiny[89.5\%, 91.6\%]} & 93.8\%\newline\mbox{\tiny[92.8\%, 94.8\%]} & 97.3\%\newline\mbox{\tiny[96.5\%, 98.0\%]} & 84.0\%\newline\mbox{\tiny[82.7\%, 85.2\%]} & 94.9\%\newline\mbox{\tiny[93.6\%, 96.2\%]} & 94.3\%\newline\mbox{\tiny[93.6\%, 95.1\%]} & 94.7\%\newline\mbox{\tiny[94.0\%, 95.3\%]} \\
\raggedright FedAvg & 89.8\%\newline\mbox{\tiny[88.6\%, 91.1\%]} & 92.2\%\newline\mbox{\tiny[91.2\%, 93.2\%]} & 94.7\%\newline\mbox{\tiny[93.9\%, 95.5\%]} & 85.0\%\newline\mbox{\tiny[83.9\%, 86.1\%]} & 96.5\%\newline\mbox{\tiny[95.3\%, 97.7\%]} & 94.1\%\newline\mbox{\tiny[93.3\%, 94.9\%]} & 93.2\%\newline\mbox{\tiny[92.2\%, 94.2\%]} \\
\raggedright pooled 79-param net & 90.4\%\newline\mbox{\tiny[89.3\%, 91.5\%]} & 92.9\%\newline\mbox{\tiny[92.1\%, 93.7\%]} & 97.3\%\newline\mbox{\tiny[96.8\%, 97.9\%]} & 83.3\%\newline\mbox{\tiny[81.9\%, 84.8\%]} & 96.3\%\newline\mbox{\tiny[94.9\%, 97.7\%]} & 94.0\%\newline\mbox{\tiny[93.3\%, 94.7\%]} & 93.3\%\newline\mbox{\tiny[92.3\%, 94.2\%]} \\
\raggedright pooled large net & 87.8\%\newline\mbox{\tiny[86.6\%, 89.0\%]} & 94.8\%\newline\mbox{\tiny[94.2\%, 95.3\%]} & 97.9\%\newline\mbox{\tiny[97.5\%, 98.4\%]} & 90.4\%\newline\mbox{\tiny[89.5\%, 91.4\%]} & 97.1\%\newline\mbox{\tiny[96.2\%, 98.0\%]} & 95.1\%\newline\mbox{\tiny[94.5\%, 95.6\%]} & 93.2\%\newline\mbox{\tiny[92.5\%, 94.0\%]} \\
\raggedright pooled gradient-boosted trees & 87.8\%\newline\mbox{\tiny[86.6\%, 89.1\%]} & 92.4\%\newline\mbox{\tiny[91.5\%, 93.4\%]} & 97.5\%\newline\mbox{\tiny[97.0\%, 98.0\%]} & 88.4\%\newline\mbox{\tiny[87.1\%, 89.6\%]} & 98.9\%\newline\mbox{\tiny[98.5\%, 99.2\%]} & 94.7\%\newline\mbox{\tiny[94.1\%, 95.3\%]} & 93.8\%\newline\mbox{\tiny[93.1\%, 94.5\%]} \\
\bottomrule
\end{tabular}
\end{table}

\begin{table}[htbp]
\centering
\caption{Confirmatory run 1 (plan v2, judgement gate): paired differences, typed learner (primary arm) minus each CV-tuned baseline, every cell. Exploratory except the owner, 0\%, 30-record cell.}
\label{tab:app-diff-v2}
\tiny
\setlength{\tabcolsep}{3pt}
\begin{tabular}{@{}p{1.8cm}p{1.7cm}p{1.7cm}p{1.7cm}p{1.7cm}p{1.7cm}p{1.7cm}p{1.7cm}@{}}
\toprule
\raggedright cell & \raggedright confined gradient & \raggedright Ditto pFL & \raggedright FedRep pFL & \raggedright FedAvg & \raggedright pooled 79-param net & \raggedright pooled large net & \raggedright pooled gradient-boosted trees \tabularnewline
\midrule
\raggedright owner, 0\%, 10 & \ensuremath{+0.3}\newline\mbox{\tiny[\ensuremath{-0.1}, \ensuremath{+0.7}]} & \ensuremath{+1.1}\newline\mbox{\tiny[\ensuremath{+0.4}, \ensuremath{+1.8}]} & \ensuremath{+0.5}\newline\mbox{\tiny[\ensuremath{0.0}, \ensuremath{+1.1}]} & \ensuremath{+1.3}\newline\mbox{\tiny[\ensuremath{+0.7}, \ensuremath{+1.9}]} & \ensuremath{+0.7}\newline\mbox{\tiny[\ensuremath{+0.3}, \ensuremath{+1.0}]} & \ensuremath{+3.3}\newline\mbox{\tiny[\ensuremath{+2.1}, \ensuremath{+4.4}]} & \ensuremath{+3.3}\newline\mbox{\tiny[\ensuremath{+2.1}, \ensuremath{+4.4}]} \\
\raggedright owner, 0\%, 30 & \ensuremath{+1.9}\newline\mbox{\tiny[\ensuremath{+1.0}, \ensuremath{+2.8}]} & \ensuremath{+1.8}\newline\mbox{\tiny[\ensuremath{+0.8}, \ensuremath{+2.9}]} & \ensuremath{+1.4}\newline\mbox{\tiny[\ensuremath{+0.1}, \ensuremath{+2.7}]} & \ensuremath{+3.0}\newline\mbox{\tiny[\ensuremath{+2.0}, \ensuremath{+3.9}]} & \ensuremath{+2.3}\newline\mbox{\tiny[\ensuremath{+1.4}, \ensuremath{+3.1}]} & \ensuremath{+0.4}\newline\mbox{\tiny[\ensuremath{-0.3}, \ensuremath{+1.1}]} & \ensuremath{+2.7}\newline\mbox{\tiny[\ensuremath{+1.8}, \ensuremath{+3.7}]} \\
\raggedright owner, 0\%, 100 & \ensuremath{+3.5}\newline\mbox{\tiny[\ensuremath{+3.0}, \ensuremath{+4.1}]} & \ensuremath{+1.7}\newline\mbox{\tiny[\ensuremath{+1.1}, \ensuremath{+2.2}]} & \ensuremath{+1.9}\newline\mbox{\tiny[\ensuremath{+1.1}, \ensuremath{+2.8}]} & \ensuremath{+4.5}\newline\mbox{\tiny[\ensuremath{+3.7}, \ensuremath{+5.4}]} & \ensuremath{+1.9}\newline\mbox{\tiny[\ensuremath{+1.3}, \ensuremath{+2.5}]} & \ensuremath{+1.3}\newline\mbox{\tiny[\ensuremath{+0.8}, \ensuremath{+1.8}]} & \ensuremath{+1.7}\newline\mbox{\tiny[\ensuremath{+1.2}, \ensuremath{+2.3}]} \\
\raggedright owner, 10\%, 30 & \ensuremath{+2.2}\newline\mbox{\tiny[\ensuremath{+1.4}, \ensuremath{+2.9}]} & \ensuremath{+12.2}\newline\mbox{\tiny[\ensuremath{+10.3}, \ensuremath{+14.0}]} & \ensuremath{+9.5}\newline\mbox{\tiny[\ensuremath{+8.2}, \ensuremath{+10.7}]} & \ensuremath{+8.4}\newline\mbox{\tiny[\ensuremath{+7.3}, \ensuremath{+9.5}]} & \ensuremath{+10.1}\newline\mbox{\tiny[\ensuremath{+8.7}, \ensuremath{+11.5}]} & \ensuremath{+3.0}\newline\mbox{\tiny[\ensuremath{+2.2}, \ensuremath{+3.8}]} & \ensuremath{+5.1}\newline\mbox{\tiny[\ensuremath{+3.9}, \ensuremath{+6.3}]} \\
\raggedright shared, 0\%, 30 & \ensuremath{+3.0}\newline\mbox{\tiny[\ensuremath{+1.6}, \ensuremath{+4.3}]} & \ensuremath{-1.1}\newline\mbox{\tiny[\ensuremath{-2.4}, \ensuremath{+0.3}]} & \ensuremath{+0.4}\newline\mbox{\tiny[\ensuremath{-0.9}, \ensuremath{+1.8}]} & \ensuremath{-1.2}\newline\mbox{\tiny[\ensuremath{-2.5}, \ensuremath{+0.1}]} & \ensuremath{-1.0}\newline\mbox{\tiny[\ensuremath{-2.4}, \ensuremath{+0.4}]} & \ensuremath{-1.8}\newline\mbox{\tiny[\ensuremath{-2.9}, \ensuremath{-0.7}]} & \ensuremath{-3.6}\newline\mbox{\tiny[\ensuremath{-4.3}, \ensuremath{-2.9}]} \\
\raggedright interaction, 0\%, 30 & \ensuremath{+0.9}\newline\mbox{\tiny[\ensuremath{+0.4}, \ensuremath{+1.4}]} & \ensuremath{+1.0}\newline\mbox{\tiny[\ensuremath{+0.5}, \ensuremath{+1.4}]} & \ensuremath{+0.9}\newline\mbox{\tiny[\ensuremath{+0.5}, \ensuremath{+1.4}]} & \ensuremath{+1.2}\newline\mbox{\tiny[\ensuremath{+0.7}, \ensuremath{+1.7}]} & \ensuremath{+1.3}\newline\mbox{\tiny[\ensuremath{+0.8}, \ensuremath{+1.7}]} & \ensuremath{+0.2}\newline\mbox{\tiny[\ensuremath{-0.1}, \ensuremath{+0.5}]} & \ensuremath{+0.6}\newline\mbox{\tiny[\ensuremath{+0.2}, \ensuremath{+1.0}]} \\
\raggedright relaxation, 0\%, 30 & \ensuremath{+0.2}\newline\mbox{\tiny[\ensuremath{0.0}, \ensuremath{+0.4}]} & \ensuremath{-0.7}\newline\mbox{\tiny[\ensuremath{-1.7}, \ensuremath{+0.2}]} & \ensuremath{-1.3}\newline\mbox{\tiny[\ensuremath{-2.1}, \ensuremath{-0.5}]} & \ensuremath{+0.1}\newline\mbox{\tiny[\ensuremath{-0.6}, \ensuremath{+0.8}]} & \ensuremath{+0.1}\newline\mbox{\tiny[\ensuremath{-0.9}, \ensuremath{+1.0}]} & \ensuremath{+0.1}\newline\mbox{\tiny[\ensuremath{-0.3}, \ensuremath{+0.4}]} & \ensuremath{-0.5}\newline\mbox{\tiny[\ensuremath{-1.2}, \ensuremath{+0.3}]} \\
\bottomrule
\end{tabular}
\end{table}

\begin{table}[htbp]
\centering
\caption{Secondary family at the primary cell (owner, 0\%, 30 records), both confirmatory runs: mean truth-level difference in points with its 95\% $t$-interval. The Holm $p$ columns give Holm-adjusted sign-flip $p$ values (Holm across S1--S10 in run 1, S1--S11 in run 2). S11 exists only in run 2.}
\label{tab:app-secondary}
\small
\setlength{\tabcolsep}{3pt}
\begin{tabular}{@{}p{4.4cm}p{2.8cm}p{1.6cm}p{2.8cm}p{1.6cm}@{}}
\toprule
\raggedright comparison & \raggedright run 1 difference & \raggedright run 1 Holm $p$ & \raggedright run 2 difference & \raggedright run 2 Holm $p$ \tabularnewline
\midrule
\raggedright S1 typed $-$ confined gradient & \ensuremath{+1.9}\newline\mbox{\tiny[\ensuremath{+1.0}, \ensuremath{+2.8}]} & 0.0011 & \ensuremath{+3.4}\newline\mbox{\tiny[\ensuremath{+2.5}, \ensuremath{+4.3}]} & <0.0001 \\
\raggedright S2 typed $-$ Ditto pFL & \ensuremath{+1.8}\newline\mbox{\tiny[\ensuremath{+0.8}, \ensuremath{+2.9}]} & 0.0019 & \ensuremath{+3.3}\newline\mbox{\tiny[\ensuremath{+2.6}, \ensuremath{+4.1}]} & <0.0001 \\
\raggedright S3 typed $-$ FedRep pFL & \ensuremath{+1.4}\newline\mbox{\tiny[\ensuremath{+0.1}, \ensuremath{+2.7}]} & 0.0618 & \ensuremath{+3.1}\newline\mbox{\tiny[\ensuremath{+2.2}, \ensuremath{+4.0}]} & <0.0001 \\
\raggedright S4 typed $-$ FedAvg & \ensuremath{+3.0}\newline\mbox{\tiny[\ensuremath{+2.0}, \ensuremath{+3.9}]} & <0.0001 & \ensuremath{+4.1}\newline\mbox{\tiny[\ensuremath{+3.1}, \ensuremath{+5.1}]} & <0.0001 \\
\raggedright S5 typed $-$ pooled 79-param net & \ensuremath{+2.3}\newline\mbox{\tiny[\ensuremath{+1.4}, \ensuremath{+3.1}]} & 0.0002 & \ensuremath{+3.6}\newline\mbox{\tiny[\ensuremath{+2.7}, \ensuremath{+4.5}]} & <0.0001 \\
\raggedright S6 typed $-$ pooled large net & \ensuremath{+0.4}\newline\mbox{\tiny[\ensuremath{-0.3}, \ensuremath{+1.1}]} & 0.2620 & \ensuremath{+2.5}\newline\mbox{\tiny[\ensuremath{+1.7}, \ensuremath{+3.3}]} & <0.0001 \\
\raggedright S7 typed $-$ pooled gradient-boosted trees & \ensuremath{+2.7}\newline\mbox{\tiny[\ensuremath{+1.8}, \ensuremath{+3.7}]} & <0.0001 & \ensuremath{+3.7}\newline\mbox{\tiny[\ensuremath{+2.8}, \ensuremath{+4.6}]} & <0.0001 \\
\raggedright S8 typed $-$ public rule & \ensuremath{+4.3}\newline\mbox{\tiny[\ensuremath{+3.2}, \ensuremath{+5.3}]} & <0.0001 & \ensuremath{+4.9}\newline\mbox{\tiny[\ensuremath{+3.5}, \ensuremath{+6.3}]} & <0.0001 \\
\raggedright S9 gated $-$ ungated & \ensuremath{-1.6}\newline\mbox{\tiny[\ensuremath{-1.9}, \ensuremath{-1.2}]} & <0.0001 & \ensuremath{-0.1}\newline\mbox{\tiny[\ensuremath{-0.1}, \ensuremath{0.0}]} & 0.1764 \\
\raggedright S10 minimal change $-$ pre-stated tie-break & \ensuremath{-0.3}\newline\mbox{\tiny[\ensuremath{-0.6}, \ensuremath{0.0}]} & 0.1066 & \ensuremath{+0.1}\newline\mbox{\tiny[\ensuremath{-0.6}, \ensuremath{+0.7}]} & 0.8309 \\
\raggedright S11 tuned gate $-$ judgement gate & -- & -- & \ensuremath{+1.3}\newline\mbox{\tiny[\ensuremath{+0.9}, \ensuremath{+1.6}]} & <0.0001 \\
\bottomrule
\end{tabular}
\end{table}

\begin{table}[htbp]
\centering
\caption{Oracle-tuned sensitivity, confirmatory run 2 (plan v3): each baseline with its configuration chosen on the deployment truth, which no real party has. These are upper bounds on what better tuning could reach and were never used in a test. The typed learner's row is repeated for comparison.}
\label{tab:app-oracle-ev3}
\scriptsize
\setlength{\tabcolsep}{3pt}
\begin{tabular}{@{}p{4.6cm}p{2.4cm}p{2.4cm}p{2.4cm}p{2.4cm}@{}}
\toprule
\raggedright arm & \raggedright owner, 0\%, 10 & \raggedright owner, 0\%, 30 & \raggedright owner, 0\%, 100 & \raggedright owner, 10\%, 30 \tabularnewline
\midrule
\raggedright typed learner (primary), run 2 (plan v3) & 93.0\%\newline\mbox{\tiny[92.1\%, 93.9\%]} & 96.6\%\newline\mbox{\tiny[96.2\%, 96.9\%]} & 99.6\%\newline\mbox{\tiny[99.5\%, 99.8\%]} & 94.5\%\newline\mbox{\tiny[94.0\%, 95.1\%]} \\
\raggedright confined gradient, oracle-tuned & 92.6\% & 94.2\% & 96.4\% & 92.7\% \\
\raggedright Ditto pFL, oracle-tuned & 93.4\% & 95.6\% & 97.8\% & 90.9\% \\
\raggedright FedRep pFL, oracle-tuned & 92.8\% & 94.8\% & 97.0\% & 87.4\% \\
\raggedright FedAvg, oracle-tuned & 92.7\% & 93.6\% & 95.2\% & 87.8\% \\
\raggedright pooled 79-param net, oracle-tuned & 92.6\% & 94.0\% & 97.0\% & 87.0\% \\
\raggedright pooled large net, oracle-tuned & 90.7\% & 94.8\% & 98.0\% & 91.7\% \\
\raggedright pooled gradient-boosted trees, oracle-tuned & 90.6\% & 93.7\% & 97.9\% & 90.2\% \\
\bottomrule
\end{tabular}
\end{table}

\begin{table}[htbp]
\centering
\caption{Oracle-tuned sensitivity, confirmatory run 1 (plan v2): each baseline with its configuration chosen on the deployment truth, which no real party has. These are upper bounds on what better tuning could reach and were never used in a test. The typed learner's row is repeated for comparison.}
\label{tab:app-oracle-ev}
\scriptsize
\setlength{\tabcolsep}{3pt}
\begin{tabular}{@{}p{4.6cm}p{2.4cm}p{2.4cm}p{2.4cm}p{2.4cm}@{}}
\toprule
\raggedright arm & \raggedright owner, 0\%, 10 & \raggedright owner, 0\%, 30 & \raggedright owner, 0\%, 100 & \raggedright owner, 10\%, 30 \tabularnewline
\midrule
\raggedright typed learner (primary), run 1 (plan v2) & 91.1\%\newline\mbox{\tiny[90.0\%, 92.2\%]} & 95.2\%\newline\mbox{\tiny[94.5\%, 95.8\%]} & 99.2\%\newline\mbox{\tiny[99.0\%, 99.4\%]} & 93.4\%\newline\mbox{\tiny[92.7\%, 94.1\%]} \\
\raggedright confined gradient, oracle-tuned & 92.4\% & 94.4\% & 96.2\% & 92.3\% \\
\raggedright Ditto pFL, oracle-tuned & 93.4\% & 95.9\% & 98.5\% & 91.0\% \\
\raggedright FedRep pFL, oracle-tuned & 92.6\% & 95.3\% & 97.8\% & 87.2\% \\
\raggedright FedAvg, oracle-tuned & 91.9\% & 93.5\% & 95.5\% & 87.6\% \\
\raggedright pooled 79-param net, oracle-tuned & 92.0\% & 94.2\% & 97.9\% & 87.0\% \\
\raggedright pooled large net, oracle-tuned & 90.3\% & 95.5\% & 98.2\% & 91.9\% \\
\raggedright pooled gradient-boosted trees, oracle-tuned & 88.6\% & 93.3\% & 98.0\% & 89.7\% \\
\bottomrule
\end{tabular}
\end{table}

\begin{table}[htbp]
\centering
\caption{The evidence gate in run 2 (tuned setting), every cell, and the exploratory null check on fresh draws. Exact recovery is the share of party-draws whose declaration is identical on the grid to the sampled truth.}
\label{tab:app-gate}
\scriptsize
\setlength{\tabcolsep}{3pt}
\begin{tabular}{@{}p{3.6cm}p{2.4cm}p{2.4cm}p{2.4cm}p{2.4cm}@{}}
\toprule
\raggedright cell & \raggedright proposes & \raggedright worse than public (gated) & \raggedright worse than public (no gate) & \raggedright exact recovery \tabularnewline
\midrule
\raggedright owner, 0\%, 10 & 22.3\%\newline\mbox{\tiny[18.0\%, 27.4\%]} & 7.5\%\newline\mbox{\tiny[3.2\%, 16.3\%]} & 17.0\%\newline\mbox{\tiny[11.8\%, 23.9\%]} & 3.7\%\newline\mbox{\tiny[2.1\%, 6.4\%]} \\
\raggedright owner, 0\%, 30 & 77.0\%\newline\mbox{\tiny[71.9\%, 81.4\%]} & 6.5\%\newline\mbox{\tiny[4.0\%, 10.4\%]} & 7.5\%\newline\mbox{\tiny[4.8\%, 11.5\%]} & 28.3\%\newline\mbox{\tiny[23.5\%, 33.7\%]} \\
\raggedright owner, 0\%, 100 & 99.0\%\newline\mbox{\tiny[97.1\%, 99.7\%]} & 1.0\%\newline\mbox{\tiny[0.3\%, 2.9\%]} & 1.0\%\newline\mbox{\tiny[0.3\%, 2.9\%]} & 77.7\%\newline\mbox{\tiny[72.6\%, 82.0\%]} \\
\raggedright owner, 10\%, 30 & 50.7\%\newline\mbox{\tiny[45.0\%, 56.3\%]} & 17.8\%\newline\mbox{\tiny[12.5\%, 24.6\%]} & 28.0\%\newline\mbox{\tiny[22.9\%, 33.8\%]} & -- \\
\raggedright shared, 0\%, 30 & 84.7\%\newline\mbox{\tiny[80.2\%, 88.3\%]} & 2.8\%\newline\mbox{\tiny[1.3\%, 5.6\%]} & 3.7\%\newline\mbox{\tiny[2.0\%, 6.7\%]} & 24.3\%\newline\mbox{\tiny[19.8\%, 29.5\%]} \\
\raggedright interaction, 0\%, 30 & 65.3\%\newline\mbox{\tiny[59.8\%, 70.5\%]} & 19.9\%\newline\mbox{\tiny[14.9\%, 26.0\%]} & 26.1\%\newline\mbox{\tiny[20.8\%, 32.2\%]} & -- \\
\raggedright relaxation, 0\%, 30 & 0.0\%\newline\mbox{\tiny[0.0\%, 1.3\%]} & -- & -- & -- \\
\midrule
\raggedright public-rule null, n10 (exploratory) & 3.0\%\newline\mbox{\tiny[1.9\%, 4.7\%]} & -- & -- & -- \\
\raggedright \quad judgement gate, n10 & 0.0\%\newline\mbox{\tiny[0.0\%, 0.6\%]} & -- & -- & -- \\
\raggedright public-rule null, n30 (exploratory) & 3.8\%\newline\mbox{\tiny[2.6\%, 5.7\%]} & -- & -- & -- \\
\raggedright \quad judgement gate, n30 & 0.2\%\newline\mbox{\tiny[0.0\%, 0.9\%]} & -- & -- & -- \\
\raggedright public-rule null, n100 (exploratory) & 6.2\%\newline\mbox{\tiny[4.5\%, 8.4\%]} & -- & -- & -- \\
\raggedright \quad judgement gate, n100 & 0.3\%\newline\mbox{\tiny[0.1\%, 1.2\%]} & -- & -- & -- \\
\raggedright public-rule null, pooled (exploratory) & 4.3\%\newline\mbox{\tiny[3.5\%, 5.4\%]} & -- & -- & -- \\
\raggedright \quad judgement gate, pooled & 0.2\%\newline\mbox{\tiny[0.1\%, 0.5\%]} & -- & -- & -- \\
\bottomrule
\end{tabular}
\end{table}

\begin{table}[htbp]
\centering
\caption{Trained base (registered), 300 shared and 30 private records per party: accuracy by arm and family, values disclosed per party, the primary differences and the secondary family. The last column gives values disclosed per party in the arm rows and the Holm-adjusted sign-flip $p$ in rows S1--S6, printed in the floating-point notation in which it is recorded. Differences are the combination minus the named comparator, except S6, which is as labelled; S5 is the combination minus the same combination without its gate.}
\label{tab:app-hybrid}
\scriptsize
\setlength{\tabcolsep}{3pt}
\begin{tabular}{@{}p{4.4cm}p{2.4cm}p{2.4cm}p{2.4cm}p{1.8cm}@{}}
\toprule
\raggedright arm or comparison & \raggedright all 40 truths & \raggedright interaction & \raggedright relaxation & \raggedright values / Holm $p$ \tabularnewline
\midrule
\raggedright public rule & 85.5\%\newline\mbox{\tiny[84.2\%, 86.7\%]} & 85.2\%\newline\mbox{\tiny[83.2\%, 87.2\%]} & 85.8\%\newline\mbox{\tiny[84.1\%, 87.4\%]} & 0.0 \\
\raggedright declarations over the public rule & 91.3\%\newline\mbox{\tiny[90.4\%, 92.1\%]} & 92.5\%\newline\mbox{\tiny[91.7\%, 93.4\%]} & 90.0\%\newline\mbox{\tiny[88.6\%, 91.4\%]} & 2.3 \\
\raggedright trained base alone & 88.9\%\newline\mbox{\tiny[88.0\%, 89.9\%]} & 88.9\%\newline\mbox{\tiny[87.5\%, 90.3\%]} & 89.0\%\newline\mbox{\tiny[87.5\%, 90.5\%]} & 0.0 \\
\raggedright combination (primary) & 94.3\%\newline\mbox{\tiny[93.8\%, 94.8\%]} & 93.4\%\newline\mbox{\tiny[92.7\%, 94.0\%]} & 95.3\%\newline\mbox{\tiny[94.8\%, 95.8\%]} & 2.2 \\
\raggedright combination, no gate & 95.6\%\newline\mbox{\tiny[95.0\%, 96.2\%]} & 94.3\%\newline\mbox{\tiny[93.6\%, 95.0\%]} & 97.0\%\newline\mbox{\tiny[96.6\%, 97.3\%]} & 3.7 \\
\raggedright confined gradient & 91.6\%\newline\mbox{\tiny[91.0\%, 92.2\%]} & 91.1\%\newline\mbox{\tiny[90.4\%, 91.8\%]} & 92.1\%\newline\mbox{\tiny[91.1\%, 93.1\%]} & 3.9 \\
\raggedright private fine-tuning & 92.4\%\newline\mbox{\tiny[91.8\%, 93.0\%]} & 91.8\%\newline\mbox{\tiny[91.3\%, 92.4\%]} & 93.0\%\newline\mbox{\tiny[92.0\%, 94.0\%]} & 0.0 \\
\raggedright Ditto pFL & 92.4\%\newline\mbox{\tiny[91.9\%, 93.0\%]} & 92.1\%\newline\mbox{\tiny[91.5\%, 92.7\%]} & 92.8\%\newline\mbox{\tiny[91.8\%, 93.8\%]} & 991.9 \\
\raggedright pooled small net & 89.6\%\newline\mbox{\tiny[88.7\%, 90.5\%]} & 89.8\%\newline\mbox{\tiny[88.5\%, 91.2\%]} & 89.3\%\newline\mbox{\tiny[88.0\%, 90.7\%]} & 210.0 \\
\raggedright declarations over public rule, no gate & 92.6\%\newline\mbox{\tiny[91.8\%, 93.4\%]} & 93.2\%\newline\mbox{\tiny[92.4\%, 94.1\%]} & 91.9\%\newline\mbox{\tiny[90.6\%, 93.3\%]} & 3.8 \\
\midrule
\raggedright D1 combination $-$ base & \ensuremath{+5.4}\newline\mbox{\tiny[\ensuremath{+4.4}, \ensuremath{+6.4}]} & -- & -- & -- \\
\raggedright D2 combination $-$ declarations on public rule & \ensuremath{+3.1}\newline\mbox{\tiny[\ensuremath{+2.0}, \ensuremath{+4.1}]} & -- & -- & -- \\
\raggedright S1 Ditto pFL & \ensuremath{+1.9}\newline\mbox{\tiny[\ensuremath{+1.2}, \ensuremath{+2.6}]} & -- & -- & 2.99999e-05 \\
\raggedright S2 private fine-tuning & \ensuremath{+1.9}\newline\mbox{\tiny[\ensuremath{+1.3}, \ensuremath{+2.6}]} & -- & -- & 2.99999e-05 \\
\raggedright S3 confined gradient & \ensuremath{+2.7}\newline\mbox{\tiny[\ensuremath{+2.0}, \ensuremath{+3.4}]} & -- & -- & 2.99999e-05 \\
\raggedright S4 pooled small net & \ensuremath{+4.8}\newline\mbox{\tiny[\ensuremath{+3.7}, \ensuremath{+5.8}]} & -- & -- & 2.99999e-05 \\
\raggedright S5 no gate & \ensuremath{-1.3}\newline\mbox{\tiny[\ensuremath{-1.5}, \ensuremath{-1.0}]} & -- & -- & 2.99999e-05 \\
\raggedright S6 base $-$ declarations on public rule & \ensuremath{-2.3}\newline\mbox{\tiny[\ensuremath{-3.6}, \ensuremath{-1.1}]} & -- & -- & 0.000454998 \\
\bottomrule
\end{tabular}
\end{table}

\begin{table}[htbp]
\centering
\caption{Trained base (registered), every cell: the combination's accuracy and its difference from each comparator. Exploratory except 300 shared, 30 private.}
\label{tab:app-hybrid-cells}
\tiny
\setlength{\tabcolsep}{3pt}
\begin{tabular}{@{}p{1.8cm}p{1.65cm}p{1.65cm}p{1.65cm}p{1.65cm}p{1.65cm}p{1.65cm}p{1.65cm}@{}}
\toprule
\raggedright cell & \raggedright combination & \raggedright base & \raggedright decl.\ on public & \raggedright Ditto pFL & \raggedright fine-tuning & \raggedright confined grad. & \raggedright pooled \tabularnewline
\midrule
\raggedright 100 shared, 10 private & 87.7\%\newline\mbox{\tiny[86.7\%, 88.7\%]} & \ensuremath{+0.4}\newline\mbox{\tiny[\ensuremath{+0.2}, \ensuremath{+0.6}]} & \ensuremath{+1.8}\newline\mbox{\tiny[\ensuremath{+1.1}, \ensuremath{+2.5}]} & \ensuremath{-0.5}\newline\mbox{\tiny[\ensuremath{-1.0}, \ensuremath{0.0}]} & \ensuremath{-0.3}\newline\mbox{\tiny[\ensuremath{-0.7}, \ensuremath{0.0}]} & \ensuremath{-0.1}\newline\mbox{\tiny[\ensuremath{-0.3}, \ensuremath{+0.2}]} & \ensuremath{-0.4}\newline\mbox{\tiny[\ensuremath{-0.7}, \ensuremath{0.0}]} \\
\raggedright 100 shared, 30 private & 92.9\%\newline\mbox{\tiny[92.2\%, 93.5\%]} & \ensuremath{+5.5}\newline\mbox{\tiny[\ensuremath{+4.5}, \ensuremath{+6.6}]} & \ensuremath{+1.7}\newline\mbox{\tiny[\ensuremath{+0.7}, \ensuremath{+2.6}]} & \ensuremath{+1.2}\newline\mbox{\tiny[\ensuremath{+0.6}, \ensuremath{+1.9}]} & \ensuremath{+1.4}\newline\mbox{\tiny[\ensuremath{+0.7}, \ensuremath{+2.0}]} & \ensuremath{+2.6}\newline\mbox{\tiny[\ensuremath{+1.9}, \ensuremath{+3.3}]} & \ensuremath{+3.2}\newline\mbox{\tiny[\ensuremath{+2.3}, \ensuremath{+4.1}]} \\
\raggedright 100 shared, 100 private & 96.9\%\newline\mbox{\tiny[96.3\%, 97.4\%]} & \ensuremath{+9.6}\newline\mbox{\tiny[\ensuremath{+8.5}, \ensuremath{+10.6}]} & \ensuremath{+1.6}\newline\mbox{\tiny[\ensuremath{+0.7}, \ensuremath{+2.4}]} & \ensuremath{+0.8}\newline\mbox{\tiny[\ensuremath{+0.4}, \ensuremath{+1.2}]} & \ensuremath{+1.0}\newline\mbox{\tiny[\ensuremath{+0.6}, \ensuremath{+1.4}]} & \ensuremath{+3.5}\newline\mbox{\tiny[\ensuremath{+3.0}, \ensuremath{+4.0}]} & \ensuremath{+2.2}\newline\mbox{\tiny[\ensuremath{+1.7}, \ensuremath{+2.8}]} \\
\raggedright 300 shared, 10 private & 89.2\%\newline\mbox{\tiny[88.3\%, 90.1\%]} & \ensuremath{+0.3}\newline\mbox{\tiny[\ensuremath{+0.1}, \ensuremath{+0.5}]} & \ensuremath{+3.4}\newline\mbox{\tiny[\ensuremath{+2.6}, \ensuremath{+4.2}]} & \ensuremath{-0.1}\newline\mbox{\tiny[\ensuremath{-0.5}, \ensuremath{+0.2}]} & \ensuremath{0.0}\newline\mbox{\tiny[\ensuremath{-0.3}, \ensuremath{+0.2}]} & \ensuremath{-0.1}\newline\mbox{\tiny[\ensuremath{-0.4}, \ensuremath{+0.2}]} & \ensuremath{+0.2}\newline\mbox{\tiny[\ensuremath{-0.1}, \ensuremath{+0.5}]} \\
\raggedright 300 shared, 30 private & 94.3\%\newline\mbox{\tiny[93.8\%, 94.8\%]} & \ensuremath{+5.4}\newline\mbox{\tiny[\ensuremath{+4.4}, \ensuremath{+6.4}]} & \ensuremath{+3.1}\newline\mbox{\tiny[\ensuremath{+2.0}, \ensuremath{+4.1}]} & \ensuremath{+1.9}\newline\mbox{\tiny[\ensuremath{+1.2}, \ensuremath{+2.6}]} & \ensuremath{+1.9}\newline\mbox{\tiny[\ensuremath{+1.3}, \ensuremath{+2.6}]} & \ensuremath{+2.7}\newline\mbox{\tiny[\ensuremath{+2.0}, \ensuremath{+3.4}]} & \ensuremath{+4.8}\newline\mbox{\tiny[\ensuremath{+3.7}, \ensuremath{+5.8}]} \\
\raggedright 300 shared, 100 private & 98.3\%\newline\mbox{\tiny[97.8\%, 98.8\%]} & \ensuremath{+9.3}\newline\mbox{\tiny[\ensuremath{+8.3}, \ensuremath{+10.4}]} & \ensuremath{+2.9}\newline\mbox{\tiny[\ensuremath{+1.9}, \ensuremath{+3.9}]} & \ensuremath{+2.1}\newline\mbox{\tiny[\ensuremath{+1.7}, \ensuremath{+2.5}]} & \ensuremath{+2.0}\newline\mbox{\tiny[\ensuremath{+1.6}, \ensuremath{+2.4}]} & \ensuremath{+3.7}\newline\mbox{\tiny[\ensuremath{+3.3}, \ensuremath{+4.1}]} & \ensuremath{+6.0}\newline\mbox{\tiny[\ensuremath{+5.1}, \ensuremath{+6.9}]} \\
\raggedright 1000 shared, 10 private & 89.7\%\newline\mbox{\tiny[88.8\%, 90.6\%]} & \ensuremath{+0.3}\newline\mbox{\tiny[\ensuremath{+0.1}, \ensuremath{+0.5}]} & \ensuremath{+3.8}\newline\mbox{\tiny[\ensuremath{+2.9}, \ensuremath{+4.6}]} & \ensuremath{0.0}\newline\mbox{\tiny[\ensuremath{-0.4}, \ensuremath{+0.4}]} & \ensuremath{0.0}\newline\mbox{\tiny[\ensuremath{-0.3}, \ensuremath{+0.2}]} & \ensuremath{0.0}\newline\mbox{\tiny[\ensuremath{-0.3}, \ensuremath{+0.2}]} & \ensuremath{+0.1}\newline\mbox{\tiny[\ensuremath{-0.2}, \ensuremath{+0.4}]} \\
\raggedright 1000 shared, 30 private & 94.8\%\newline\mbox{\tiny[94.4\%, 95.2\%]} & \ensuremath{+5.4}\newline\mbox{\tiny[\ensuremath{+4.3}, \ensuremath{+6.5}]} & \ensuremath{+3.6}\newline\mbox{\tiny[\ensuremath{+2.5}, \ensuremath{+4.7}]} & \ensuremath{+2.5}\newline\mbox{\tiny[\ensuremath{+1.7}, \ensuremath{+3.4}]} & \ensuremath{+2.3}\newline\mbox{\tiny[\ensuremath{+1.5}, \ensuremath{+3.0}]} & \ensuremath{+2.9}\newline\mbox{\tiny[\ensuremath{+2.2}, \ensuremath{+3.7}]} & \ensuremath{+5.2}\newline\mbox{\tiny[\ensuremath{+4.1}, \ensuremath{+6.3}]} \\
\raggedright 1000 shared, 100 private & 98.8\%\newline\mbox{\tiny[98.5\%, 99.1\%]} & \ensuremath{+9.4}\newline\mbox{\tiny[\ensuremath{+8.3}, \ensuremath{+10.4}]} & \ensuremath{+3.4}\newline\mbox{\tiny[\ensuremath{+2.5}, \ensuremath{+4.4}]} & \ensuremath{+2.9}\newline\mbox{\tiny[\ensuremath{+2.5}, \ensuremath{+3.4}]} & \ensuremath{+2.6}\newline\mbox{\tiny[\ensuremath{+2.2}, \ensuremath{+3.0}]} & \ensuremath{+3.9}\newline\mbox{\tiny[\ensuremath{+3.5}, \ensuremath{+4.4}]} & \ensuremath{+9.0}\newline\mbox{\tiny[\ensuremath{+7.9}, \ensuremath{+10.2}]} \\
\bottomrule
\end{tabular}
\end{table}

\begin{table}[htbp]
\centering
\caption{Trained-base sensitivity check (exploratory, after unblinding): the registered combination minus each comparator, 300 shared records. Oracle arms choose their configuration on deployment truth, which no real party can do.}
\label{tab:app-hybrid-sens}
\scriptsize
\setlength{\tabcolsep}{3pt}
\begin{tabular}{@{}p{4.6cm}p{3.2cm}p{3.2cm}p{3.2cm}@{}}
\toprule
\raggedright comparator & \raggedright 10 private & \raggedright 30 private & \raggedright 100 private \tabularnewline
\midrule
\raggedright combination over better base & \ensuremath{-0.4}\newline\mbox{\tiny[\ensuremath{-0.5}, \ensuremath{-0.2}]} & \ensuremath{-0.3}\newline\mbox{\tiny[\ensuremath{-0.5}, \ensuremath{-0.1}]} & \ensuremath{-0.3}\newline\mbox{\tiny[\ensuremath{-0.4}, \ensuremath{-0.1}]} \\
\raggedright trained base (CV) & \ensuremath{+0.3}\newline\mbox{\tiny[\ensuremath{+0.1}, \ensuremath{+0.5}]} & \ensuremath{+5.4}\newline\mbox{\tiny[\ensuremath{+4.4}, \ensuremath{+6.4}]} & \ensuremath{+9.3}\newline\mbox{\tiny[\ensuremath{+8.3}, \ensuremath{+10.4}]} \\
\raggedright trained base, oracle & \ensuremath{-0.2}\newline\mbox{\tiny[\ensuremath{-0.5}, \ensuremath{+0.1}]} & \ensuremath{+4.9}\newline\mbox{\tiny[\ensuremath{+3.8}, \ensuremath{+6.0}]} & \ensuremath{+8.8}\newline\mbox{\tiny[\ensuremath{+7.7}, \ensuremath{+10.0}]} \\
\raggedright better-trained base & \ensuremath{-0.1}\newline\mbox{\tiny[\ensuremath{-0.3}, \ensuremath{+0.2}]} & \ensuremath{+5.0}\newline\mbox{\tiny[\ensuremath{+3.9}, \ensuremath{+6.1}]} & \ensuremath{+9.0}\newline\mbox{\tiny[\ensuremath{+7.9}, \ensuremath{+10.1}]} \\
\raggedright private fine-tuning, oracle & \ensuremath{-2.4}\newline\mbox{\tiny[\ensuremath{-2.9}, \ensuremath{-1.8}]} & \ensuremath{-0.3}\newline\mbox{\tiny[\ensuremath{-0.9}, \ensuremath{+0.2}]} & \ensuremath{+1.0}\newline\mbox{\tiny[\ensuremath{+0.7}, \ensuremath{+1.3}]} \\
\raggedright fine-tuning, better base, CV & \ensuremath{-0.5}\newline\mbox{\tiny[\ensuremath{-0.9}, \ensuremath{-0.2}]} & \ensuremath{+1.7}\newline\mbox{\tiny[\ensuremath{+1.0}, \ensuremath{+2.3}]} & \ensuremath{+1.9}\newline\mbox{\tiny[\ensuremath{+1.5}, \ensuremath{+2.3}]} \\
\raggedright fine-tuning, better base, oracle & \ensuremath{-2.6}\newline\mbox{\tiny[\ensuremath{-3.3}, \ensuremath{-2.0}]} & \ensuremath{-0.4}\newline\mbox{\tiny[\ensuremath{-1.0}, \ensuremath{+0.1}]} & \ensuremath{+0.9}\newline\mbox{\tiny[\ensuremath{+0.5}, \ensuremath{+1.3}]} \\
\raggedright Ditto pFL, oracle & \ensuremath{-2.1}\newline\mbox{\tiny[\ensuremath{-2.6}, \ensuremath{-1.6}]} & \ensuremath{-0.1}\newline\mbox{\tiny[\ensuremath{-0.6}, \ensuremath{+0.4}]} & \ensuremath{+0.9}\newline\mbox{\tiny[\ensuremath{+0.6}, \ensuremath{+1.3}]} \\
\raggedright Ditto pFL, better base, CV & \ensuremath{-0.5}\newline\mbox{\tiny[\ensuremath{-1.0}, \ensuremath{-0.1}]} & \ensuremath{+1.8}\newline\mbox{\tiny[\ensuremath{+1.1}, \ensuremath{+2.5}]} & \ensuremath{+2.1}\newline\mbox{\tiny[\ensuremath{+1.6}, \ensuremath{+2.5}]} \\
\raggedright Ditto pFL, better base, oracle & \ensuremath{-2.3}\newline\mbox{\tiny[\ensuremath{-2.8}, \ensuremath{-1.8}]} & \ensuremath{-0.2}\newline\mbox{\tiny[\ensuremath{-0.7}, \ensuremath{+0.4}]} & \ensuremath{+0.9}\newline\mbox{\tiny[\ensuremath{+0.5}, \ensuremath{+1.2}]} \\
\raggedright confined gradient, oracle & \ensuremath{-1.5}\newline\mbox{\tiny[\ensuremath{-1.9}, \ensuremath{-1.0}]} & \ensuremath{+1.4}\newline\mbox{\tiny[\ensuremath{+0.7}, \ensuremath{+2.1}]} & \ensuremath{+3.0}\newline\mbox{\tiny[\ensuremath{+2.6}, \ensuremath{+3.4}]} \\
\raggedright pooled small net, long grid, CV & \ensuremath{+0.2}\newline\mbox{\tiny[\ensuremath{-0.1}, \ensuremath{+0.5}]} & \ensuremath{+4.8}\newline\mbox{\tiny[\ensuremath{+3.7}, \ensuremath{+5.8}]} & \ensuremath{+6.1}\newline\mbox{\tiny[\ensuremath{+5.2}, \ensuremath{+6.9}]} \\
\raggedright pooled small net, oracle & \ensuremath{-0.7}\newline\mbox{\tiny[\ensuremath{-1.0}, \ensuremath{-0.5}]} & \ensuremath{+3.1}\newline\mbox{\tiny[\ensuremath{+2.2}, \ensuremath{+4.0}]} & \ensuremath{+3.9}\newline\mbox{\tiny[\ensuremath{+3.3}, \ensuremath{+4.5}]} \\
\raggedright pooled 128-unit net, CV & \ensuremath{+0.2}\newline\mbox{\tiny[\ensuremath{-0.3}, \ensuremath{+0.7}]} & \ensuremath{+4.1}\newline\mbox{\tiny[\ensuremath{+3.0}, \ensuremath{+5.2}]} & \ensuremath{+5.2}\newline\mbox{\tiny[\ensuremath{+4.3}, \ensuremath{+6.1}]} \\
\raggedright pooled 128-unit net, oracle & \ensuremath{0.0}\newline\mbox{\tiny[\ensuremath{-0.5}, \ensuremath{+0.6}]} & \ensuremath{+3.9}\newline\mbox{\tiny[\ensuremath{+2.8}, \ensuremath{+5.0}]} & \ensuremath{+5.0}\newline\mbox{\tiny[\ensuremath{+4.1}, \ensuremath{+5.8}]} \\
\raggedright pooled boosted trees, CV & \ensuremath{-0.3}\newline\mbox{\tiny[\ensuremath{-0.8}, \ensuremath{+0.2}]} & \ensuremath{+4.6}\newline\mbox{\tiny[\ensuremath{+3.4}, \ensuremath{+5.8}]} & \ensuremath{+6.7}\newline\mbox{\tiny[\ensuremath{+5.6}, \ensuremath{+7.7}]} \\
\raggedright pooled boosted trees, oracle & \ensuremath{-0.7}\newline\mbox{\tiny[\ensuremath{-1.2}, \ensuremath{-0.2}]} & \ensuremath{+3.8}\newline\mbox{\tiny[\ensuremath{+2.7}, \ensuremath{+5.0}]} & \ensuremath{+5.4}\newline\mbox{\tiny[\ensuremath{+4.5}, \ensuremath{+6.4}]} \\
\raggedright per-party boosted trees, CV & \ensuremath{-0.5}\newline\mbox{\tiny[\ensuremath{-0.9}, \ensuremath{0.0}]} & \ensuremath{+4.2}\newline\mbox{\tiny[\ensuremath{+3.1}, \ensuremath{+5.4}]} & \ensuremath{+5.5}\newline\mbox{\tiny[\ensuremath{+4.6}, \ensuremath{+6.5}]} \\
\raggedright per-party boosted trees, oracle & \ensuremath{-0.9}\newline\mbox{\tiny[\ensuremath{-1.4}, \ensuremath{-0.4}]} & \ensuremath{+3.4}\newline\mbox{\tiny[\ensuremath{+2.2}, \ensuremath{+4.5}]} & \ensuremath{+4.3}\newline\mbox{\tiny[\ensuremath{+3.4}, \ensuremath{+5.1}]} \\
\bottomrule
\end{tabular}
\end{table}

\begin{table}[htbp]
\centering
\caption{Fixed-truth replication (plan v1, exploratory): three fixed truths, 30 draws, owner truth, 0\% noise, 30 records per party. Intervals are across draws.}
\label{tab:app-rep}
\small
\setlength{\tabcolsep}{3pt}
\begin{tabular}{@{}p{6cm}p{4cm}@{}}
\toprule
\raggedright arm or comparison & \raggedright accuracy or difference \tabularnewline
\midrule
\raggedright public rule & 88.3\% \\
\raggedright typed, pre-stated, gated (v1 primary) & 96.1\% \\
\raggedright typed, no gate & 97.1\% \\
\raggedright typed, minimal change, gated & 96.3\% \\
\raggedright typed, minimal change, no gate & 97.5\% \\
\raggedright free-form linear, gated & 94.4\% \\
\raggedright no compiled prior, no gate & 87.3\% \\
\raggedright confined gradient, CV & 90.0\% \\
\raggedright confined gradient, oracle & 92.4\% \\
\raggedright Ditto pFL, CV & 90.0\% \\
\raggedright Ditto pFL, oracle & 93.8\% \\
\raggedright FedRep pFL, CV & 90.3\% \\
\raggedright FedAvg, CV & 89.1\% \\
\raggedright pooled 79-param net, CV & 89.7\% \\
\raggedright pooled large net, CV & 92.4\% \\
\raggedright pooled large net, oracle & 93.2\% \\
\midrule
\raggedright P1 typed $-$ Ditto pFL & \ensuremath{+6.1} \\
\raggedright P2 typed $-$ confined gradient & \ensuremath{+6.1} \\
\raggedright P3 typed $-$ pooled large net & \ensuremath{+3.7} \\
\raggedright typed $-$ public rule & \ensuremath{+7.8} \\
\bottomrule
\end{tabular}
\end{table}

\end{document}